\documentclass[acmsmall]{acmart}

\usepackage{multirow}
\usepackage{enumitem}

\usepackage{subfigure}
\usepackage{xspace}
\usepackage{multirow}
\usepackage{xcolor}
\usepackage{amsmath}
\usepackage{amsfonts}
\usepackage{bm}
\usepackage{soul}
\usepackage[ruled]{algorithm2e}
\usepackage{balance}

\AtBeginDocument{%
  \providecommand\BibTeX{{%
    \normalfont B\kern-0.5em{\scshape i\kern-0.25em b}\kern-0.8em\TeX}}}

\setcopyright{acmcopyright}
\copyrightyear{2018}
\acmYear{2018}
\acmDOI{10.1145/1122445.1122456}

\acmJournal{JACM}
\acmVolume{37}
\acmNumber{4}
\acmArticle{111}
\acmMonth{8}

\newcommand{\eg}{\emph{e.g.},\xspace}

\newcommand{\ie}{\emph{i.e.},\xspace}

\newcommand{\eat}[1]{}

\begin{document}

\title{Machine Learning for Urban Air Quality Analytics: A Survey}


\author{Jindong Han}
\affiliation{%
  \institution{Hong Kong University of Science and Technology}
  \city{Hong Kong}
  \country{China}}
\email{jhanao@connect.ust.hk}

\author{Weijia Zhang}
\affiliation{%
  \institution{Hong Kong University of Science and Technology (Guangzhou)}
  \city{Guangzhou}
  \country{China}
}
\email{wzhang411@connect.hkust-gz.edu.cn}

\author{Hao Liu, Hui Xiong}
\affiliation{%
  \institution{Hong Kong University of Science and Technology (Guangzhou) and Hong Kong University of Science and Technology}
  \country{China}
}
\email{{liuh, xionghui}@ust.hk}

\renewcommand{\shortauthors}{Trovato and Tobin, et al.}

\begin{abstract}
The increasing air pollution poses an urgent global concern with far-reaching consequences, such as premature mortality and reduced crop yield, which significantly impact various aspects of our daily lives. Accurate and timely analysis of air pollution is crucial for understanding its underlying mechanisms and implementing necessary precautions to mitigate potential socio-economic losses. 
Traditional analytical methodologies, such as atmospheric modeling, heavily rely on domain expertise and often make simplified assumptions that may not be applicable to complex air pollution problems. In contrast, Machine Learning (ML) models are able to capture the intrinsic physical and chemical rules by automatically learning from a large amount of historical observational data, showing great promise in various air quality analytical tasks.
In this article, we present a comprehensive survey of ML-based air quality analytics, following a roadmap spanning from data acquisition to pre-processing, and encompassing various analytical tasks such as pollution pattern mining, air quality inference, and forecasting. Moreover, we offer a systematic categorization and summary of existing methodologies and applications, while also providing a list of publicly available air quality datasets to ease the research in this direction. Finally, we identify several promising future research directions.
This survey can serve as a valuable resource for professionals seeking suitable solutions for their specific challenges and advancing their research at the cutting edge.
\end{abstract}

\ccsdesc[500]{Applied computing~Environmental sciences}
\ccsdesc[300]{Computing methodologies~Machine learning}

\keywords{Air quality, machine learning, urban computing}

\maketitle
\section{Introduction}
Air pollution poses tremendous threats to public health and environmental sustainability across cities worldwide. According to the World Health Organization (WHO), air pollution has increased the risk of various health issues among citizens and imposed a significant economic burden on society~\cite{world2015economic}. Therefore, air quality analytics has become vital for society and its individuals. On the one hand, accurate analysis of air pollution enables policymakers to formulate effective environmental regulations and targeted interventions for mitigating pollution emissions. On the other hand, it can also empower individuals to make informed decisions, such as adjusting travel routes or reducing outdoor activities, to minimize exposure to harmful pollutants. Consequently, air quality analytics has gained significant attention in recent decades, leading to the emergence of diverse research directions and applications, such as pollution pattern mining~\cite{akbari2015generic}, air quality inference~\cite{zheng2013u}, and forecasting~\cite{zheng2015forecasting}. 
These advancements have paved the way for a better understanding of air pollution and have enabled the development of more accurate air quality monitoring and forecasting systems.

Traditional air quality analytics typically relies on numerical simulation of atmospheric models. However, such simulation-based methods are computationally expensive and demand extensive domain knowledge~\cite{bi2022pangu,lam2022graphcast}. Moreover, such approaches inevitably fall short in the ability to capture complex mechanisms of air pollution due to incomplete knowledge about the atmospheric system and a variety of factors involved~\cite{vardoulakis2003modelling}.
As an appealing alternative, Machine Learning (ML) has provided great opportunities to tackle urban air pollution tasks from a data-driven perspective~\cite{zheng2013u,zheng2015forecasting}. Unlike traditional methods that require a comprehensive understanding of the inner mechanisms of air pollution, ML techniques capture "underlying laws" directly from historical data, achieving promising results with a much lower computational budget. Given such capability, ML-based air quality analytics has attracted considerable attention from multiple research areas, including computer science~\cite{zheng2014urban}, environmental science~\cite{zhong2021machine,liu2022data}, and sociology~\cite{zheng2019air}.

However, developing effective and efficient ML algorithms for air pollution analysis is particularly challenging due to the heterogeneous, sparse, and spatio-temporal nature of air quality observational data. Specifically, these challenges manifest in three key aspects, including: \textbf{(1) Heterogeneous model inputs}. Analyzing and modeling urban air pollution needs to harness information from multiple data sources, such as local meteorology, traffic flow, pollution emissions, and human activities. However, the data from different sources are highly heterogeneous, with each characterized by distinct spatial resolutions, modalities, structures, and densities, making it difficult to integrate them. According to previous studies~\cite{zheng2015methodologies}, simply combining features from different sources yields unsatisfactory performance or even compromises the models. Therefore, it is crucial to devise advanced ML techniques that can effectively assimilate knowledge from heterogeneous pollution-related data. \textbf{(2) Insufficient data coverage}. ML models usually require a large amount of observational data to achieve good performance. However, due to economic concerns, there are only a limited number of monitoring sensors deployed in a city, resulting in data sparsity issue (e.g., only 0.2\% of data are observed in Beijing)~\cite{zheng2013u,xu2019fine}. The sparsely and non-uniformly observed air quality may deviate from the true distribution of the entire dataset, thereby introducing bias in subsequent analytical tasks. Thus, how to develop data-efficient ML techniques for air quality analytics is also a prominent challenge. \textbf{(3) Complex spatio-temporal dependencies among pollutants}. Air pollution exhibits complicated spatio-temporal dependencies due to the propagation and chemical reactions of different pollutants over time and space. For instance, strong winds blowing from one location to another can transport pollutants, thereby enhancing the correlation among locations. Conversely, a change in wind direction can weaken such correlations~\cite{liang2018geoman}. Traditional ML models that rely on feature engineering, such as support vector machine (SVM)~\cite{drucker1996support} and random forest (RF)~\cite{breiman2001random}, are unable to handle such complex and non-linear dynamic dependencies. So, there is a rising demand to design more sophisticated ML models that can effectively capture spatio-temporal dependencies among pollutants.

In the past decade, a vast amount of ML techniques have been developed and integrated into urban air quality analytics to address the aforementioned challenges. 
Thus, a comprehensive survey of existing methodologies and applications in this area becomes highly scientific and practical.
Although some surveys have been published in this field~\cite{zheng2014urban,maag2018survey,huang2021overview,concas2021low,kaur2023computational}, most of them mainly focus on specific parts of the
analytics pipeline or overlook the latest high-quality literature. For instance, Zheng et al.~\cite{zheng2014urban} specify several applications based on multi-source urban data, Kaur et al.~\cite{kaur2023computational} review forecasting methods while ignoring other aspects of air quality analytics, Concas et al.~\cite{concas2021low} mainly pays attention to low-cost air quality sensor calibration and discusses ML algorithms utilized for calibration.
As a result, newcomers to this field may still encounter difficulties and obstacles due to a lack of comprehensive understanding regarding the key research problems, existing methodologies, potential traps, and promising directions for subsequent research.
To this end, we provide a systematic overview of the current state-of-the-art in ML-based air quality analytics, as depicted in Figure 1.

\begin{figure}[t]
	\centering
	\includegraphics[width=13cm]{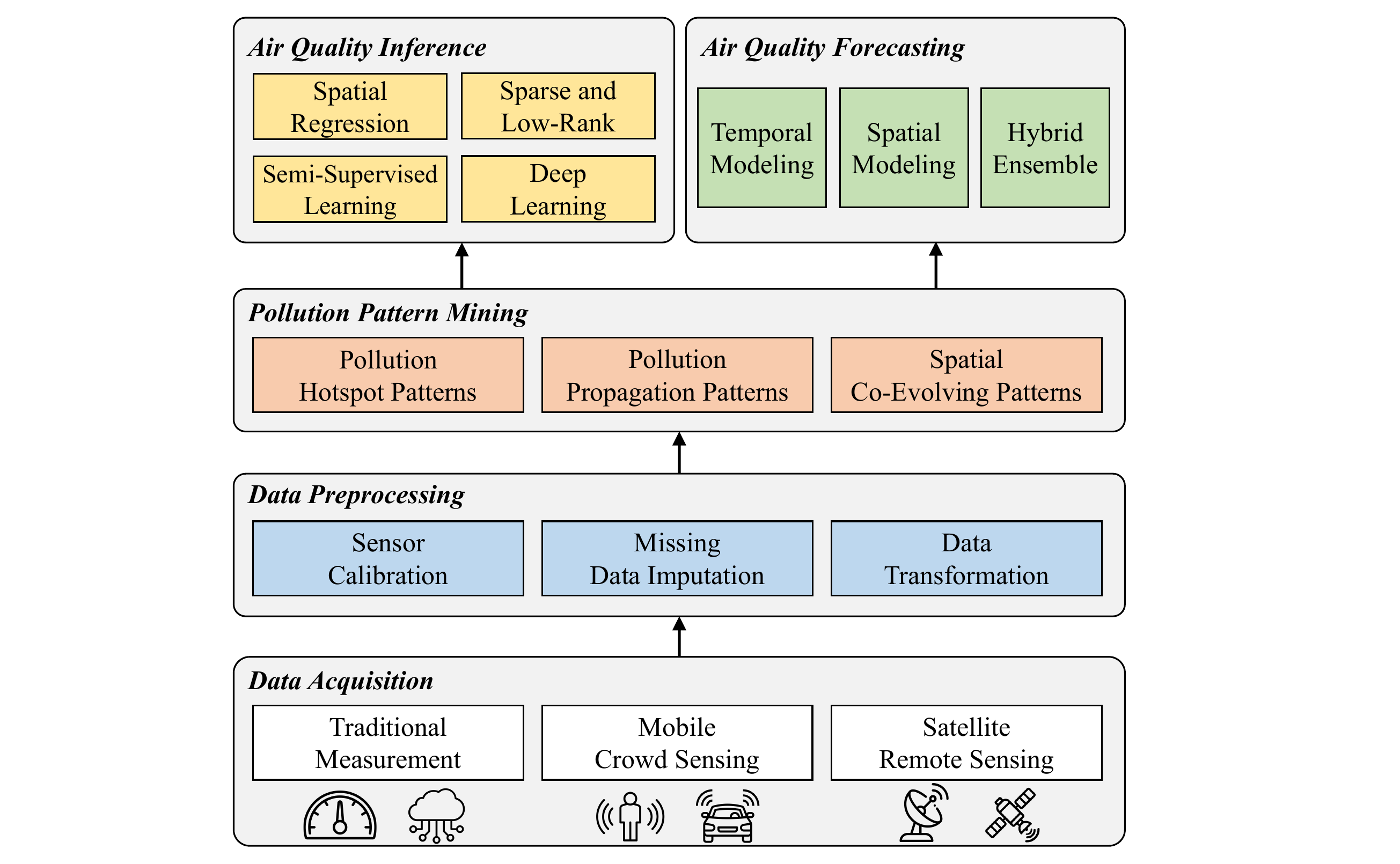}
	\caption{Framework of urban air quality analytics.}
	\label{fig:infer}
\end{figure}

First, in Section~2.2, we classify the existing data acquisition technologies for urban air quality into three categories: traditional measurement, mobile crowd sensing, and satellite remote sensing. We also list and discuss key techniques under each category. Second, before using air quality data, we need to conduct pre-processing on the raw measurements, as they are noisy and usually contain missing values. We summarize three data pre-processing techniques for urban air quality in Section~2.3, including sensor calibration, missing data imputation, and data transformation. Sensor calibration aims to correct or adjust unreliable sensor readings using known reference values. Missing data imputation is reconstructing missing air quality values by leveraging past and future observations and concurrent measurements from neighboring sensors. Data transformation enables the conversion of raw data into specific formats suitable for the application of various ML models. Third, building upon the first two steps, we can then conduct air quality analytical tasks, including: \emph{(1) Pollution pattern mining}. The huge volume of urban air quality data provides an opportunity for discovering critical pollution patterns. These patterns have significant implications for various applications, such as enhancing the performance of air quality inference and forecasting, and even pinpointing the root causes of air pollution. In Section 3, we explore the literature related to three categories of patterns: hotspot patterns, pollution propagation patterns, and spatial co-evolving patterns.
\emph{(2) Air quality inference}. As previously discussed, obtaining fine-grained air quality information is difficult as we only have limited monitoring sensors. To achieve this goal, a series of research tried to reconstruct the entire pollution map from sparse observations with ML models. We review air quality inference in Section 4. \emph{(3) Air quality forecasting}. Predicting future air quality based on past observations is a challenging and significant problem. ML techniques have demonstrated their effectiveness in making predictions by learning intricate relationships from historical data. In Section 5, we present representative algorithms of air quality forecasting. 

To summarize, this article makes contributions in the following aspects:
\begin{itemize}
	\item We present a generic framework for ML-based urban air quality analytics, summarizing key research problems and methodologies in this field. The framework establishes the primary scope and roadmap for urban air quality analytics, thereby enabling the community to gain a better understanding and engagement in this emerging area.
	\item Existing research works are well organized and interconnected in this framework. We categorize the analytical models according to problems and corresponding techniques, discussing the relationships, strengths, and weaknesses among different subcategories, as well as the technical details within each subcategory. The proposed framework can help professionals identify most suitable methods to address their specific problems.
	\item We collect and list existing public air quality datasets, which could facilitate the development of new algorithms and applications in this field. Moreover, we conclude by providing valuable insights into the current challenges and outlining potential directions for future research.
\end{itemize}

\section{Data Acquisition and Pre-processing}
In this section, we first introduce the types of air pollutants. Then we present an overview of the fundamental technologies commonly used for air quality data acquisition. Finally, we elaborate on several pre-processing techniques applied to the collected data.

\subsection{Types of Air Pollutants}
The sources of air pollutants can be either natural (\eg dust storms and forest fires) or generated through human activities (\eg waste incineration, building construction, and fossil fuel power plants). According to the composition, air pollutants can be categorized into two classes: particulate matter and gaseous pollutants~\cite{concas2021low}. Particulate matter, also known as aerosols, are minuscule particles of liquid or solid substances suspended in the air. Existing research classifies particles based on their sizes. For example, coarse particulates (PM$_{10}$) refer to particulate matter with diameters of 10 micrometers or less, whereas the diameter of fine particulates (PM$_{2.5}$) is less than 2.5 micrometers. In a given environment, we can employ particulate matter sensors to detect and count particles within specific diameter ranges.
Gaseous pollutants usually consist of various chemical compounds, including ozone (O$_3$), carbon monoxide (CO), carbon dioxide (CO$_2$), sulfur oxides, nitrogen oxides, and volatile organic compounds. Unlike particulate matter sensors, which lack the capacity to distinguish specific compositions or sources, gaseous sensors can identify different kinds of gases by utilizing different sensing materials or by altering sensor parameters.

To enhance public awareness about air pollution and facilitate decision-making, government agencies have developed the air quality index (AQI) as the indicator~\cite{zheng2013u}. A higher AQI implies that people will suffer from increasingly severe adverse health effects. In practice, the AQI is computed based on the concentration levels of several key pollutants, including PM$_{2.5}$, PM$_{10}$, ozone, carbon monoxide, nitrogen dioxide (NO$_2$), and sulfur dioxide (SO$_2$). Since different regions may have their own standards and methods, the formula used to transform pollutant concentrations into specific AQI values varies by countries~\cite{bishoi2009comparative,xu2020air}.

\subsection{Data Acquisition}
Air quality data can be represented as multivariate time-series comprising concentrations of different pollutants.
In this section, we classify the current data acquisition technologies into three types: traditional measurement, mobile crowd sensing, and satellite remote sensing. We provide detailed discussion for each category as follows.

\subsubsection{Traditional measurement}
Traditional air quality measurements typically rely on professional monitoring stations or low-cost sensors integrated into the urban infrastructure. Over the past decade, many cities have established ground-based monitoring stations to provide accurate air quality information for government authorities and citizens. However, professional monitoring stations suffer from limited spatial coverage due to expensive installation and maintenance costs~\cite{zheng2013u}. On the other hand, low-cost air quality sensor usually costs less than 500 US dollars~\cite{concas2021low}, making them an affordable alternative to measurement stations. A series of research studies have been done to measure fine-grained air quality by integrating low-cost sensors into fixed (\emph{e.g.} light poles)~\cite{cheng2019ict,motlagh2020toward} or public transportation (\emph{e.g.} trams)~\cite{hasenfratz2014pushing,hasenfratz2015deriving,gao2016mosaic} infrastructures. 
For example, Cheng et al.~\cite{cheng2019ict} constructed a large-scale sensor network with thousands of low-cost sensor nodes in Beijing to measure the concentration of PM2.5, humidity, and temperature. The data from these individual sensor nodes were collected via a cloud server at one-minute intervals to generate up-to-date city-wide pollution maps. Moreover, Hasenfratz et al.~\cite{hasenfratz2014pushing,hasenfratz2015deriving} installed the air quality sensors on the top of streetcars to monitor the concentration of ultrafine particles. Although high-density deployment of low-cost sensors can be achieved, these sensors are susceptible to low accuracy issues due to environmental interference and inherent vulnerability, necessitating laborious periodic re-calibration to obtain accurate measurement values~\cite{concas2021low}.

\subsubsection{Mobile crowd sensing}
Mobile crowd sensing~\cite{guo2015mobile} is an emerging sensing paradigm that actively collects environmental information around the crowds through mobile sensing devices, such as smartphones, GPS trackers, and wearable sensors. Compared to traditional methods, mobile crowd sensing has two major advantages. First, it leverages existing mobile devices and communication infrastructures, which largely reduces the cost of sensor deployment and maintenance. Second, the inherent mobility of mobile device users significantly expands the spatial coverage of air quality monitoring. Consequently, mobile crowd sensing has been widely used in air quality data acquisition~\cite{pan2017crowdsensing,liu2018third,maag2018w,wu2020sharing}. For instance, Third-Eye~\cite{liu2018third} developed a mobile crowd sensing system to produce accurate estimation of air quality levels. Specifically, the authors first recruited 200 people living near monitoring stations to take outdoor photos via their mobile phones. Then the system estimates air quality based on these images using deep learning techniques. Moreover, UbiAir~\cite{wu2020sharing} installed mobile sensors on shared bikes to monitor ambient air quality. Cyclists are rewarded with gifts if they successfully collect corresponding air quality observations at desired locations. Despite these advances, mobile crowd sensing approaches are still not satisfactory due to the lack of effective task assignment and incentive schemes~\cite{tong2020spatial}. Additionally, since the preference and bias of participating users are usually diverse, the data quality cannot be guaranteed during the sensing process.

In addition to physical sensors, one can also consider social media users as "social sensors". The data generated from mobile social networking services, such as geotagged tweets and photos, can describe the surrounding events and environment changes of users. Consequently, the analysis of social media data can help estimate the real-time pollution situations~\cite{mei2014inferring,jiang2015using} or predict future air quality~\cite{jiang2019enhancing}.
However, such sensor-free solutions cannot provide precise pollution measurements and are not applicable in urban areas where social media records are unavailable.

\subsubsection{Satellite remote sensing}
While ground-based monitoring can provide accurate air quality observations, they are sparsely distributed in suburban and rural areas, making it inadequate to characterize the national-scale or global-scale spatial variability of pollutant concentration. Satellite remote sensing (SRS) is an alternative solution that can enable air quality monitoring on a large scale. By scanning the surface of the earth, SRS generates wide spectral images reflecting the physical characteristics of land and atmosphere, which can be adopted to assess the pollution situations of aerosols (such as PM$_{2.5}$, PM$_{10}$) and trace gases (such as O$_3$, N$_2$O)~\cite{martin2008satellite}. For example, the satellite-derived aerosol optical depth (AOD) data can be employed for the inversion of aerosol pollution maps~\cite{xu2016remote,xu2019fine}. However, SRS can be easily disturbed by meteorological conditions such as clouds, and fails to differentiate between pollutants from atmosphere and ground surface~\cite{van2006estimating}.

\subsection{Data Pre-Processing}
Data pre-processing is a crucial step in which raw data are cleaned and transformed into formats suitable for further analysis. Raw air quality measurements often suffer from inherent noise~\cite{concas2021low} and contain numerous missing values~\cite{yi2016st}, which may introduce bias and inaccuracies in analytical results. Therefore, it is imperative to perform pre-processing on the raw data. In the following section, we provide a concise overview of basic pre-processing techniques utilized for air quality data before starting an analytical task.

\subsubsection{Sensor calibration}
As mentioned in Section 2.1.1, low-cost sensors are susceptible to the influence of meteorological conditions~\cite{masson2015quantification} and cross-sensitivities among different pollutants~\cite{cross2017use}, thus they are less reliable than professional monitoring stations. While periodic re-calibration can enhance the accuracy of low-cost sensors, it is a time-consuming and labor-intensive process~\cite{ramanathan2006rapid}.
In recent years, machine learning techniques have emerged as an effective and practical tool for improving calibration accuracy and reducing associated workloads~\cite{cheng2019ict,lin2018calibrating,yu2020airnet}. These approaches treat sensor calibration as a regression problem and aim to train machine learning models that map unreliable sensor readings to accurate reference values (i.e., the measurements of a professional station close to the target sensor) with the help of external information, such as temperature, humidity, and wind speed. For example, AirNet~\cite{yu2020airnet} first treated sensor calibration as a sequence-to-point mapping problem, and then proposed a dual sequence encoder network to obtain the corrected measurement by integrating the historical observations from both sensor and reference station. The calibrated sensor measurements can then be utilized in subsequent applications. More details about machine learning-based sensor calibration can be found in Concas et al.~\cite{concas2021low}.

\subsubsection{Missing data imputation}
The presence of missing values is a common occurrence in the air quality data acquisition process due to sensor malfunction or communication failures~\cite{yi2016st}. Missing data problem not only poses a significant challenge to real-time monitoring, but also undermines the performance of subsequent analytical tasks, such as air quality inference and forecasting. Extensive research has been conducted over the past decades to address the imputation of missing values in time series data. In this article, we briefly discuss several approaches specifically relevant to air quality analysis. Early works for the missing data imputation usually leverage either interpolation or linear regression approaches~\cite{junninen2004methods}. However, these methods are incapable of capturing complicated spatio-temporal dependencies, leading to unsatisfactory results. Yi et al.~\cite{yi2016st} proposed a multi-view learning approach to enhance the accuracy of air quality data imputation by jointly considering spatio-temporal correlations from both global and local perspectives. More recently, deep autoregressive models have been employed as the common workhorse in time series imputation~\cite{cao2018brits,luo2018multivariate,luo2019e2gan,miao2021generative,cini2021filling}. For example, BRITS~\cite{cao2018brits} utilizes a bidirectional Recurrent Neural Network (RNN) to reconstruct missing parts of air quality data, while capturing spatial correlations through a linear regression layer. Based upon the deep autoregressive paradigm, several recent studies have developed more sophisticated architectures coupled with advanced techniques, including adversarial learning~\cite{luo2018multivariate,luo2019e2gan}, semi-supervised learning~\cite{miao2021generative}, and graph neural networks~\cite{cini2021filling}. However, autoregressive models suffer from the error accumulation issue and can be easily disrupted to make biased imputation results when air quality observations are highly sparse. To alleviate such issue, SPIN~\cite{marisca2022learning} proposed a pure attention-based model that can impute missing data points without propagating prediction errors.

\subsubsection{Data transformation}
Air quality data can be represented in four common formats when used as input for machine learning models, including sequence, two-dimensional matrix, three-dimensional tensor, and graph~\cite{wang2020deep}. In practice, air quality measurements from a single location naturally form a geo-tagged sequence, allowing the utilization of sequence learning techniques~\cite{sutskever2014sequence} for data processing. Sometimes it is necessary to take into account the concentrations of multiple pollutants or air quality observations across all the locations as a whole for analysis. To this end, we can convert the data into matrices or tensors. For the case of matrices, the first dimension of the matrices can be pollution types or locations and the second dimension represents time stamps~\cite{shang2014inferring,yi2018deep}. For the case of tensors, the three dimensions represent locations, time stamps, and pollution types, respectively~\cite{xu2016remote,xu2019fine}. Nevertheless, matrices and tensors only capture temporal information and neglect spatial relations among different locations, leading to potential information loss. To overcome this limitation, researchers proposed to incorporate spatial structure information by leveraging graphs, where locations serve as nodes and spatial relations between locations are represented as edges~\cite{han2021joint,wang2021modeling,han2022semi}.

Besides air quality data, analyzing air pollution often requires the simultaneous consideration of multiple influential factors such as meteorology and traffic. For instance, the U-Air project~\cite{zheng2013u} aims to estimate fine-grained air quality throughout the city by leveraging data from heterogeneous sources, including meteorology, point-of-interests (POIs), road networks, and trajectories. More details about pre-processing the urban data from other sources can refer to Zheng et al.~\cite{zheng2014urban}.
\section{Pollution Pattern Mining}
Air pollution exhibits numerous complicated patterns in space and time, ranging from individual behaviors to group dynamics. Accurate analysis of these patterns provides valuable insights into diverse pollution-related applications, including root cause diagnosis, propagation path discovery, air quality inference, and forecasting~\cite{li2017discovering,deng2019airvis}.
In this section, we discuss three categories of pollution patterns: hotspot patterns, pollution propagation patterns, and spatial co-evolving patterns.

\subsection{Pollution Hotspot Patterns}
A branch of research aims to discover specific regions with high pollution emissions, known as pollution hotspots. In real-world scenarios, pollution hotspots are highly dynamic, changing across both space and time. For instance, vehicle emissions on different roads are largely influenced by current traffic conditions, while factories are considered as pollution sources only during periods when they release waste gases. Additionally, various random events, such as straw burning, fireworks displays, and building construction, may also form potential pollution hotspots.

Traditionally, environmental scientists rely on either dispersion models~\cite{keats2007bayesian} or chemical receptor models~\cite{lee2008source} to detect pollution hotspots. These methods heavily depend on coarse-grained air quality measurements obtained from specific monitoring stations, thereby overlooking a significant number of local pollution hotspots. Li et al.~\cite{li2017discovering} focused on discovering local hotspots of PM$_{2.5}$ by leveraging frequent subgraph mining algorithm~\cite{jiang2013survey}, where the source nodes in the mined frequent subgraph are treated as pollution hotspots. Nevertheless, this approach still relies on data collected from sparse sensors, making it unsuitable for pollutants that exhibit highly localized hotspots, such as nitrogen oxides. In addition, important contextual information, such as point-of-interests (POIs) and traffic data, are also ignored in the model. Given the aforementioned limitations, Zhang et al.~\cite{zhang2021air} proposed to detect pollution hotspots using irregularly sampled mobile sensing data, which provides air quality information at a finer granularity. They first extract local spikes from raw observations and then aggregate them via mean shift clustering algorithm~\cite{cheng1995mean}. 
Moreover, they further train a random forest model to predict hotspots in cities without mobile sensing data by incorporating various contextual information like POI features as input. 
Besides identifying pollution hotspots, air pollution data can also be utilized to predict other interesting hotspot, such as COVID-19 outbreak hotspots~\cite{segovia2021does}.

\subsection{Pollution Propagation Patterns}
Pollution propagation patterns can help uncover the movement behaviors of air pollutants across different locations. Such movement behaviors provide a concise and intuitive understanding of pollutant transportation at a large spatial scale, helping governments take timely interventions and formulate effective policies to mitigate air pollution.

Propagation pattern mining has been extensively investigated for spatio-temporal data. For example, Hoang et al.~\cite{nguyen2016discovering} focused on identifying representative congestion propagation patterns from traffic data, while Xiong et al.~\cite{xiong2018predicting} attempted to predict the future footprint of congestion propagation. However, these methods are not directly applicable to air pollution due to the data heterogeneity. Li et al.~\cite{li2017discovering} devised an efficient algorithm for mining propagation patterns based on the air quality data obtained from monitoring sensors. This algorithm first constructs causality graphs by quantifying the causal relationships between geo-distributed sensors, and then utilizes frequent subgraph mining~\cite{yan2002gspan} to retrieve propagation patterns from these causality graphs. Nevertheless, this study primarily focuses on the propagation patterns of PM$_{2.5}$, neglecting the interactions among different pollutants and the impact of various external factors. To this end, Zhu et al.~\cite{zhu2017pg} proposed pg-Causality, which combines frequent pattern mining with Bayesian learning, to discover influence pathways between multiple pollutants under different meteorological contexts. Using existing pattern mining techniques, pg-Causality first extracts frequent evolving patterns from air quality data. Subsequently, a Bayesian-based graphical model is trained to identify the causal relationships using the extracted patterns, where meteorological factors are also incorporated into the model to minimize result biases. However, propagation patterns could be more complex in real-world scenarios, involving fine-grained interactions (\eg district-level), uncertainties, and cascading processes.

To address the issues previously mentioned, Deng et al.~\cite{deng2019airvis} proposed an analytics system that combines frequent subgraph mining with interactive visualizations. This system empowers domain experts to explore and interpret the uncertainties of pollution propagation patterns across different districts. Furthermore, Deng et al.~\cite{deng2021visual} extended their work by integrating a cascading network inference model to capture the cascading pollution patterns that involve multiple locations.

\subsection{Spatial Co-Evolving Patterns}
This research branch focuses on identifying a group of spatially correlated sensors that share similar behaviors in their readings. The co-evolving patterns have significant implications for the study of propagation pattern discovery~\cite{zhu2017pg}, pollution inference and air quality forecasting~\cite{cheng2016finding}. Zhang et al.~\cite{zhang2015assembler} proposed a two-stage approach to discover co-evolving patterns in air quality data. In the first stage, a wavelet transform algorithm is employed to extract evolving intervals for each individual sensor. Then a non-parametric clustering approach is utilized to detect frequent evolutions by splitting and grouping the extracted evolving intervals. In the second stage, co-evolving patterns are generated by merging the detected frequent evolutions from individual sensors. To accelerate the generation process and reduce the search space, all the co-evolving patterns are organized into a tree structure according to the spatial constraint. Cheng et al.~\cite{cheng2016finding} further extend the research~\cite{zhang2015assembler} to discover dynamic co-evolving zones by using a time series clustering approach, based on air quality data collected from densely deployed sensors. Interestingly, incorporating the information of the discovered patterns significantly improves the performance of air quality forecasting, providing further evidence of the practical utility of co-evolving pattern mining.

\section{Air Quality Inference}
Due to economic concerns, such as high installation cost of sensors, we cannot conduct exhaustive air quality measurement across the whole urban space, \ie only a limited number of monitoring stations can be deployed in a city~\cite{zheng2013u,wang2019real}. However, since urban air quality depends on various complex factors (\eg land use, local emissions, human activities) non-linearly and varies by location, even two spatially close regions may have very different observations~\cite{zheng2013u}. Thus, obtaining real-time fine-grained air quality information is of great importance to citizens' decision-making and governments' policy formulation. To fuse the gap, in the past decade, numerous ML-based inference approaches~\cite{qi2018deep,cheng2018neural,han2021fine} have been proposed to reconstruct the entire pollution map based on sparse observations and a set of data sources collected in the city, such as weather, road network topology, traffic flow, and human mobility.

The air quality inference problem is formally defined as follows: suppose we have a set of predefined locations $\{l_{1}, l_{2}, \cdots, l_{N}\}$ in a city, each location $l_{i}$ is associated with a set of contextual features $\mathbf{c}_i$ (\emph{e.g.} traffic volume, building density), and has a pollution value to be inferred or already exists a pollution value $x_i$ if having a sensor measurement at time $t$. The goal is to infer the air quality for all the unmonitored locations at time $t$ by minimizing the following loss function
\begin{equation}
\mathcal{L}=\frac{1}{N} \sum_{i=1}^{N} \mathcal{L}_s(\hat{x_i},x_i),
\end{equation}
where $\hat{x_i}$ denotes the inferred value of location $l_{i}$ at time $t$, $\mathcal{L}_s(\cdot,\cdot)$ is an error function depending on the specific inference task, such as cross-entropy loss for classification~\cite{zheng2013u} or mean absolute error for regression~\cite{cheng2018neural}.
To solve the air quality inference problem, a plethora of approaches have been explored over the past decade. As illustrated in Figure~\ref{fig:infer}, we classify the techniques used in existing literature into four categories: spatial regression methods, sparse and low-rank methods, semi-supervised learning methods, and deep learning methods. Next, we will introduce representative algorithms from each of these categories.

\begin{figure}[t]
	\centering
	\includegraphics[width=13cm]{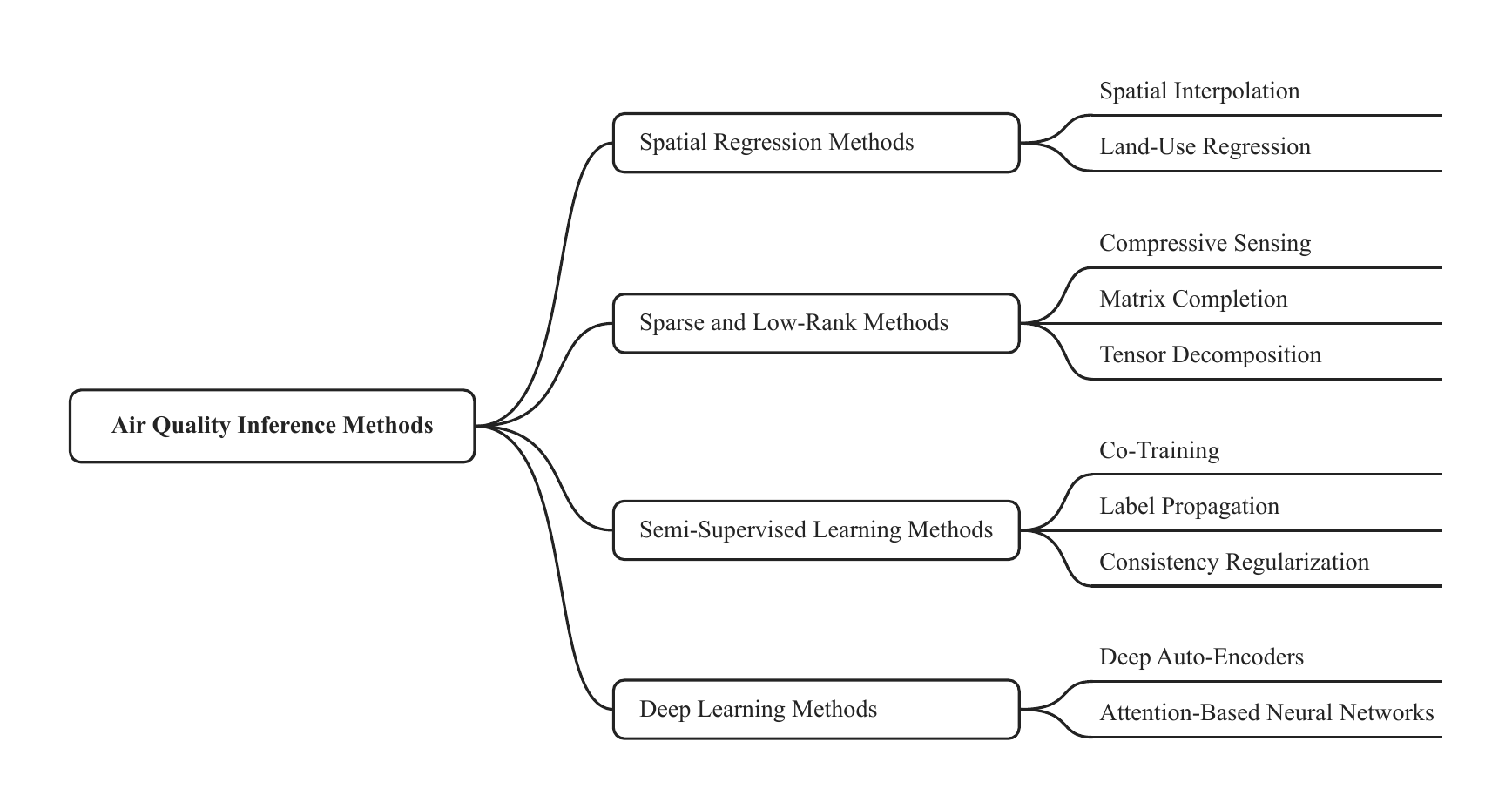}
	\caption{Categorizations of air quality inference methods.}
	\label{fig:infer}
\end{figure}

\subsection{Spatial Regression Methods}
According to the First Law of Geography~\cite{miller2004tobler}, the air quality of spatially adjacent locations tends to have similar values. Spatial regression~\cite{cressie2015statistics} explicitly incorporates such spatial correlations into the statistical regression, which has been widely adopted in air quality inference tasks. In this section, we discuss two categories of spatial regression methods: spatial interpolation and land-use regression.

\subsubsection{Spatial interpolation}
Spatial interpolation estimates values for unknown locations based on the locations with known values. For instance, to derive a real-time air pollution map throughout a city, spatial interpolation can infer the air quality of unmonitored locations by using observed values from other locations with deployed sensors. Formally, the general spatial interpolation formulation can be written as follows
\begin{equation}
\hat{x_i}=\sum_{j=1}^{M} w_{j} x_{j},
\end{equation}
where $\hat{x_i}$ is the estimated pollution value of unknown location $l_i$, $M$ is the number of locations with real-time air quality measurements, $w_i$ denotes the weight in the estimation process. As one of the most representative spatial interpolation approaches, inverse distance weighting (IDW)~\cite{wong2004comparison} aggregates the observed values by using distance decay coefficients $\hat{x_i}=\sum_{i=1}^{M} \frac{1}{d_i^p} x_{j}$, where $d_j$ is the distance between location $l_i$ and $l_j$, $p$ is a fixed real number. Another widely used interpolation method, ordinary kriging~\cite{li2011review}, learns weights $w_i$ by minimizing the variance of estimation error. The major issue on interpolation methods is that they often suffer from severe performance degradation when the air quality varies non-linearly in geographical space. A comprehensive survey on spatial interpolation can be found in Li et al.~\cite{li2014spatial}.

\subsubsection{Land-use regression}
Land-Use Regression (LUR) models infer air quality of unknown locations by learning the relationship between air quality and a set of land-use variables, such as the types of land use, population density, and terrain elevation~\cite{hoek2008review}.
To achieve this goal, the first step is to identify the relationship between observed air quality and these variables by training a linear regression model. Then the trained model is employed to estimate air quality for those unmonitored locations with available land-use data. Compared to spatial interpolation methods, LUR models further incorporate external factors associated with air pollution and demonstrate superior performance.

Hasenfratz et al.~\cite{hasenfratz2014pushing,hasenfratz2015deriving} proposed a non-linear LUR model for air pollution map derivation based on sparse measurements collected from public transport vehicles. Specifically, they partition region of interest into a number of grid cells (100m $\times$ 100m), and each grid cell encompasses a set of land-use variables. For a certain time period, the pollution measurements are assigned to their corresponding grid cells as labels. Then the pollution labels of unmonitored locations can be inferred via the following Generalized Additive Models (GAMs)
\begin{equation}
\ln (\hat{x_j})=a+f_{1}(\mathbf{e}_{1})+f_{2}(\mathbf{e}_{2})+\cdots+f_{d}(\mathbf{e}_{d})+\epsilon,
\end{equation}
where $f_{1},f_{2}\cdots f_{d}$ denote smooth regression splines, $\mathbf{e}_{1},\mathbf{e}_{2}\cdots \mathbf{e}_{d}$ are a set of land-use variables extracted from location $l_j$, $a$ and $\epsilon$ are intercept and error term, respectively. In practice, since land-use variables usually remain stable during a long time period, it is difficult to obtain estimated results with high temporal resolutions. To overcome such limitation, they train independent LUR models for each specific time intervals, yielding 989 models in total. Additionally, past measurements are also introduced into the LUR models, which substantially reduces the inference error. Building upon the aforementioned algorithm, Jutzeler et al.~\cite{jutzeler2014region} introduced an improved version that utilizes a Gaussian process model for inferring air quality. Moreover, the authors split the city into irregular regions with homogeneous emission based on the road network, which is more reasonable than grid-based partitioning. Recent years various machine learning approaches have been proposed based on land-use regression for more accurate air quality inference~\cite{lin2017mining,wang2019real,patel2022accurate}. For example, Lin et al.~\cite{lin2017mining} adopted a random forecast model to produce PM2.5 concentration of locations without monitoring stations based on publicly available OpenStreetMap data. Wang et al.~\cite{wang2019real} proposed a Gaussian process regression method for air pollution inference in a mobile sensing system. More recently, Patel et al.~\cite{patel2022accurate} extended a non-stationary Gaussian process based approach by further incorporating two kernels, including a Hamming distance-based kernel to encode categorical features and a local periodic kernel to model temporal periodicity. 

\subsection{Sparse and Low-Rank Methods}
This line of research aims to reconstruct a given high-dimensional object (\eg a vector, a matrix, or a tensor) from limited measurements by exploiting the sparse or low-rank nature of data~\cite{wang2018sparse}. 
For example, if we put air quality data into a matrix where each entry represents the measurement at a particular time and location, the matrix is incomplete as we only have a few locations deployed with monitoring sensors. The sparse and low-rank methods can be naturally utilized to fill in the missing part of such matrices. We hereafter discuss three sparse and low-rank learning techniques that can be applied to air quality inference problem: compressive sensing, matrix completion, and tensor decomposition.

\subsubsection{Compressive sensing}
Compressive sensing aims to recover a $n$-dimensional signal vector through $m$ sparse real measurements~\cite{baraniuk2010model}, where $m \ll n$. The basic assumption of compressive sensing is that the underlying signal is sparse in some basis, such as wavelets.
Here, we introduce the compressive sensing algorithm with a concrete example.

The example~\cite{wu2020sharing} was introduced in Section 2.2.2 as an illustration of a real-world application of crowd sensing. In this study, a fine-grained pollution map is inferred using calibrated air quality observations obtained from sensors deployed on shared bikes. Specifically, the authors partition the city into $n$ disjoint grid cells, and map the sampled air quality measurements into the corresponding cell. At a specific time interval, an observation vector $\mathbf{x}$ was built, where each entry in $\mathbf{x} \in \mathbb{R}^{m \times 1}$ stands for the air quality of $m$ observed grid cells. Afterwards, a Bayesian compressive sensing based inference model is proposed, taking the observation vector $\mathbf{x}$ as input, to reconstruct the pollution values for grid cells that are not covered by mobile sensors.

Specifically, suppose $\mathbf{x}_r \in \mathbb{R}^{n \times 1}$ represents the real pollution distribution vector of $n$ grid cells, $\mathbf{e}$ is a zero-mean Gaussian noise in the measurements, the pollution map reconstruction is defined as a linear regression problem
\begin{equation}
\mathbf{x}=\Phi \mathbf{x}_r+\mathbf{e}=\Theta \mathbf{w}+\mathbf{e},
\end{equation}
where $\Theta=\Phi \mathbf{B}$ is a mapping matrix, $\Phi \in \mathbb{R}^{m \times n}$ and $\mathbf{B} \in \mathbb{R}^{n \times n}$ are sampling matrix and basis matrix, and $\mathbf{w}$ denotes sparse weights to be estimated.

In practice, each row of $\Phi$ is a unit vector that only has one non-zero entry, indicating which grid cell one real air quality measurement belongs to. The matrix $\mathbf{B}$ is instantiated by a Gaussian kernel basis matrix $[\Psi(\mathbf{x}_{r 1}) \Psi(\mathbf{x}_{r 2}) \cdots \Psi(\mathbf{x}_{r n})]^{T}$, where $\Psi(\mathbf{x}_{r i})=[K(\mathbf{x}_{r i}, \mathbf{x}_{r 1}) \cdots \mathbf{x}(\mathbf{x}_{r i}, \mathbf{x}_{r n})]$, $\mathbf{x}_{r i}$ is the $i$-th grid cell, and $K(\mathbf{x}_{r i}, \mathbf{x}_{r j})$ denotes a geographical distance based Gaussian kernel function. Given $\mathbf{x}$ and $\Theta$, the authors utilize a fast sparse Bayesian learning method~\cite{ji2008bayesian,tipping2003fast} to derive the sparse vector $\mathbf{w}$ by maximizing the posterior probability. Finally, the algorithm recovers the pollution distribution over the whole space of interest via $\mathbf{x}_r=\mathbf{B} \mathbf{w}$.

\subsubsection{Matrix completion}
Matrix completion recovers the missing entries of a partially observed matrix $\mathbf{X}$ by decomposing it into the production of multiple low-rank matrices. Representative matrix completion approaches include eigen decomposition, singular value decomposition (SVD), and Simon Funk SVD.
Typically, a matrix $\mathbf{X}$ can be decomposed into multiple low-rank matrices using linear algebra techniques, such as $\mathbf{X}=\mathbf{U}\mathbf{V}^\top$, provided the matrix is complete. However, the matrix is usually incomplete and highly sparse in many application scenarios. Consequently, a general method to matrix completion is through solving the following optimization problem
\begin{equation}
\mathcal{L}=\|\mathbf{X}-\hat{\mathbf{X}}\|^{2}=\|\mathbf{X}-\mathbf{U} \mathbf{V}^{\top}\|^{2},
\end{equation}
where $\hat{\mathbf{X}}$ is an approximation of the original matrix $\mathbf{X}$. By treating $\mathbf{U}$ and $\mathbf{V}$ as trainable parameters, the objective of matrix completion is to minimize the error between the existing entries in $\mathbf{X}$ and their approximated values in $\hat{\mathbf{X}}$ using gradient descent. This allows us to fill in the missing entries of $\mathbf{X}$ by using the corresponding value in $\hat{\mathbf{X}}$.

Shang et al.~\cite{shang2014inferring} proposed a matrix completion approach to estimate real-time pollution emissions on each road segment using traffic speed and volume learned from taxi trajectories. Specifically, the authors first calculate the traffic speed from raw GPS trajectories and then construct an observation matrix $\mathbf{M}_{r}^{\prime}$, where each entry stands for the traffic speed on a particular road segment and at a specific time interval. However,  since only a small portion of road segments are covered by taxi trajectories during a certain time slot, there exists a substantial number of missing entries in $\mathbf{M}_{r}^{\prime}$. The highly sparse traffic observations may undermine the effectiveness of the above matrix completion algorithm.

To alleviate the data scarcity issue, the authors constructed three matrices $\mathbf{X}=\mathbf{M}_{r}^{\prime}||\mathbf{M}_{r}$, $\mathbf{Y}$, and $\mathbf{Z}$, where 
$\mathbf{M}_{r}$, $\mathbf{Y}$, and $\mathbf{Z}$ consists of rich contextual information of roads, namely historical periodic traffic speed patterns, POI distribution, and road network features. In comparison to $\mathbf{M}_{r}^{\prime}$, the above proposed matrices are much denser. With the help of these auxiliary matrices, $\mathbf{X}$ can be decomposed as follows
\begin{equation}
\mathbf{X} \approx \mathbf{T} \times(\mathbf{R}; \mathbf{R})^{\mathbf{T}};
\mathbf{Y} \approx \mathbf{T} \times(\mathbf{G}; \mathbf{G})^{\mathbf{T}};  \mathbf{Z} \approx \mathbf{R} \times \mathbf{F}^{\mathbf{T}},
\end{equation}
where $\mathbf{T}$, $\mathbf{G}$, $\mathbf{R}$, and $\mathbf{F}$ are decomposed low-rank matrices. As can be seen, $\mathbf{X}$ and $\mathbf{Y}$ share the same low-rank matrix $\mathbf{T}$, and $\mathbf{X}$ and $\mathbf{Z}$ share the same low-rank matrix $\mathbf{R}$. As a result, during the decomposition process, $\mathbf{X}$ can automatically absorb knowledge from context matrices $\mathbf{Y}$ and $\mathbf{Z}$ as complementary information, thereby improving the inference accuracy. The context-aware matrix completion framework can be trained via the following objective
\begin{equation}
\begin{aligned}
&\mathcal{L}(\mathbf{T}, \mathbf{R}, \mathbf{G}, \mathbf{F})=\|\mathbf{Y}-\mathbf{T}(\mathbf{G}; \mathbf{G})^{\mathbf{T}}\|^{2}+\lambda_{1} \|\mathbf{X}-\mathbf{T}(\mathbf{R}; \mathbf{R})^{\mathbf{T}}\|^{2}+ \\
&\lambda_{2}\|\mathbf{Z}-\mathbf{R} \mathbf{F}^{\mathbf{T}}\|^{2}+\lambda_{3}(\|\mathbf{T}\|^{2}+\|\mathbf{R}\|^{2}+\|\mathbf{G}\|^{2}+\|\mathbf{F}\|^{2}),
\end{aligned}
\end{equation}
where $\lambda_1$, $\lambda_2$, and $\lambda_3$ are hyper-parameters controlling the contributions of different model part. After obtaining the reconstructed traffic speed information, the authors further devise an unsupervised Bayesian network to infer the real-time traffic flow. Finally, the pollution emissions are computed by using an empirical equation from environmental theory based on the inferred traffic conditions.

\subsubsection{Tensor decomposition}
Tensor decomposition extends matrix completion techniques to higher-order tensors. There are various tensor decomposition methods in the existing literature, such as CP decomposition and Tucker decomposition. CP decomposition denotes a tensor as a sum of outer products of vectors, while Tucker decomposition decomposes the target tensor into a compact core tensor and a collection of low-rank matrices. For instance, considering a three-dimensional tensor $\mathbf{X} \in \mathbb{R}^{n_1 \times n_2 \times n_3}$, the general formula for Tucker decomposition can be expressed as follows
\begin{equation}
\mathbf{X} \approx \mathbf{Q} \times_{1} \mathbf{U} \times_{2} \mathbf{V} \times_{3} \mathbf{W},
\end{equation}
where $\mathbf{Q} \in \mathbb{R}^{r_1 \times r_2 \times r_3}$ denotes the core tensor, $\mathbf{U} \in \mathbb{R}^{n_1 \times r_1}$, $\mathbf{V} \in \mathbb{R}^{n_2 \times r_2}$, and $\mathbf{W} \in \mathbb{R}^{n_3 \times r_3}$ are three low-rank matrices, and $\times_{n}$ denotes the mode-$n$ product. Using the tensor decomposition technique, we can impute the missing entries of a sparse tensor by performing the multiplication of the decomposed tensors and matrices. Similar to the objective of matrix completion discussed in Equation 5, we can achieve this goal by optimizing the approximation error via gradient descent. 

Based on the theory of Tucker decomposition, Xu et al.~\cite{xu2016remote,xu2019fine} proposed a context-aware tensor decomposition approach to infer the fine-grained air pollution, utilizing a similar idea of context-aware matrix completion described in Section 3.3.2. Based on the ground and satellite remote sensing data measurements, the authors start by formulating a three-dimensional tensor $\mathbf{X}$ to represent air quality observations, with dimensions corresponding to urban regions, pollutant types (\eg PM$_{2.5}$, PM$_{10}$), and time intervals. However, approximately 86\% of the values in this tensor remain unknown due to the influence of clouds on remote sensing data. To tackle data scarcity issue, the authors built three context matrices $\mathbf{R}$, $\mathbf{S}$, and $\mathbf{T}$ to help the decomposition of $\mathbf{X}$ with various external information. Specifically, matrices $\mathbf{R}$ and $\mathbf{S}$ are constructed based on POI and meteorological data respectively, while matrix $\mathbf{T}$ is a graph adjacency matrix indicating the correlations between different air pollutants. The above matrices serve as complementary information to minimize the following loss function during decomposition process

\begin{equation}
\begin{aligned}
\mathcal{L}(\mathbf{Q}, \mathbf{U}, \mathbf{V}, \mathbf{W}, \mathbf{G})=& \|\mathbf{X}-\mathbf{Q} \times_{1} \mathbf{U} \times_{2} \mathbf{V} \times_{3} \mathbf{W}\|^{2}+\lambda_{1} \|\mathbf{R}-\mathbf{U} \mathbf{G}\|^{2}+\lambda_{2} \operatorname{tr}(\mathbf{V}^{\top} \mathbf{L} \mathbf{V}) \\
&+\lambda_{3}\|\mathbf{S}-\mathbf{W} \mathbf{U}^{\top}\|^{2}+\lambda_{4}(\|\mathbf{U}\|^{2}+\|\mathbf{V}\|^{2}+\|\mathbf{W}\|^{2}+\|\mathbf{G}\|^{2}),
\end{aligned}
\end{equation}
where $\operatorname{tr}(\cdot)$ is the matrix trace,  $\mathbf{L}=\mathbf{D}-\mathbf{T}$ denotes the Laplacian matrix of pollutant type correlation graph, $\mathbf{D}$ is the diagonal matrix of adjacency matrix $\mathbf{T}$, $\lambda_1$, $\lambda_2$, $\lambda_3$, and $\lambda_4$ are hyper-parameters controlling the importance of different parts in the decomposition. From the objective function, we can see that tensor $\mathbf{X}$ share a part of information with matrices $\mathbf{R}$ and $\mathbf{S}$, which can facilitate the knowledge in POI and meteorological data transferring into $\mathbf{X}$. In addition, the pollutant type correlations are also injected into $\mathbf{X}$ through the third term of Equation 9, which largely reduces the inference error.

\subsection{Semi-Supervised Learning Methods}
All the aforementioned methods rely on abundant supervised labels, \ie air quality observations, to achieve satisfactory performance. However, collecting a large annotated dataset is hard due to the inherent sparseness of measurement stations. To address this limitation, a common technique is to utilize sparsely labeled samples together with plentiful unlabeled data for semi-supervised learning (SSL)~\cite{van2020survey}. SSL takes advantage of the latent structure present in the unlabeled data to compensate for the lack of supervision, resulting in improved performance. Thus, researchers have proposed various specially designed SSL approaches for air pollution inference, including co-training, label propagation, and consistency regularization.

\subsubsection{Co-training}
The co-training algorithm~\cite{blum1998combining} is a well-known SSL approach that jointly trains separated models on two complementary views of the data. Zheng et al.~\cite{zheng2013u} proposed U-Air, a co-training based SSL framework to infer urban air pollution information based on air quality collected from measurement stations and heterogenous urban data. After splitting a target city into numerous disjoint grid cells, a cell is labeled with the observed air quality if it contains a monitoring station, and otherwise left unlabeled. The goal is to infer the air quality of unlabeled cells at the current time interval. Although spatial regression or low-rank learning methods can be directly employed for this problem, their performance is often unsatisfactory as we have many grids to infer while the labeled grids are extremely scarce.
To fully exploit the hidden information in unlabeled data, U-Air designed a co-training framework that comprises two mutually reinforced classifiers: the spatial classifier and the temporal classifier. Specifically, the spatial classifier utilizes an artificial neural network to capture the spatial correlation among grids, whereas the temporal classifier employs a linear-chain conditional random field model to capture the temporal dependency within a grid. The two classifiers are trained using feature sets extracted from two conditionally independent views and iteratively adopted to infer the air quality of unlabeled grids. Instances with the highest classified confidence are added to the labeled set for the subsequent round of training. This process continues until either the predefined round threshold is reached or the set of unlabeled grids becomes empty.

However, constructing such two conditionally independent feature sets is usually difficult in real-world applications. Moreover, the increase of pseudo-labeled examples in training set may introduce additional label noise. To address these issues, Chen et al.~\cite{chen2016spatially} proposed a framework called Semi-EP that consists of an ensemble SSL model and a pruning method. Likewise, Semi-EP is also a co-training style algorithm, which trains multiple classifiers simultaneously, but the training of different models is only based on a single view of data. During each round, different classifiers are trained on different labeled subsets generated using bootstrap sampling. In particular, to reduce the label noise, Semi-EP develops two tailored strategies. One is a confidence measurement strategy based on majority voting, which combines the results of multiple classifiers to improve the accuracy of confidence estimation. Another is a selection strategy based on the criteria used in tri-training~\cite{zhou2005tri} to filter incorrectly labeled samples. Additionally, Semi-EP further integrates an ensemble pruning process to enhance the diversity between classifiers.

\subsubsection{Label propagation}
Label propagation is a widely used SSL method that utilizes explicit graph structure to propagate label information between interconnected samples~\cite{song2022graph}. In particular, each sample is represented as a node on a graph, and the edge weight between nodes is determined based on some predefined criteria, such as physical distance or semantic similarity. The basic assumption behind label propagation is that nodes connected by edges are likely to share the same labels. Hence, labels can be naturally propagated across the graph to enhance the generalization performance of SSL tasks. For air quality inference applications, locations can be regarded as nodes on a graph, and relationships among nodes can be characterized by using various geographical features. By constructing such a graph, we can apply label propagation methods to estimate the air quality of unknown locations.

Take the model AQInf proposed in~\cite{hsieh2015inferring} as an example, we demonstrate how to leverage label propagation for urban air quality inference. Overall, AQInf is comprised of three major steps. The first step is to construct an affinity graph that describes the spatio-temporal correlations among locations. For instance, a location can be connected to its neighboring locations and the same location in previous time slots. The second step is ``weights learning'', which aims to automatically learn the edge weights of the affinity graph from data. The key idea is that nodes with higher feature similarity should have larger edge weights. In the final step, the authors employ a label propagation approach to infer the real-time air quality distribution for unmonitored locations by propagating the air quality information from labeled nodes to unlabeled nodes on the affinity graph. Since the observed air quality signals are very sparse on the graph, the authors propose to enrich the supervision signals by jointly optimizing the entropy and the KL-divergence between adjacent nodes, defined as
\begin{equation}
\mathcal{L}=-\sum_{i \in \mathcal{V}} \sum_{k \in \mathcal{K}} (\mathbf{p}_i(k) \log \mathbf{p}_i(k)+(1-\mathbf{p}_i(k)) \log (1-\mathbf{p}_i(k)))+\sum_{(i,j) \in \mathcal{E}} w_{i,j} D_{KL}(\mathbf{p}_i \| \mathbf{p}_j),
\end{equation}
where $\mathbf{p}_i$ is the inferred air quality distribution of node $i$, $\mathcal{K}=\{1,2,...,K\}$ is the set of $K$ possible air quality classes (\eg good, moderate, unhealthy),  $\mathcal{V}$ and $\mathcal{E}$ denotes the set of nodes and edges, $w_{i,j}$ is the edge weight between node $i$ and $j$. The model can be trained through back-propagation and gradient descent.

\subsubsection{Consistency regularization}
Consistency regularization exploits unlabeled data by injecting a regularization term that reflects the hidden data structure. The regularization term enforces the trained model to adhere to specific prior assumptions on the unlabeled data, such as cluster or smoothness assumptions. In recent years, consistency regularization has become increasingly popular in the development of deep SSL models, with representative works including ladder network~\cite{rasmus2015semi}, temporal ensembling~\cite{laine2016temporal}, and mean teacher~\cite{tarvainen2017mean}.

A common assumption used in spatiotemporal data is the smoothness assumption, which refers to nearby observations in both space and time that tend to be more similar than distant observations. Building upon this intuition, Zhao et al.~\cite{zhao2017incorporating} integrated a regularization strategy into a multi-task regression model for air quality inference. Specifically, they imposed spatial smoothness constraint on spatially adjacent monitoring stations and temporal smoothness constraint within each individual station so that their air quality is enforced to be similar. Despite the usefulness of such regularization strategy, the authors overlooked the spatiotemporal smoothness in unlabeled data, leading to inefficient data utilization under label-scarcity scenarios.
To deal with this issue, Qi et al.~\cite{qi2018deep} extended the smoothness assumption to unlabeled data and proposed a semi-supervised neural network. Similarly, the neural network is regularized by maintaining spatiotemporal smoothness of the inferred air quality during training phase, defined as follows,
\begin{equation}
\mathcal{L}_r=\frac{1}{N} \sum_{i=1}^{N} \sum_{j \in \mathcal{N}_{i}} e^{-d(i,j)} ||f_{\theta}(\mathbf{x}_i)-f_{\theta}(\mathbf{x}_j)||^2,
\end{equation}
where $N$ is the number of training set including both labeled and unlabeled samples, $\mathbf{x}_i$ is input features of sample $i$, $\mathcal{N}_{i}$ is spatial or temporal neighborhood of sample $i$, $f_{\theta}(\cdot)$ denotes a neural network that output the inferred air quality, and $d(\cdot)$ measures the spatio-temporal distance between adjacent sample $i$ and $j$. With this regularization term, the model can enrich supervision information by enhancing the spatiotemporal consistency of unlabeled data.


\subsection{Deep Learning Methods}
In recent years, deep learning has emerged as a rapidly advancing field with diverse applications in computer vision~\cite{lecun2015deep}, natural language processing~\cite{ouyang2022training}, and recommender systems~\cite{zhang2019deep}. Due to the powerful spatio-temporal representation capability~\cite{wang2020deep}, deep learning techniques have significantly improved the state-of-the-art in urban air quality inference. This progress primarily falls into two research lines: deep autoencoders and attention-based neural networks.

\subsubsection{Deep auto-encoders} 
Auto-encoder and its variants are commonly used techniques in many spatiotemporal inference tasks~\cite{qin2021network,liu2022unified}. The goal of auto-encoders is to reconstruct the input observations with the lowest error, by using an encoder-decoder architecture. 
In general, the formulation of auto-encoders can be defined as 
\begin{equation}
\begin{aligned}
\mathbf{h} & =f\left(\mathbf{x}^{i}; \theta_1\right), \\
\hat{\mathbf{x}}^{i} & =g\left(\mathbf{h}; \theta_2\right),
\end{aligned}
\end{equation}
where $f(\cdot)$ and $g(\cdot)$ are respectively encoder and decoder parameterized by $\theta_1$ and $\theta_2$, $\mathbf{h}$ is a latent representation vector, which can be then utilized to reconstruct the input observation $\mathbf{x}^{i}$. In typical auto-encoder architecture, the encoder and decoder can be instantiated with deep neural networks. We can train the auto-encoder by simply minimizing the objective $\mathcal{L}(\hat{\mathbf{x}}^{i},\mathbf{x}^{i})$, where $\mathcal{L}(\cdot)$ is a loss function measuring the difference between the input $\mathbf{x}^{i}$ and output $\hat{\mathbf{x}}^{i}$.

Ma et al.~\cite{ma2020fine} proposed a weather-aware deep auto-encoder framework to infer fine-grained air pollution, using sparse observations collected from a mobile sensing system. The principled idea is to learn a mapping function between incomplete pollution map and complete pollution map through auto-encoder, which is implemented by a convolution recurrent neural network~\cite{shi2015convolutional}. In the method, they first transform the irregularly sampled measurements into a three-dimensional tensor, where each entry in the matrix represents an exact pollution value in a particular location and in a certain time slot. Intuitively, the tensor is very sparse as we only have observations at a few locations and time slots. After that, an encoder is leveraged to convert the partially observed tensor into a low-dimensional condensed representation, and a decoder is employed to recover the entire tensor from this representation with the help of weather information. 

\subsubsection{Attention-based neural networks}
Many of the above methods either merely depend on contextual features (e.g., traffic, land-use) extracted from the target location, or employ random selection and k-nearest search to incorporate the information of spatial neighbors at the current time slot. However, since the spatio-temporal dependencies among different locations are highly dynamic and diverse, simply selecting random or nearest neighbors may not be reasonable in practice. On the one hand, the air quality of a location is not only influenced by its spatial neighborhood but also correlated with distant locations with similar environmental contexts. On the other hand, the dependencies between locations also change dynamically over time due to the effect of many complex factors, such as wind direction and traffic conditions. Hence, how to accurately quantify the complicated dependencies between different locations remains a challenge.

The emergence of attention mechanism~\cite{bahdanau2014neural} provides new opportunities for addressing the aforementioned issue. Specifically, when inferring the current air quality of a location, the attention mechanism enables the automatic selection of the most relevant information by learning the weights between different locations. For example, Cheng et al.~\cite{cheng2018neural} proposed a neural attention model, named ADAIN, to estimate the air quality at arbitrary locations within a given time slot using observed data from existing measurement stations. The inference process involves three steps. First, ADAIN leverages two parts of features as input, \ie static features (e.g., POIs, road network) and dynamic features (e.g., air quality, meteorology). The static features are fed into a feed-forward neural network (FNN), while the dynamic features are processed by a recurrent neural network (RNN). By combing the output of the FNN and RNN, a latent representation is obtained for each measurement station $s_i \in S$, denoted as $\mathbf{z}_{i}$. Second, ADAIN adopt the attention mechanism to learn the importance of different measurement stations to the target location $l$. More specifically, an attention score is calculated via a multi-layer perceptron (MLP)
\begin{equation}
e_{s_i,l}=\mathbf{w}_{2}^{\top} \sigma(\mathbf{W}_1(\mathbf{z}_{i} || \mathbf{z}_{l})+\mathbf{b}_{1})+b_{2},
\end{equation}
\begin{equation}
\alpha_{s_i,l}=\frac{\exp (e_{i})}{\sum_{s_i \in S} \exp (e_{i})},
\end{equation}
where $\mathbf{z}_{l}$ is the hidden representation of location $l$, attention score $\alpha_{s_i,l}$ measures the importance of station $s_i$'s features to target location $l$, $\mathbf{W}_1$, $\mathbf{b}_{1}$, $\mathbf{w}_2$ and $b_{2}$ are learned parameters of the MLP, $\sigma(\cdot)$ denotes activation function. Afterwards, a unified representation of location $i$ is computed based on the attention scores
\begin{equation}
\mathbf{z}^{\prime}_{l}=\mathbf{z}_{l}|| \sum_{i \in S} \alpha_{i} \mathbf{z}_{i}.
\end{equation}
Finally, the representation $\mathbf{z}^{\prime}_{l}$ are fed to another MLP model to generate the estimated air quality.

More recently, Han et al.~\cite{han2021fine} proposed a multi-channel attention model (MCAM) to explicitly decouple static and dynamic aspects of spatial correlations. MCAM is comprised of two channels: static channel and dynamic channel. In static channel, the target location $l$ and a set of measurement stations $S$ are regarded as nodes on a directed graph. Edges are built between the target location and measurement stations to indicate the time-invariant spatial influence. The attention mechanism is then utilized to quantify the importance of measurement stations on target location, based on the idea borrowed from bilateral filtering
\begin{equation}
e_{(s_{i}, l)}=\exp (-\frac{1}{2}(\frac{d(s_{i}, l)}{\sigma_{d}})^{2}) \exp (-\frac{1}{2}(\frac{\rho(s_{i}, l)}{\sigma_{r}})^{2}),
\end{equation}
where $d(s_{i}, l)$ and $\rho(s_{i}, l)$ measure the distance and Pearson correlation of static features (POIs, road network) between target location $l$ and measurement station $s_i \in S$, $\sigma_{d}$ and $\sigma_{r}$ are learnable parameters. 
In dynamic channel, a similar weighted graph is constructed, but the edge weights of this graph evolve over time. The authors employ a similar attention mechanism in~\cite{cheng2018neural} to compute time-varying weight of an edge in this graph by using air quality and meteorological features. Finally, a graph convolution network is utilized to further integrate static and dynamic channels for subsequent inference.


\section{Air Quality Forecasting}
Air quality forecasting is a long-standing fundamental research topic in environmental science~\cite{zhang2012real1,zhang2012real2}. Accurate and timely forecasting of future air quality states is essential for urban governance and human livelihood. In recent years, machine learning has emerged as the most popular technique in air quality forecasting area~\cite{zheng2015forecasting,yi2018deep,hao2021demand}. Compared to physics-informed numerical models~\cite{zhang2012real1,zhang2012real2}, machine learning models can directly learn useful knowledge from large historical data, which reduces the reliance on a great deal of computing powers and domain expertise. Here, we first formally define the air quality forecasting problem, and then provide a taxonomy to organize the existing ML-based predictive models.

Suppose we have a set of $N$ monitoring locations $\{l_1, l_2, \cdots, l_N\}$, each location $l_i$ is associated with a sequence of air quality observations $\mathbf{x}_i=(x^1_i,x^2_i,...,x^T_i)$, where $\mathbf{x}_i$ is a univariate time series and $x^t_i$ represents the observed value at time step $t$. We use $\mathbf{X}^{1:T}=(\mathbf{x}_1,\mathbf{x}_2,...,\mathbf{x}_N)$ to indicate the air quality measured in time interval $(1,T]$ for $N$ locations. Moreover, we denote $\mathbf{Z}^t=(\mathbf{z}^t_1, \mathbf{z}^t_2, \cdots, \mathbf{z}^t_N)$ as the covariate vectors associated to each location at time step $t$. Specifically, each vector $\mathbf{z}^t_i$ consists of both time-invariant (\emph{e.g.,} POI features) and time-varying features (\emph{e.g.,} meteorology or weather forecast). Based on the above notations, we define the air quality forecasting problem as follows: given a sequence of $T$ past observations $\mathbf{X}^{1:t}$, the goal is to predict air quality for all the locations over next $\tau$ time steps with the help of $\mathbf{Z}^{1:t+\tau}$,
\begin{equation}
\hat{\mathbf{X}}^{t+1:t+\tau}=f(\mathbf{X}^{1:t},\mathbf{Z}^{1:t+\tau};\theta),
\end{equation}
where $\theta$ are the parameters of a forecasting model, $\hat{\mathbf{X}}^{t+1:t+\tau}$ is the predicted air quality from future time $t+1$ to $t+\tau$. 

Based on the types of information utilized in prediction, the existing works can be categorized into temporal modeling methods, spatial modeling methods, and hybrid ensemble methods. Next, we elaborate on typical algorithms in each category and discuss their respective strengths and weaknesses in real-world scenarios.

\begin{figure}[t]
	\centering
	\includegraphics[width=13cm]{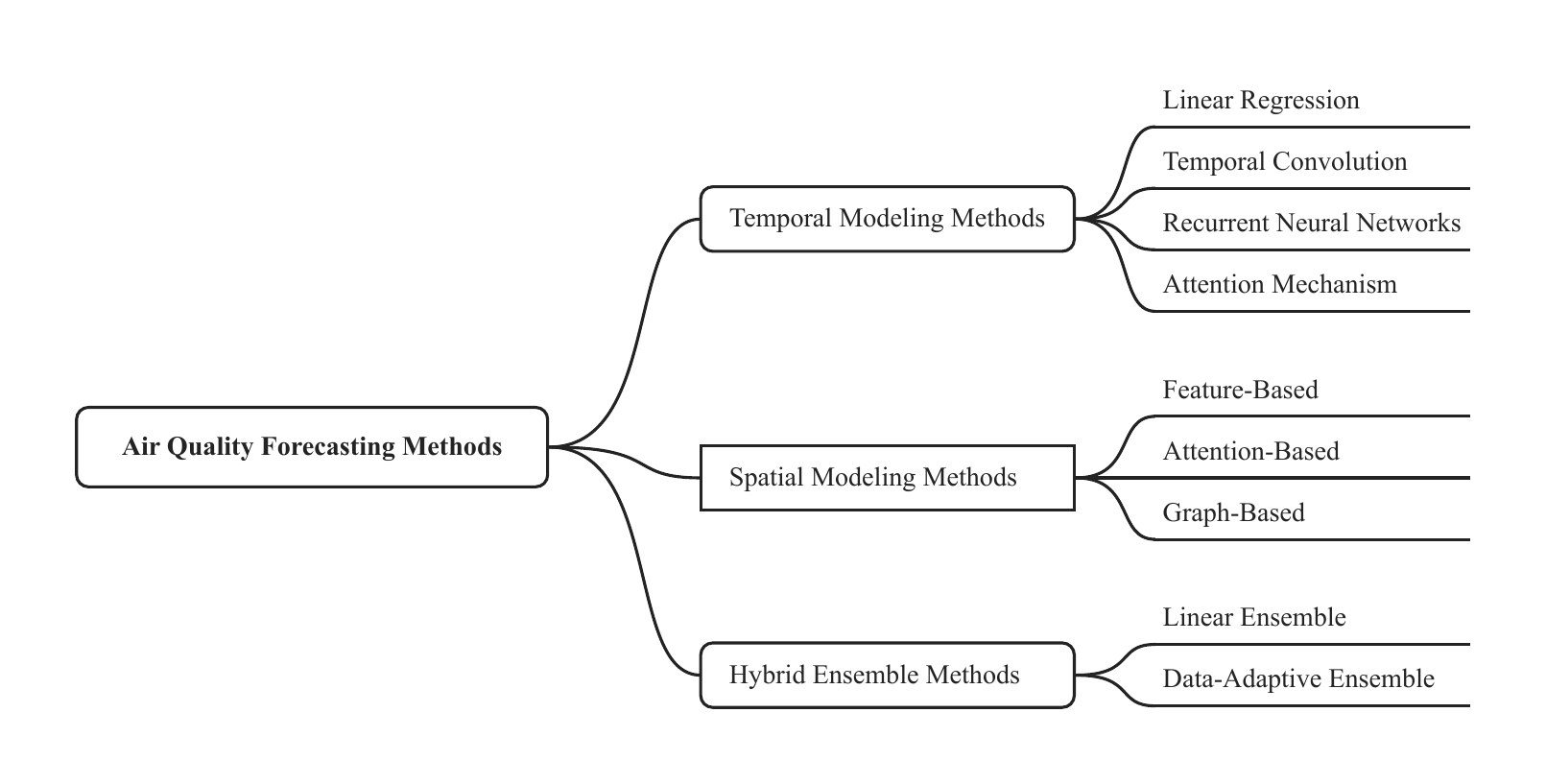}
	\caption{Categorizations of air quality forecasting methods.}
	\label{fig:pred}
\end{figure}

\subsection{Temporal Modeling Methods}
Future air quality signals yield complicated dependencies on
neighboring historical observations~\cite{zheng2015forecasting}, and taking them into account is beneficial for accurate prediction. This section discusses several machine learning approaches utilized for extracting temporal patterns in air quality prediction, including linear regression, temporal convolution, recurrent neural networks, and attention mechanism. Below we present representative methods for each category.

\subsubsection{Linear regression}
Linear regression directly maps historical time series into future prediction by using a weighted sum operation.
Zheng et al.~\cite{zheng2015forecasting} introduced a temporal approach for air quality forecasting, where air quality and meteorological data within a temporal sliding window as well as weather predictions are leveraged as input features. 
The method independently trains different linear regression predictors for forecasting horizon from $t+1$ to $t+\tau$, where the only difference among predictors are the input weather forecast features. The advantages are mainly two-fold. First, since the correlations between future air quality and past observations are diverse and vary from time to time, multiple predictors enable more accurate fitting of local data by employing independent parameter sets, which increases the overall model capacity. Second, it also helps to alleviate error accumulation effects induced by iterative prediction, thereby improving the overall performance. However, air quality usually exhibits complex non-linear and dynamic temporal correlations, which may diminish the effectiveness of linear regression models.

\subsubsection{Temporal convolution}
Temporal convolution~\cite{oord2016wavenet}, also known as dilated causal convolution, is a special class of convolution neural networks~\cite{lecun1989generalization} that are designed for time series data. Formally, given a sequence of air quality history $\mathbf{x}_i \in \mathbb{R}^T$ at location $l_i$, the temporal convolution at time step $t$ can be defined as
\begin{equation}
h_t=\sum_{s=0}^{S-1} f(s) \mathbf{x}_i(t-\omega \times s),
\end{equation}
where $h_t$ is the output of temporal convolution, $\omega$ is the dilation factor, $f(\cdot)$ is the learnable filter kernel with size $S$. Temporal convolution processes the input time series via a moving weighted average that only takes into account time points before $t$, which maintains the temporal order and causality of input data. Moreover, the dilation factor $w$ allows the filter $f(\cdot)$ to be applied over a larger receptive field by skipping input time series with every $w$ step. Hence, the receptive field grows exponentially with the increased number of convolution layers, which allows efficient modeling of long-term temporal dependencies. Due to its advantages, temporal convolution has been successfully used to extract temporal patterns for air quality prediction~\cite{wu2019msstn,du2019deep,samal2021temporal,liu2022fine}.

\subsubsection{Recurrent neural networks}
Recurrent Neural Networks (RNNs) occupy important positions in capturing inner relationships of time series data. The key idea of RNNs is to process sequential data by utilizing recurrent connections, allowing the flow of information from previous time steps to the current one. However, a common issue encountered during RNN model training is the vanishing and exploding gradients, which pose significant learning challenges. Several RNN-based architectures, such as Long Short-Term Memory (LSTM)~\cite{hochreiter1997long} and its simplified variant Gated Recurrent Unit (GRU)~\cite{cho2014properties}, are proposed to mitigate this issue by introducing a gating mechanism. 

Building upon the success of RNNs in sequence modeling~\cite{makin2020machine}, a vast number of methods have been developed for air quality forecasting. Most of these methods adopt the encoder-decoder architecture~\cite{sutskever2014sequence}, which first encodes the previous observations into a vector representation using the encoder and then generates predictions through the decoder, defined as
\begin{equation}
\begin{aligned}
\mathbf{h} & =f\left(\mathbf{X}^{1:t},\mathbf{Z}^{1:t+\tau}; \theta_1\right), \\
\hat{\mathbf{X}}^{t+1:t+\tau} & =g\left(\mathbf{h}; \theta_2\right),
\end{aligned}
\end{equation}
where $\mathbf{h}$ denotes the hidden vector representation, $\theta_1$ and $\theta_2$ are parameters of the encoder and decoder, $f(\cdot)$ and $g(\cdot)$ can be instantiated with RNNs.

Zhang et al.~\cite{zhang2019multi} introduced MGED-Net, a multi-group encoder-decoder model specifically designed to predict air quality in the next 24 hours. MGED-Net consists of two major modules: feature decoupling and feature fusion. The feature decoupling module automatically partitions the features into multiple groups based on Pearson correlations. Subsequently, an encoder-decoder fusion architecture is utilized to integrate the multi-group features for air quality prediction. In particular, each group of features is processed by using a specific LSTM model in the encoder. Another type of approach focuses on leveraging a probabilistic encoder and decoder to quantify the input and model uncertainty, thereby improving prediction reliability and interpretability. For example, Han et al.~\cite{han2020domain} developed a Bayesian RNN model for air pollution forecasting in China and the United Kingdom, defined as follows
\begin{equation}
\begin{aligned}
\mathbf{h} & \sim \pi_f\left(\mathbf{X}^{1:t},\mathbf{Z}^{1:t+\tau}; \theta_1\right) \\
\hat{\mathbf{X}}^{t+1:t+\tau} & \sim \pi_g\left(\mathbf{h} ; \theta_2\right),
\end{aligned}
\end{equation}
where $\mathbf{h}$ and $\hat{\mathbf{X}}^{t+1:t+\tau}$ are random variables. $\pi_f(\cdot)$ and $\pi_g(\cdot)$ represent Bayesian RNNs, which combine Bayesian inference and RNNs to enable uncertainty estimation and probabilistic modeling of urban air quality. In addition, the statistical relationship between pollutants is also integrated into the loss function as domain-specific knowledge, serving as a regularization of the model to improve the prediction accuracy. 

Despite fruitful progress, existing methods often generate predictions in an autoregressive manner, leading to error accumulation as early prediction errors may propagate over the forecast horizon~\cite{marisca2022learning}.
To address this issue, CGF~\cite{yu2023cgf} designed a training framework by introducing pollutant category guidance into encoder-decoder architecture. The pollutant concentration values are first converted into different categories according to predefined rules. For instance, pollution concentrations below 25 $\mu g / \mathrm{m}^3$ are classified as "Good", whereas the values ranging from 50 $\mu g / \mathrm{m}^3$ to 100 $\mu g / \mathrm{m}^3$ are labeled as "Poor". Based on the category information, CGF proposed two training strategies: category-based representation learning (CRL) and category-based self-paced learning (CSL). 
CRL incorporates category values as auxiliary supervision signals during the training process, leveraging their robustness and insensitivity to local fluctuations. By doing so, the model becomes more attentive to the general trends of air quality. CSL focuses on addressing the issue of error accumulation. If the current forecasting error is large, the real air quality value is used for the next-step prediction; otherwise, autoregressive results serve as input. To ensure effective learning, a self-paced learning strategy gradually reduces the proportion of real air quality values until a well-trained model is obtained. Additionally, category information is also incorporated into the self-paced learning strategy, as different categories may have distinct preferences when selecting large forecasting errors.

\subsubsection{Attention mechanism}
Due to gradient vanishing and limited memory capacity, the performance of RNN-based methods often deteriorates as the length of input sequence increases, making them less effective in handling long historical time series data~\cite{benidis2022deep}. One possible solution is to learn a hidden representation for each time step and determine which representations are crucial for generating accurate predictions. Attention mechanism~\cite{bahdanau2014neural} automatically learns a weighted sum of the representations using similarity measures like dot product or cosine similarity. This enables the model to selectively focus on the most relevant information at each time step, thus enhancing the model's capacity to capture long-range dependencies and extract informative features. For example, Chen et al~\cite{chen2020evanet} proposed EvaNet, an attention-based model for long-term air quality forecasting. EvaNet employs an encoder-decoder structure and integrates two attention mechanisms: extreme value attention and temporal attention. Specifically, the input features are first encoded by using an LSTM model to generate hidden representations at different time steps. After that, the extreme value attention is applied to select the representations based on extreme event features. Finally, a temporal attention is applied to capture long-range correlations and produce predictions by adaptively combining hidden representations in the decoder part. 

\subsection{Spatial Modeling Methods}
Besides temporal patterns, the air quality of a specific location also exhibits spatial correlations with its neighboring areas, as air pollutants can be transported and dispersed over long distances. Several methods have been developed to model the spatial correlations of air quality. In this section, we classify these existing methods into three categories: feature-based, attention-based, and graph-based approaches.

\subsubsection{Feature-based}
Feature-based approaches address the spatial correlation modeling problem by training a predictive model based on some hand-crafted features. In general, feature-based approaches assume that air quality observations at different locations are independent given the extracted features. The general formulation of feature-based methods is stated as follows
\begin{equation}
\hat{\mathbf{X}}^{t+1:t+\tau}_{i}=f(\Phi_{i}(\mathbf{X}^{1:t},\mathbf{Z}^{1:t+\tau});\theta),
\end{equation}
where $\Phi_{i}(\cdot)$ is a feature extractor that extracts spatial features at location $l_i$, $f(\cdot)$ denotes a regression model, $\hat{\mathbf{X}}^{t+1:t+\tau}_{i}$ is the model prediction at location $l_i$. In real-world scenarios, air quality measurements are irregularly and sparsely distributed in geographical space. Directly leveraging all these measurements as input features may cause a burden for model training and prediction, as these data are redundant and contain a large amount of noise. 

Zheng et al.~\cite{zheng2015forecasting} proposed a novel feature transformation method to reduce the input feature size based on spatial partition and aggregation. 
The method constructs a fixed-size set of spatial features for each target monitoring station. First, the geographical space is partitioned into 24 distinct regions by employing four lines and three concentric circles with different diameters. The innermost circle has the smallest diameter (e.g., 30 km), and the outermost circle has the largest diameter (e.g., 300 km). All three circles are centered around the target station. Afterwards, the existing stations within the outermost circle are mapped to the corresponding region based on their coordinates. Finally, the air quality and meteorological information observed from stations located within the same region are aggregated using a mean operator. Hence, each region contains a fixed size set of air quality and meteorological features, and regions without any stations are ignored. These features are then fed into a spatial predictor, namely an artificial neural network, for predicting future air quality. DeepAir~\cite{yi2018deep,yi2020predicting} improved the above feature transformation method by using a spatial interpolation method. For these regions without stations, some fake monitoring stations are randomly generated as reference points. Then a spatial interpolation method is applied to fill in the missing values of these reference points. After that, the interpolated values are aggregated to obtain the average features for the region. In addition, a deep neural network based method is further utilized to fuse the extracted features for urban air quality prediction. 

One major drawback of feature-based methods is that they depend on the quality of hand-crafted spatial features to obtain satisfactory performance. In practice, manually designing and selecting high-quality features heavily relies on extensive expert knowledge, which is a time-consuming and tedious process.

\subsubsection{Attention-based}
Attention-based methods have been extensively explored in the air quality forecasting area, where the attention weights are learned for sensor readings of monitoring stations to quantify the dynamic spatial correlations. Its generic framework is $\mathbf{h}^t_i=(\alpha_{1,i} x^t_1, \alpha_{2,i} x^t_2, \ldots, \alpha_{N,i} x^t_{N})$, where $\alpha_{j,i}$ represents the attention weight from station $l_j$ to $l_i$, and $x^t_i$ corresponds to the air quality reading of station $l_i$ at time step $t$. The weighted air quality readings $\mathbf{h}^t_i$ for station $l_i$ are then fed into a temporal attention module or an encoder-decoder architecture to further capture temporal dependencies and make predictions. 

Liang et al.~\cite{liang2018geoman} introduced a multi-level attention network called GeoMAN for air quality forecasting. GeoMAN consists of an encoder with local and global spatial attention, as well as a decoder with temporal attention. The local spatial attention focuses on the local correlations within a station, and employs an attention mechanism to capture complex relationships among observations of air quality and other pollutants, \eg SO$_2$ and NO. The global attention focuses on the correlations between stations, and leverages attention weights to adaptively select relevant stations by jointly considering the recent air quality, context information, and geographical distance. Finally, the output of spatial attention is fed into the temporal attention module to select relevant temporal information for prediction. 
Building upon the success of GeoMAN, Cheng et al.~\cite{cheng2021tip} further incorporated air flow trajectory information into spatial attention mechanism. 
The air flow trajectory represents the propagation patterns of air pollutants among different locations. By incorporating this information as prior knowledge, spatial attention can effectively track the path of pollution propagation, leading to more accurate air quality forecasting. For example, if an air flow trajectory is expected to arrive from the east of the target station in the upcoming hours, the model will focus more on monitoring stations along this trajectory. 

However, one shortcoming of attention-based methods is the quadratic complexity of attention weight calculation with the number of stations. This computational complexity makes it prohibitively expensive to capture spatial dependencies in large-scale air quality monitoring systems.
More recently, Liang et al.~\cite{liang2022airformer} proposed AirFormer, a method specifically designed to reduce the quadratic complexity of attention mechanisms for nationwide air quality prediction in China, which involves over 1,000 stations. AirFormer tackles this problem by dividing the surrounding area of each target station into 24 independent regions using the feature transformation module described in Section 5.2.1. The attention weights are then calculated only between the target station and its surrounding regions. Compared to the standard attention mechanism with quadratic cost, this operation reduces the computational complexity to linear \emph{w.r.t.} the number of stations, allowing for more efficient modeling of spatial dependencies in a large scale.

\subsubsection{Graph-based}
Graph neural networks (GNNs)~\cite{wu2020comprehensive} have emerged as a prominent research area in deep learning, exhibiting remarkable performance across a wide range of applications. In recent years, GNN-based approaches have demonstrated superior capabilities in various spatio-temporal forecasting tasks~\cite{jin2023spatio}. This can be attributed to their ability to capture spatial dependencies by propagating information on a predefined graph, which encodes the pair-wise relationships among spatial nodes~\cite{zhang2020semi}. In this section, we focus on GNN-based approaches pertaining to air quality forecasting. These approaches share a common routine that typically consists of two steps: (1) establish a non-Euclidean graph structure to represent the spatial relationships, and (2) perform spatio-temporal message passing based on the graph constructed in Step 1. The specific methodologies employed in each step are elaborated as follows.

\textbf{Step 1}: 
One fundamental problem for using GNNs in air quality forecasting is the graph construction, which directly decides the effectiveness of spatial correlation modeling. Here, we classify the graphs used in previous studies into three types: distance-based graphs, similarity-based graphs, and time-evolving graphs. 

Distance-based graphs capture spatial proximity by using geographical distance among stations~\cite{wu2019msstn,wang2020pm2,chen2021group,han2021joint,han2022semi,zhang2022multi}, defined as 
\begin{equation}
A_{i j}= \begin{cases} \kappa(\operatorname{dist}(l_i, l_j)) & \operatorname{dist}(l_i, l_j)<\epsilon \\ 0, & \text { otherwise }\end{cases},
\end{equation}
where $\operatorname{dist}(l_i, l_j)$ denotes the geographical distance between station $l_i$ and $l_j$, and the edge weight $A_{i j}$ is computed by a non-linear function $\kappa(\cdot)$ when the distance is lower than a predefined threshold $\epsilon$. One shortcoming of distance-based graphs is that they fail to consider the similarity between distant stations, \eg stations with similar environmental contexts. 

Similarity-based graphs address this issue by calculating either the functional similarities~\cite{lin2018exploiting,han2023kill} or similarities in observed time series~\cite{wang2021modeling}, defined as
\begin{equation}
A_{i j}= \begin{cases} \operatorname{sim}(l_i, l_j) & \operatorname{sim}(l_i, l_j)>\epsilon \\ 0, & \text { otherwise }\end{cases},
\end{equation}
where $\operatorname{sim}(\cdot)$ represents a similarity function. To capture functional similarities, a Euclidean distance-based function is commonly used to measure the similarity of node attributes, \eg POI distribution. In contrast, when calculating time series similarity, the Pearson Correlation Coefficient (PCC) is preferred. 

All of the above graphs are static in most cases. However, in real-world scenarios, the dependencies among stations can vary dynamically over time. For instance, the correlations between adjacent stations can change significantly due to variations in wind direction~\cite{liang2018geoman}. Therefore, using graphs with fixed edge weights in forecasting may introduce considerable bias. To this end, Li et al.~\cite{li2021ddgnet} proposed a strategy to learn a time-evolving graph structure by using geographical and recent air quality information. This method reduces reliance on predefined graphs and enables adaptive modeling of dynamic dependencies among stations.

\textbf{Step 2}: 
The goal of this step is to propagate and aggregate neighboring information based on the graph structure defined in Step 1. The core building block is a message passing layer, which can be formulated as follows:
\begin{equation}
\mathbf{h}_t^{i,(l+1)}=\phi^l(\mathbf{h}_{t}^{i,(l)},\underset{j \in \mathcal{N}(i)}{\operatorname{AGG}}\{\omega^l(\mathbf{h}_{t}^{i,(l)}, \mathbf{h}_{t}^{j,(l)}, A_{ij}\}),
\end{equation}
where $\mathbf{h}_{t}^{i,(l)}$ stands for the hidden representation of node $i$ at time step $t$ and layer $l$, $\mathcal{N}(i)$ is the set of spatial neighbors of node $i$, $A_{ij}$ indicates the edge weight connecting node $i$ and $j$, update function $\phi^l(\cdot)$ and message function $\omega^l(\cdot)$ can be instantiated with multi-layer perceptrons (MLPs), $\operatorname{AGG}\{\cdot\}$ represents an aggregation operator. 

Based on the terminology of previous work~\cite{cini2023taming}, we summarize existing GNN-based air quality forecasting methods into two categories: space-then-time methods and space-and-time methods. The space-then-time methods perform message passing before a temporal encoding neural network (TENN), while in space-and-time methods the dependencies along time and space are modeled in a more compact manner, \eg integrating message passing into RNNs. Specifically, the general formulation of space-then-time methods is defined as
\begin{equation}
\begin{aligned}
\mathbf{H}_t^{(l)} & =\operatorname{MP}^{(l)}(\mathbf{H}_t^{(l-1)},\mathbf{A}), \quad \forall l=1, \ldots, L \\
\mathbf{H}_t^{(L+1)} & =\operatorname{TENN}(\mathbf{H}_t^{(L)}), \\
\end{aligned}
\end{equation}
where $\mathbf{H}_t^{(l)}$ stands for the stack of node representations $\mathbf{h}_{t}^{i,(l)}$, $\operatorname{TENN}(\cdot)$ is a temporal encoding neural network implemented by RNNs~\cite{wang2020pm2,li2021ddgnet,han2021joint,han2022semi,han2023kill} or temporal convolution~\cite{wu2019msstn}, processing the output of message passing layer $\operatorname{MP}(\cdot)$ in a node-wise fashion. The major difference of existing methods lies in the aggregation functions $\operatorname{AGG}\{\cdot\}$ used in message passing. For example, Wu et al.~\cite{wu2019msstn} proposes to aggregate information along spatial dimension using a simple mean operator. Several recent studies~\cite{wang2020pm2,li2021ddgnet,han2021joint,han2023kill}  utilized attention-based aggregation for message passing, which largely enhances the representation capability. Moreover, Han et al.~\cite{han2022semi} proposed a hierarchical message passing layer, where the information can be propagated and aggregated based on the spatial hierarchy of urban regions to capture long-range spatial dependencies. 

Similarly, the space-and-time methods process input air quality data as follows
\begin{equation}
\mathbf{H}_t^{(l)}=\operatorname{TENN}^{(l)}(\operatorname{MP}^{(l)}(\mathbf{H}_t^{(l)},\mathbf{A})), \quad \forall l=1, \ldots, L.
\end{equation}
The operation in $\operatorname{TENN}(\cdot)$ is replaced by message passing layer, thus achieving the simultaneous modeling of spatio-temporal dependencies. Examples of air quality forecasting methods that fall into this category include GC-DCRNN~\cite{lin2018exploiting}, ATGCN~\cite{wang2021modeling} and the Group-Aware Graph Neural Network (GAGNN) introduced by Chen et al.~\cite{chen2021group}. Specifically, GC-DCRNN~\cite{lin2018exploiting} integrates the graph diffusion convolution~\cite{li2017diffusion} into gated recurrent units (GRUs), and perform message passing based on a functional similarity graph. ATGCN~\cite{wang2021modeling} first combines distance-based graph, functional similarity graph, and time series similarity graph and then leverages attention-based graph convolution coupled with GRUs for air quality prediction. In addition, GAGNN~\cite{chen2021group} is a hierarchical graph neural network for nationwide air quality forecasting. This method first automatically discovers the relationships between cities and city groups through a differentiable grouping network. Then message passing is adopted along the identified graph relationships to model spatial dependencies among cities and city groups. 

\subsection{Hybrid Ensemble Methods}
Hybrid ensemble approaches integrate multiple heterogeneous models to improve the predictive performance. Existing studies on air quality prediction mainly utilize two types of strategies to combine the results of these models: linear ensemble and data-adaptive ensemble.

\subsubsection{Linear ensemble}
Linear ensemble methods aggregate the predictions from different models by employing a weighted sum function.
For example, Luo et al.~\cite{luo2019accuair} propose a linear ensemble framework for air quality prediction in the next 48 hours. The framework consists of three distinct models: a gradient boosting decision tree (GBDT), a spatio-temporal gated deep neural network (DNN), and a GRU-based model. The GBDT model focuses on forecasting air quality at each station by utilizing a set of hand-crafted features, \eg recent weather conditions, air quality from the past few hours, and corresponding features from nearby stations. Instead, the spatio-temporal gated DNN directly captures the spatio-temporal correlations from raw air quality data for prediction. Moreover, a GRU-based model is utilized to further improve sequence prediction capability by incorporating weather forecast features at each forecast horizon, which largely enhances the accuracy of long-term and sudden change predictions. The three models generate their own prediction results independently, which are subsequently aggregated using a linear regressor. Mathematically, suppose the output of the three models are $\hat{x}_{gbdt}$, $\hat{x}_{dnn}$, and $\hat{x}_{seq}$, respectively, the linear ensemble is formally defined as:
\begin{equation}
\hat{x} = w_1 \hat{x}_{gbdt}+w_2 \hat{x}_{dnn}+w_3 \hat{x}_{seq},
\end{equation}
where $w_i$ is learnable weight assigned to each model. To ensure that the average prediction value remains consistent, $w_i$ is constrained to the range [0,1]. By integrating these three models into a unified framework, we can fully exploit their respective strengths, addressing the challenges posed by complicated spatial-temporal correlations, long-term prediction, and sudden change prediction.

\subsubsection{Data-adaptive ensemble}
According to empirical observations, it has been noted that the combination of models varies over time and locations in a dynamic manner~\cite{zheng2015forecasting}. Therefore, employing static-weight combination may introduce considerable bias. Data-adaptive ensemble methods calculate distinct weights for each data instance according to various contextual factors, such as weather conditions. By doing so, they can dynamically quantify the importance of each model to the final prediction~\cite{zheng2015methodologies}.

Along this line, Zheng et al.~\cite{zheng2015forecasting} present a hybrid framework for station-level air quality forecasting. The framework consists of two base predictive models, namely the temporal predictor and the spatial predictor, along with a prediction aggregator. The temporal predictor models and predicts the air quality of a station based on its own weather and air quality data, whereas spatial predictor captures the spatial correlations and predicts air quality by leveraging features (\eg, air quality, wind speed) extracted from spatial neighbors' data. The prediction aggregator dynamically combines the results of the two predictors based on current weather conditions. Specifically, the results of the two predictors are first paired with the corresponding weather features, and then a decision tree is constructed to partition the paired data into different groups, with the aim of minimizing the variance of data through weather features. Each group of data in a leaf node is fitted using different linear regression models. During the inference stage, the dynamic combination can be achieved by first selecting the specific regression model according to the current weather features and then generating a final prediction using the selected model.

\section{Public Datasets}
This section introduces the public air quality datasets that are available for analysis and research. We present the sources of these datasets, the variables they measure, and the extent of their spatial and temporal coverage.

\textbf{OpenSense}\footnote[1]{https://gitlab.ethz.ch/tec/public/opensense} is a public air pollution dataset from ETH Zurich OpenSense project~\cite{li2012sensing}, collected by sensors installed on top of 10 streetcars. The dataset consists of ultrafine particulate (UFP), ozone (O$_{3}$), and carbon monoxide (CO) measurements, covering a large urban area of 100 Km$^2$ and a long time period of 4 years. It has been used by Hasenfratz et al.~\cite{hasenfratz2014pushing} to investigate the task of inferring detailed pollution maps.

\textbf{Grid-PM2.5}\footnote[2]{https://drive.google.com/file/d/1hdiYt3scAI0hxlvuGLBYoZzlWjZFNPat/view} contains a half-year of PM$_{2.5}$ records generated by a mobile sensing system deployed in Binhai district in Tianjin, China. Each record includes a record ID, a car ID, a timestamp, a latitude, a longitude, and the corresponding PM$_{2.5}$ measurement. The dataset was used for fine-grained pollution map recovery~\cite{ma2020fine}.

\textbf{UrbanAir}\footnote[3]{http://urban-computing.com/data/Data-1.zip} dataset
records several air quality measurements, including PM$_{2.5}$, PM$_{10}$, NO$_{2}$, CO, O$_{3}$, and SO$_{2}$, from 437 monitoring stations across 43 Chinese cities. The measurements are collected hourly from May 1, 2014, to April 30, 2015. It also provides several meteorology observations such as weather conditions, temperature, humidity, pressure, wind speed and wind direction. This dataset is widely used in the tasks like air quality forecasting~\cite{wang2021modeling} and multivariate time
series imputation~\cite{cini2021filling,cao2018brits,yi2016st}.

\textbf{KDDCUP}\footnote[4]{https://www.biendata.xyz/competition/kdd\_2018} is a dataset provided by the air quality forecasting competition in KDD CUP 2018.
It contains hourly air quality measurements, \eg PM$_{2.5}$, PM$_{10}$, and SO$_{2}$, collected from 35 monitoring stations in Beijing and 13 monitoring stations in London, as well as meteorological observations such as weather conditions, temperature, humidity, pressure, wind speed and wind direction. The data covers the period from January 1, 2017, to March 31, 2018.

\textbf{AQ-India}\footnote[5]{https://www.kaggle.com/datasets/rohanrao/air-quality-data-in-india} dataset contains air pollution data and AQI at hourly and daily levels collected from various stations across 26 cities in India. The air pollutants measured include PM$_{2.5}$, PM$_{10}$, NO$_{2}$, CO, NH$_{3}$, SO$_{2}$, and AQI, covering the period from 2015 to 2020.

\textbf{KnowAir}\footnote[6]{https://github.com/shuowang-ai/PM2.5-GNN} dataset records PM$_{2.5}$ measurements from 1,498 monitoring stations across 184 Chinese cities, published by \cite{wang2020pm2}. All the measurements are collected hourly from January 1, 2015, to December 31, 2018.

\textbf{EPAAQS}\footnote[7]{https://www.epa.gov/outdoor-air-quality-data}~(Environmental Protection Agency Air Quality System) dataset is a collection of ambient air pollution data~(\eg O$_{3}$, PM$_{2.5}$, CO, SO$_{2}$, and NO$_{2}$) collected by Air Quality System of US Environmental Protection Agency from over thousands of measurement stations across the United States. The dataset also provides meteorological data, meta information of each measurement station, such as longitude, latitude and operator, as well as data quality control and assurance information. The EPAAQS dataset spans over 40 years, from 1980 to present.



\textbf{OpenAQ}\footnote[8]{https://openaq.org/} is one of the largest open-resource air quality data platforms in the world.  
It continually produces hourly pollutant measurements, \eg PM$_{2.5}$, PM$_{10}$, SO$_{2}$, CO, NO$_{2}$, O$_{3}$, by aggregating and harnessing data from over 8,000 monitoring stations in 65 countries since 2016.



\textbf{WAQI}\footnote[9]{https://aqicn.org/data-platform/register/}~(World Air Quality Index) dataset provides daily air quality observations, including measurements such as PM$_{2.5}$, PM$_{10}$, O$_{3}$, CO, NO$_{2}$, SO$_{2}$, for cities from over 100 countries around the world since 2014 to present. 

\textbf{CAMS}\footnote[10]{https://ads.atmosphere.copernicus.eu/cdsapp\#!/dataset/cams-europe-air-quality-reanalyses}~(Copernicus Atmosphere Monitoring Service) provides air quality reanalysis data in European area. The reanalysis integrate model simulation with in-situ measurements (\eg sensor, satellite) into gridded dataset via data assimilation techniques, producing hourly analyses of various pollutants (\eg PM$_{2.5}$, PM$_{10}$, and CO$_{2}$) at fine spatial granularity (0.1 degree, approximately 10Km).

\textbf{CityAir}\footnote[11]{https://drive.google.com/file/d/1I\_vpbLJhOJpNh-TpLdSWsaG3xCpzMVSQ/view} is a nationwide dataset containing AQI and weather data of 209 Chinese cities from January 1, 2017 to April 30, 2019. It was used to evaluate the performance of nationwide city-level air quality forecasting task in~\cite{chen2021group}.

\textbf{AirNet}\footnote[12]{https://github.com/usail-hkust/AirNet} consists of hourly pollution measurements (\eg PM$_{2.5}$, PM$_{10}$, and CO$_{2}$) from 2,028 monitoring stations distributed in 375 Chinese cities, which is provided by this paper. The dataset spans over 8 years, from January 1, 2015 to December 31, 2022. We hope this dataset can facilitate broad air quality analysis research and applications in the future.
\section{Open Challenges and Future Directions}
While machine learning has demonstrated remarkable success in urban air quality analytics, there still remain a number of open issues and promising research directions that require further investigation in the future. In this discussion, we aim to highlight the existing open challenges and potential future directions in the field of urban air quality analytics, with the intent to inspire researchers to push the boundaries and make significant advancements in this area.

\subsection{Learning from Multimodal Data}
In the era of big data, a huge volume of data reflecting air pollution have been generated and collected from a diversity of sources, such as satellite imagery, sensor networks, and social media feeds. These datasets encompass multiple modalities, each of which has a distinct data type, structure, distribution, and scale. For example, satellite imagery utilizes pixel intensities to characterize aerosol distribution, whereas air quality measurements from sensor networks can be represented as a set of geo-tagged time series. By extracting and fusing information from different modalities, we can improve spatial and temporal resolution, enhance the accuracy of predictive models, and uncover underlying pollution patterns that may not be discernible within a single modality. Therefore, it is imperative to explore tailored machine learning techniques that can effectively integrate knowledge from heterogeneous air quality data. Multimodal learning~\cite{baltruvsaitis2018multimodal} is the mainstream approach to harness complementary information between modalities, which has achieved remarkable success in computer vision and nature language processing tasks~\cite{radford2021learning,saharia2022photorealistic}. However, the application of multimodal learning in the field of air quality analytics is not well studied yet, which has the potential to become a promising and beneficial direction for future research.

\subsection{Incorporating Physical Laws}
In recent decades, environmental scientists have made significant advancements in the development of physics-based models, such as street canyon models and Gaussian plume models, which have provided valuable insights into air pollution modeling. These models aim to simplify the complex process of atmospheric dispersion into concise scientific principles to capture core mechanisms, but they overlook detailed information due to limited knowledge of specific processes. Machine learning models have demonstrated superior capability for the accurate fitting of large-scale historical data, but they may fail to produce physically consistent results and cannot accurately model the causality of various influential factors. Consequently, a promising direction is to integrate physics-based models and machine learning models to leverage their complementary strengths. Existing works~\cite{ma2020fine} have started to explore the relationship between physics-based models and deep learning models. Nevertheless, how to combine them in a unified way has not been well studied yet. Simply using the output of physics-based models as input features of machine learning models cannot achieve satisfactory performance~\cite{willard2020integrating}. As a result, more advanced integration techniques are required to incorporate the core physical principles and domain expertise into machine learning models.

\subsection{Quantifying Predictive Uncertainty}
Air pollution is influenced by a variety of factors, including meteorological conditions, pollutant emissions, and human activities, resulting in significant variability and uncertainty in the data. Although machine learning models are powerful in capturing complicated spatio-temporal patterns, they often provide point estimates without quantifying the associated uncertainty. Uncertainty quantification bridges this gap by generating confidence intervals for point estimates~\cite{wu2021quantifying,wang2019deep}. Existing methods on uncertainty quantification fall into two categories: one involves introducing perturbations to model parameters or the dataset during the inference stage~\cite{gal2016dropout,lakshminarayanan2017simple}, while the other focuses on modeling the posterior distribution of model parameters based on observed data~\cite{neal2012bayesian}. However, air quality data also pose unique challenges. For example, how to leverage spatio-temporal structure of data for uncertainty quantification? How to reduce error propagation of uncertainty quantification in multi-step air quality forecasting? Being able to provide probabilistic air quality estimation will be tremendously helpful for decision-making and environmental management. Unfortunately, the application of uncertainty quantification for air quality analysis has yet to be explored in the community.

\subsection{Model Interpretability}
Currently, numerous sophisticated machine learning models have been proposed to enhance predictive performance. However, despite their ability to improve prediction accuracy, the ever-increasing model complexity presents a huge obstacle for humans in interpreting the model results. Consequently, it has become a pressing issue to enhance models' interpretability. On the one hand, interpretability provide supporting evidence to decision makers regarding model results, thereby enhancing trust in the models. On the other hand, interpretability empowers researchers and data scientists to gain profound insights into the strengths and weaknesses of the models, which shed light on model design and optimization. Recent studies have explored causal effect learning to achieve causal-level interpretation in specific applications, such as event prediction~\cite{deng2022robust} and bike flow prediction~\cite{deng2023spatio}. Thus, a potential future direction is to extend causal effect learning to air quality analytics tasks. By explicitly modeling the causal relationships among diverse factors, such as meteorological conditions, pollutant emissions, and their impact on analysis, we can uncover the critical drivers behind air quality variations, identify salient spatio-temporal features, and enhance the transparency and acceptability of the models.

\subsection{From Analysis to Decision-Making}
At present, a vast amount of ML-related works have focused on analyzing and predicting air quality. However, there still remains a lack of literature connecting the analysis results with actionable decision-making strategies, which are crucial for mitigating pollution emissions. A recent study DeepThermal~\cite{zhan2022deepthermal} demonstrated the potential of reinforcement learning~\cite{zhang2021intelligent,fan2021interactive} in this domain. This work developed a reinforcement learning framework to optimize combustion control strategies in thermal power plants, reducing nitrogen oxides emissions by hundreds of tons annually. Hence, reinforcement learning could be a promising direction for enhancing decision-making processes in air pollution management. Another interesting problem is intervention-based air quality analysis. In real-world scenarios, it is essential to quantify the potential impact of modifications to an urban system on air quality. For example, how much will air pollution decrease if we relocate a factory away from the city? Will the reduction in pollution emissions be substantial enough if we electrify a certain share of vehicles? Addressing this kind of questions is vital for informed decision-making by stakeholders and governmental authorities. Recently, interventional causal machine learning~\cite{ahuja2023interventional} has emerged as an effective tool for analysis in the presence of external interventions, which could be a potential direction for future research.

\section{Conclusion}
This survey has provided a comprehensive overview of machine learning techniques developed specifically for urban air quality analytics. The article presents an extensive review of the challenges, research problems, methodologies, and future prospects in this field, summarizing significant research findings from more than 100 publications, with a focus on works published within the past five years. We provide a systematic taxonomy of existing analytical models based on research problems and associated techniques. Furthermore, an in-depth analysis is conducted to illustrate the relationships, differences, strengths, and weaknesses among these machine learning methods. This paper elucidates the application of machine learning in the field of air quality analytics, catering not only to researchers from computer science but also to a wide spectrum of communities like environmental science. Finally, a list of public air quality datasets is provided, along with a discussion of the open challenges and future directions in this emerging field.


\bibliography{ref}


\begin{thebibliography}{157}


\ifx \showCODEN    \undefined \def \showCODEN     #1{\unskip}     \fi
\ifx \showDOI      \undefined \def \showDOI       #1{#1}\fi
\ifx \showISBNx    \undefined \def \showISBNx     #1{\unskip}     \fi
\ifx \showISBNxiii \undefined \def \showISBNxiii  #1{\unskip}     \fi
\ifx \showISSN     \undefined \def \showISSN      #1{\unskip}     \fi
\ifx \showLCCN     \undefined \def \showLCCN      #1{\unskip}     \fi
\ifx \shownote     \undefined \def \shownote      #1{#1}          \fi
\ifx \showarticletitle \undefined \def \showarticletitle #1{#1}   \fi
\ifx \showURL      \undefined \def \showURL       {\relax}        \fi
\providecommand\bibfield[2]{#2}
\providecommand\bibinfo[2]{#2}
\providecommand\natexlab[1]{#1}
\providecommand\showeprint[2][]{arXiv:#2}

\bibitem[\protect\citeauthoryear{Ahuja, Mahajan, Wang, and Bengio}{Ahuja
  et~al\mbox{.}}{2023}]%
        {ahuja2023interventional}
\bibfield{author}{\bibinfo{person}{Kartik Ahuja}, \bibinfo{person}{Divyat
  Mahajan}, \bibinfo{person}{Yixin Wang}, {and} \bibinfo{person}{Yoshua
  Bengio}.} \bibinfo{year}{2023}\natexlab{}.
\newblock \showarticletitle{Interventional causal representation learning}. In
  \bibinfo{booktitle}{\emph{International Conference on Machine Learning}}.
  PMLR, \bibinfo{pages}{372--407}.
\newblock


\bibitem[\protect\citeauthoryear{Akbari, Samadzadegan, and Weibel}{Akbari
  et~al\mbox{.}}{2015}]%
        {akbari2015generic}
\bibfield{author}{\bibinfo{person}{Mohammad Akbari}, \bibinfo{person}{Farhad
  Samadzadegan}, {and} \bibinfo{person}{Robert Weibel}.}
  \bibinfo{year}{2015}\natexlab{}.
\newblock \showarticletitle{A generic regional spatio-temporal co-occurrence
  pattern mining model: a case study for air pollution}.
\newblock \bibinfo{journal}{\emph{Journal of Geographical Systems}}
  \bibinfo{volume}{17} (\bibinfo{year}{2015}), \bibinfo{pages}{249--274}.
\newblock


\bibitem[\protect\citeauthoryear{Bahdanau, Cho, and Bengio}{Bahdanau
  et~al\mbox{.}}{2014}]%
        {bahdanau2014neural}
\bibfield{author}{\bibinfo{person}{Dzmitry Bahdanau},
  \bibinfo{person}{Kyunghyun Cho}, {and} \bibinfo{person}{Yoshua Bengio}.}
  \bibinfo{year}{2014}\natexlab{}.
\newblock \showarticletitle{Neural machine translation by jointly learning to
  align and translate}.
\newblock \bibinfo{journal}{\emph{arXiv preprint arXiv:1409.0473}}
  (\bibinfo{year}{2014}).
\newblock


\bibitem[\protect\citeauthoryear{Baltru{\v{s}}aitis, Ahuja, and
  Morency}{Baltru{\v{s}}aitis et~al\mbox{.}}{2018}]%
        {baltruvsaitis2018multimodal}
\bibfield{author}{\bibinfo{person}{Tadas Baltru{\v{s}}aitis},
  \bibinfo{person}{Chaitanya Ahuja}, {and} \bibinfo{person}{Louis-Philippe
  Morency}.} \bibinfo{year}{2018}\natexlab{}.
\newblock \showarticletitle{Multimodal machine learning: A survey and
  taxonomy}.
\newblock \bibinfo{journal}{\emph{IEEE transactions on pattern analysis and
  machine intelligence}} \bibinfo{volume}{41}, \bibinfo{number}{2}
  (\bibinfo{year}{2018}), \bibinfo{pages}{423--443}.
\newblock


\bibitem[\protect\citeauthoryear{Baraniuk, Cevher, Duarte, and Hegde}{Baraniuk
  et~al\mbox{.}}{2010}]%
        {baraniuk2010model}
\bibfield{author}{\bibinfo{person}{Richard~G Baraniuk}, \bibinfo{person}{Volkan
  Cevher}, \bibinfo{person}{Marco~F Duarte}, {and} \bibinfo{person}{Chinmay
  Hegde}.} \bibinfo{year}{2010}\natexlab{}.
\newblock \showarticletitle{Model-based compressive sensing}.
\newblock \bibinfo{journal}{\emph{IEEE Transactions on information theory}}
  \bibinfo{volume}{56}, \bibinfo{number}{4} (\bibinfo{year}{2010}),
  \bibinfo{pages}{1982--2001}.
\newblock


\bibitem[\protect\citeauthoryear{Benidis, Rangapuram, Flunkert, Wang, Maddix,
  Turkmen, Gasthaus, Bohlke-Schneider, Salinas, Stella, et~al\mbox{.}}{Benidis
  et~al\mbox{.}}{2022}]%
        {benidis2022deep}
\bibfield{author}{\bibinfo{person}{Konstantinos Benidis},
  \bibinfo{person}{Syama~Sundar Rangapuram}, \bibinfo{person}{Valentin
  Flunkert}, \bibinfo{person}{Yuyang Wang}, \bibinfo{person}{Danielle Maddix},
  \bibinfo{person}{Caner Turkmen}, \bibinfo{person}{Jan Gasthaus},
  \bibinfo{person}{Michael Bohlke-Schneider}, \bibinfo{person}{David Salinas},
  \bibinfo{person}{Lorenzo Stella}, {et~al\mbox{.}}}
  \bibinfo{year}{2022}\natexlab{}.
\newblock \showarticletitle{Deep learning for time series forecasting: Tutorial
  and literature survey}.
\newblock \bibinfo{journal}{\emph{Comput. Surveys}} \bibinfo{volume}{55},
  \bibinfo{number}{6} (\bibinfo{year}{2022}), \bibinfo{pages}{1--36}.
\newblock


\bibitem[\protect\citeauthoryear{Bi, Xie, Zhang, Chen, Gu, and Tian}{Bi
  et~al\mbox{.}}{2022}]%
        {bi2022pangu}
\bibfield{author}{\bibinfo{person}{Kaifeng Bi}, \bibinfo{person}{Lingxi Xie},
  \bibinfo{person}{Hengheng Zhang}, \bibinfo{person}{Xin Chen},
  \bibinfo{person}{Xiaotao Gu}, {and} \bibinfo{person}{Qi Tian}.}
  \bibinfo{year}{2022}\natexlab{}.
\newblock \showarticletitle{Pangu-weather: A 3d high-resolution model for fast
  and accurate global weather forecast}.
\newblock \bibinfo{journal}{\emph{arXiv preprint arXiv:2211.02556}}
  (\bibinfo{year}{2022}).
\newblock


\bibitem[\protect\citeauthoryear{Bishoi, Prakash, Jain, et~al\mbox{.}}{Bishoi
  et~al\mbox{.}}{2009}]%
        {bishoi2009comparative}
\bibfield{author}{\bibinfo{person}{Biswanath Bishoi}, \bibinfo{person}{Amit
  Prakash}, \bibinfo{person}{VK Jain}, {et~al\mbox{.}}}
  \bibinfo{year}{2009}\natexlab{}.
\newblock \showarticletitle{A comparative study of air quality index based on
  factor analysis and US-EPA methods for an urban environment}.
\newblock \bibinfo{journal}{\emph{Aerosol and Air Quality Research}}
  \bibinfo{volume}{9}, \bibinfo{number}{1} (\bibinfo{year}{2009}),
  \bibinfo{pages}{1--17}.
\newblock


\bibitem[\protect\citeauthoryear{Blum and Mitchell}{Blum and Mitchell}{1998}]%
        {blum1998combining}
\bibfield{author}{\bibinfo{person}{Avrim Blum} {and} \bibinfo{person}{Tom
  Mitchell}.} \bibinfo{year}{1998}\natexlab{}.
\newblock \showarticletitle{Combining labeled and unlabeled data with
  co-training}. In \bibinfo{booktitle}{\emph{Proceedings of the eleventh annual
  conference on Computational learning theory}}. \bibinfo{pages}{92--100}.
\newblock


\bibitem[\protect\citeauthoryear{Breiman}{Breiman}{2001}]%
        {breiman2001random}
\bibfield{author}{\bibinfo{person}{Leo Breiman}.}
  \bibinfo{year}{2001}\natexlab{}.
\newblock \showarticletitle{Random forests}.
\newblock \bibinfo{journal}{\emph{Machine learning}}  \bibinfo{volume}{45}
  (\bibinfo{year}{2001}), \bibinfo{pages}{5--32}.
\newblock


\bibitem[\protect\citeauthoryear{Cao, Wang, Li, Zhou, Li, and Li}{Cao
  et~al\mbox{.}}{2018}]%
        {cao2018brits}
\bibfield{author}{\bibinfo{person}{Wei Cao}, \bibinfo{person}{Dong Wang},
  \bibinfo{person}{Jian Li}, \bibinfo{person}{Hao Zhou}, \bibinfo{person}{Lei
  Li}, {and} \bibinfo{person}{Yitan Li}.} \bibinfo{year}{2018}\natexlab{}.
\newblock \showarticletitle{Brits: Bidirectional recurrent imputation for time
  series}.
\newblock \bibinfo{journal}{\emph{Advances in neural information processing
  systems}}  \bibinfo{volume}{31} (\bibinfo{year}{2018}).
\newblock


\bibitem[\protect\citeauthoryear{Chen, Cai, Ding, Lv, Yuan, and Chen}{Chen
  et~al\mbox{.}}{2016}]%
        {chen2016spatially}
\bibfield{author}{\bibinfo{person}{Ling Chen}, \bibinfo{person}{Yaya Cai},
  \bibinfo{person}{Yifang Ding}, \bibinfo{person}{Mingqi Lv},
  \bibinfo{person}{Cuili Yuan}, {and} \bibinfo{person}{Gencai Chen}.}
  \bibinfo{year}{2016}\natexlab{}.
\newblock \showarticletitle{Spatially fine-grained urban air quality estimation
  using ensemble semi-supervised learning and pruning}. In
  \bibinfo{booktitle}{\emph{Proceedings of the 2016 ACM international joint
  conference on pervasive and ubiquitous computing}}.
  \bibinfo{pages}{1076--1087}.
\newblock


\bibitem[\protect\citeauthoryear{Chen, Xu, Wu, Qian, Du, Li, and Zhang}{Chen
  et~al\mbox{.}}{2021}]%
        {chen2021group}
\bibfield{author}{\bibinfo{person}{Ling Chen}, \bibinfo{person}{Jiahui Xu},
  \bibinfo{person}{Binqing Wu}, \bibinfo{person}{Yuntao Qian},
  \bibinfo{person}{Zhenhong Du}, \bibinfo{person}{Yansheng Li}, {and}
  \bibinfo{person}{Yongjun Zhang}.} \bibinfo{year}{2021}\natexlab{}.
\newblock \showarticletitle{Group-aware graph neural network for nationwide
  city air quality forecasting}.
\newblock \bibinfo{journal}{\emph{arXiv preprint arXiv:2108.12238}}
  (\bibinfo{year}{2021}).
\newblock


\bibitem[\protect\citeauthoryear{Chen, Yu, Geng, Li, and Zhang}{Chen
  et~al\mbox{.}}{2020}]%
        {chen2020evanet}
\bibfield{author}{\bibinfo{person}{Zechuan Chen}, \bibinfo{person}{Haomin Yu},
  \bibinfo{person}{Yangli-ao Geng}, \bibinfo{person}{Qingyong Li}, {and}
  \bibinfo{person}{Yingjun Zhang}.} \bibinfo{year}{2020}\natexlab{}.
\newblock \showarticletitle{Evanet: An extreme value attention network for
  long-term air quality prediction}. In \bibinfo{booktitle}{\emph{2020 IEEE
  International Conference on Big Data (Big Data)}}. IEEE,
  \bibinfo{pages}{4545--4552}.
\newblock


\bibitem[\protect\citeauthoryear{Cheng, Shen, Zhu, and Huang}{Cheng
  et~al\mbox{.}}{2018}]%
        {cheng2018neural}
\bibfield{author}{\bibinfo{person}{Weiyu Cheng}, \bibinfo{person}{Yanyan Shen},
  \bibinfo{person}{Yanmin Zhu}, {and} \bibinfo{person}{Linpeng Huang}.}
  \bibinfo{year}{2018}\natexlab{}.
\newblock \showarticletitle{A neural attention model for urban air quality
  inference: Learning the weights of monitoring stations}. In
  \bibinfo{booktitle}{\emph{Proceedings of the AAAI Conference on Artificial
  Intelligence}}, Vol.~\bibinfo{volume}{32}.
\newblock


\bibitem[\protect\citeauthoryear{Cheng}{Cheng}{1995}]%
        {cheng1995mean}
\bibfield{author}{\bibinfo{person}{Yizong Cheng}.}
  \bibinfo{year}{1995}\natexlab{}.
\newblock \showarticletitle{Mean shift, mode seeking, and clustering}.
\newblock \bibinfo{journal}{\emph{IEEE transactions on pattern analysis and
  machine intelligence}} \bibinfo{volume}{17}, \bibinfo{number}{8}
  (\bibinfo{year}{1995}), \bibinfo{pages}{790--799}.
\newblock


\bibitem[\protect\citeauthoryear{Cheng, He, Zhou, and Thiele}{Cheng
  et~al\mbox{.}}{2019}]%
        {cheng2019ict}
\bibfield{author}{\bibinfo{person}{Yun Cheng}, \bibinfo{person}{Xiaoxi He},
  \bibinfo{person}{Zimu Zhou}, {and} \bibinfo{person}{Lothar Thiele}.}
  \bibinfo{year}{2019}\natexlab{}.
\newblock \showarticletitle{Ict: In-field calibration transfer for air quality
  sensor deployments}.
\newblock \bibinfo{journal}{\emph{Proceedings of the ACM on Interactive,
  Mobile, Wearable and Ubiquitous Technologies}} \bibinfo{volume}{3},
  \bibinfo{number}{1} (\bibinfo{year}{2019}), \bibinfo{pages}{1--19}.
\newblock


\bibitem[\protect\citeauthoryear{Cheng, Li, and Li}{Cheng
  et~al\mbox{.}}{2016}]%
        {cheng2016finding}
\bibfield{author}{\bibinfo{person}{Yun Cheng}, \bibinfo{person}{Xiucheng Li},
  {and} \bibinfo{person}{Yan Li}.} \bibinfo{year}{2016}\natexlab{}.
\newblock \showarticletitle{Finding dynamic co-evolving zones in
  spatial-temporal time series data}. In \bibinfo{booktitle}{\emph{Machine
  Learning and Knowledge Discovery in Databases: European Conference, ECML PKDD
  2016, Riva del Garda, Italy, September 19-23, 2016, Proceedings, Part III
  16}}. Springer, \bibinfo{pages}{129--144}.
\newblock


\bibitem[\protect\citeauthoryear{Cheng, Saukh, and Thiele}{Cheng
  et~al\mbox{.}}{2021}]%
        {cheng2021tip}
\bibfield{author}{\bibinfo{person}{Yun Cheng}, \bibinfo{person}{Olga Saukh},
  {and} \bibinfo{person}{Lothar Thiele}.} \bibinfo{year}{2021}\natexlab{}.
\newblock \showarticletitle{TIP-Air: Tracking pollution transfer for accurate
  air quality prediction}. In \bibinfo{booktitle}{\emph{Adjunct Proceedings of
  the 2021 ACM International Joint Conference on Pervasive and Ubiquitous
  Computing and Proceedings of the 2021 ACM International Symposium on Wearable
  Computers}}. \bibinfo{pages}{589--599}.
\newblock


\bibitem[\protect\citeauthoryear{Cho, Van~Merri{\"e}nboer, Bahdanau, and
  Bengio}{Cho et~al\mbox{.}}{2014}]%
        {cho2014properties}
\bibfield{author}{\bibinfo{person}{Kyunghyun Cho}, \bibinfo{person}{Bart
  Van~Merri{\"e}nboer}, \bibinfo{person}{Dzmitry Bahdanau}, {and}
  \bibinfo{person}{Yoshua Bengio}.} \bibinfo{year}{2014}\natexlab{}.
\newblock \showarticletitle{On the properties of neural machine translation:
  Encoder-decoder approaches}.
\newblock \bibinfo{journal}{\emph{arXiv preprint arXiv:1409.1259}}
  (\bibinfo{year}{2014}).
\newblock


\bibitem[\protect\citeauthoryear{Cini, Marisca, and Alippi}{Cini
  et~al\mbox{.}}{2021}]%
        {cini2021filling}
\bibfield{author}{\bibinfo{person}{Andrea Cini}, \bibinfo{person}{Ivan
  Marisca}, {and} \bibinfo{person}{Cesare Alippi}.}
  \bibinfo{year}{2021}\natexlab{}.
\newblock \showarticletitle{Filling the g\_ap\_s: Multivariate time series
  imputation by graph neural networks}.
\newblock \bibinfo{journal}{\emph{arXiv preprint arXiv:2108.00298}}
  (\bibinfo{year}{2021}).
\newblock


\bibitem[\protect\citeauthoryear{Cini, Marisca, Zambon, and Alippi}{Cini
  et~al\mbox{.}}{2023}]%
        {cini2023taming}
\bibfield{author}{\bibinfo{person}{Andrea Cini}, \bibinfo{person}{Ivan
  Marisca}, \bibinfo{person}{Daniele Zambon}, {and} \bibinfo{person}{Cesare
  Alippi}.} \bibinfo{year}{2023}\natexlab{}.
\newblock \showarticletitle{Taming Local Effects in Graph-based Spatiotemporal
  Forecasting}.
\newblock \bibinfo{journal}{\emph{arXiv preprint arXiv:2302.04071}}
  (\bibinfo{year}{2023}).
\newblock


\bibitem[\protect\citeauthoryear{Concas, Mineraud, Lagerspetz, Varjonen, Liu,
  Puolam{\"a}ki, Nurmi, and Tarkoma}{Concas et~al\mbox{.}}{2021}]%
        {concas2021low}
\bibfield{author}{\bibinfo{person}{Francesco Concas}, \bibinfo{person}{Julien
  Mineraud}, \bibinfo{person}{Eemil Lagerspetz}, \bibinfo{person}{Samu
  Varjonen}, \bibinfo{person}{Xiaoli Liu}, \bibinfo{person}{Kai Puolam{\"a}ki},
  \bibinfo{person}{Petteri Nurmi}, {and} \bibinfo{person}{Sasu Tarkoma}.}
  \bibinfo{year}{2021}\natexlab{}.
\newblock \showarticletitle{Low-cost outdoor air quality monitoring and sensor
  calibration: A survey and critical analysis}.
\newblock \bibinfo{journal}{\emph{ACM Transactions on Sensor Networks (TOSN)}}
  \bibinfo{volume}{17}, \bibinfo{number}{2} (\bibinfo{year}{2021}),
  \bibinfo{pages}{1--44}.
\newblock


\bibitem[\protect\citeauthoryear{Cressie}{Cressie}{2015}]%
        {cressie2015statistics}
\bibfield{author}{\bibinfo{person}{Noel Cressie}.}
  \bibinfo{year}{2015}\natexlab{}.
\newblock \bibinfo{booktitle}{\emph{Statistics for spatial data}}.
\newblock \bibinfo{publisher}{John Wiley \& Sons}.
\newblock


\bibitem[\protect\citeauthoryear{Cross, Williams, Lewis, Magoon, Onasch,
  Kaminsky, Worsnop, and Jayne}{Cross et~al\mbox{.}}{2017}]%
        {cross2017use}
\bibfield{author}{\bibinfo{person}{Eben~S Cross}, \bibinfo{person}{Leah~R
  Williams}, \bibinfo{person}{David~K Lewis}, \bibinfo{person}{Gregory~R
  Magoon}, \bibinfo{person}{Timothy~B Onasch}, \bibinfo{person}{Michael~L
  Kaminsky}, \bibinfo{person}{Douglas~R Worsnop}, {and} \bibinfo{person}{John~T
  Jayne}.} \bibinfo{year}{2017}\natexlab{}.
\newblock \showarticletitle{Use of electrochemical sensors for measurement of
  air pollution: correcting interference response and validating measurements}.
\newblock \bibinfo{journal}{\emph{Atmospheric Measurement Techniques}}
  \bibinfo{volume}{10}, \bibinfo{number}{9} (\bibinfo{year}{2017}),
  \bibinfo{pages}{3575--3588}.
\newblock


\bibitem[\protect\citeauthoryear{Deng, Zhao, Liu, Jia, and Wang}{Deng
  et~al\mbox{.}}{2023}]%
        {deng2023spatio}
\bibfield{author}{\bibinfo{person}{Pan Deng}, \bibinfo{person}{Yu Zhao},
  \bibinfo{person}{Junting Liu}, \bibinfo{person}{Xiaofeng Jia}, {and}
  \bibinfo{person}{Mulan Wang}.} \bibinfo{year}{2023}\natexlab{}.
\newblock \showarticletitle{Spatio-temporal neural structural causal models for
  bike flow prediction}.
\newblock \bibinfo{journal}{\emph{arXiv preprint arXiv:2301.07843}}
  (\bibinfo{year}{2023}).
\newblock


\bibitem[\protect\citeauthoryear{Deng, Rangwala, and Ning}{Deng
  et~al\mbox{.}}{2022}]%
        {deng2022robust}
\bibfield{author}{\bibinfo{person}{Songgaojun Deng}, \bibinfo{person}{Huzefa
  Rangwala}, {and} \bibinfo{person}{Yue Ning}.}
  \bibinfo{year}{2022}\natexlab{}.
\newblock \showarticletitle{Robust Event Forecasting with Spatiotemporal
  Confounder Learning}. In \bibinfo{booktitle}{\emph{Proceedings of the 28th
  ACM SIGKDD Conference on Knowledge Discovery and Data Mining}}.
  \bibinfo{pages}{294--304}.
\newblock


\bibitem[\protect\citeauthoryear{Deng, Weng, Chen, Liu, Wang, Bao, Zheng, and
  Wu}{Deng et~al\mbox{.}}{2019}]%
        {deng2019airvis}
\bibfield{author}{\bibinfo{person}{Zikun Deng}, \bibinfo{person}{Di Weng},
  \bibinfo{person}{Jiahui Chen}, \bibinfo{person}{Ren Liu},
  \bibinfo{person}{Zhibin Wang}, \bibinfo{person}{Jie Bao}, \bibinfo{person}{Yu
  Zheng}, {and} \bibinfo{person}{Yingcai Wu}.} \bibinfo{year}{2019}\natexlab{}.
\newblock \showarticletitle{AirVis: Visual analytics of air pollution
  propagation}.
\newblock \bibinfo{journal}{\emph{IEEE transactions on visualization and
  computer graphics}} \bibinfo{volume}{26}, \bibinfo{number}{1}
  (\bibinfo{year}{2019}), \bibinfo{pages}{800--810}.
\newblock


\bibitem[\protect\citeauthoryear{Deng, Weng, Liang, Bao, Zheng, Schreck, Xu,
  and Wu}{Deng et~al\mbox{.}}{2021}]%
        {deng2021visual}
\bibfield{author}{\bibinfo{person}{Zikun Deng}, \bibinfo{person}{Di Weng},
  \bibinfo{person}{Yuxuan Liang}, \bibinfo{person}{Jie Bao},
  \bibinfo{person}{Yu Zheng}, \bibinfo{person}{Tobias Schreck},
  \bibinfo{person}{Mingliang Xu}, {and} \bibinfo{person}{Yingcai Wu}.}
  \bibinfo{year}{2021}\natexlab{}.
\newblock \showarticletitle{Visual cascade analytics of large-scale
  spatiotemporal data}.
\newblock \bibinfo{journal}{\emph{IEEE Transactions on Visualization and
  Computer Graphics}} \bibinfo{volume}{28}, \bibinfo{number}{6}
  (\bibinfo{year}{2021}), \bibinfo{pages}{2486--2499}.
\newblock


\bibitem[\protect\citeauthoryear{Drucker, Burges, Kaufman, Smola, and
  Vapnik}{Drucker et~al\mbox{.}}{1996}]%
        {drucker1996support}
\bibfield{author}{\bibinfo{person}{Harris Drucker},
  \bibinfo{person}{Christopher~J Burges}, \bibinfo{person}{Linda Kaufman},
  \bibinfo{person}{Alex Smola}, {and} \bibinfo{person}{Vladimir Vapnik}.}
  \bibinfo{year}{1996}\natexlab{}.
\newblock \showarticletitle{Support vector regression machines}.
\newblock \bibinfo{journal}{\emph{Advances in neural information processing
  systems}}  \bibinfo{volume}{9} (\bibinfo{year}{1996}).
\newblock


\bibitem[\protect\citeauthoryear{Du, Li, Yang, and Horng}{Du
  et~al\mbox{.}}{2019}]%
        {du2019deep}
\bibfield{author}{\bibinfo{person}{Shengdong Du}, \bibinfo{person}{Tianrui Li},
  \bibinfo{person}{Yan Yang}, {and} \bibinfo{person}{Shi-Jinn Horng}.}
  \bibinfo{year}{2019}\natexlab{}.
\newblock \showarticletitle{Deep air quality forecasting using hybrid deep
  learning framework}.
\newblock \bibinfo{journal}{\emph{IEEE Transactions on Knowledge and Data
  Engineering}} \bibinfo{volume}{33}, \bibinfo{number}{6}
  (\bibinfo{year}{2019}), \bibinfo{pages}{2412--2424}.
\newblock


\bibitem[\protect\citeauthoryear{Fan, Liu, Liu, Ge, Xiong, and Fu}{Fan
  et~al\mbox{.}}{2021}]%
        {fan2021interactive}
\bibfield{author}{\bibinfo{person}{Wei Fan}, \bibinfo{person}{Kunpeng Liu},
  \bibinfo{person}{Hao Liu}, \bibinfo{person}{Yong Ge}, \bibinfo{person}{Hui
  Xiong}, {and} \bibinfo{person}{Yanjie Fu}.} \bibinfo{year}{2021}\natexlab{}.
\newblock \showarticletitle{Interactive reinforcement learning for feature
  selection with decision tree in the loop}.
\newblock \bibinfo{journal}{\emph{IEEE Transactions on Knowledge and Data
  Engineering}} (\bibinfo{year}{2021}).
\newblock


\bibitem[\protect\citeauthoryear{Gal and Ghahramani}{Gal and
  Ghahramani}{2016}]%
        {gal2016dropout}
\bibfield{author}{\bibinfo{person}{Yarin Gal} {and} \bibinfo{person}{Zoubin
  Ghahramani}.} \bibinfo{year}{2016}\natexlab{}.
\newblock \showarticletitle{Dropout as a bayesian approximation: Representing
  model uncertainty in deep learning}. In
  \bibinfo{booktitle}{\emph{international conference on machine learning}}.
  PMLR, \bibinfo{pages}{1050--1059}.
\newblock


\bibitem[\protect\citeauthoryear{Gao, Dong, Guo, Liu, Chen, Liu, Bu, and
  Chen}{Gao et~al\mbox{.}}{2016}]%
        {gao2016mosaic}
\bibfield{author}{\bibinfo{person}{Yi Gao}, \bibinfo{person}{Wei Dong},
  \bibinfo{person}{Kai Guo}, \bibinfo{person}{Xue Liu}, \bibinfo{person}{Yuan
  Chen}, \bibinfo{person}{Xiaojin Liu}, \bibinfo{person}{Jiajun Bu}, {and}
  \bibinfo{person}{Chun Chen}.} \bibinfo{year}{2016}\natexlab{}.
\newblock \showarticletitle{Mosaic: A low-cost mobile sensing system for urban
  air quality monitoring}. In \bibinfo{booktitle}{\emph{IEEE INFOCOM 2016-The
  35th Annual IEEE International Conference on Computer Communications}}. IEEE,
  \bibinfo{pages}{1--9}.
\newblock


\bibitem[\protect\citeauthoryear{Guo, Wang, Yu, Wang, Yen, Huang, and Zhou}{Guo
  et~al\mbox{.}}{2015}]%
        {guo2015mobile}
\bibfield{author}{\bibinfo{person}{Bin Guo}, \bibinfo{person}{Zhu Wang},
  \bibinfo{person}{Zhiwen Yu}, \bibinfo{person}{Yu Wang},
  \bibinfo{person}{Neil~Y Yen}, \bibinfo{person}{Runhe Huang}, {and}
  \bibinfo{person}{Xingshe Zhou}.} \bibinfo{year}{2015}\natexlab{}.
\newblock \showarticletitle{Mobile crowd sensing and computing: The review of
  an emerging human-powered sensing paradigm}.
\newblock \bibinfo{journal}{\emph{ACM computing surveys (CSUR)}}
  \bibinfo{volume}{48}, \bibinfo{number}{1} (\bibinfo{year}{2015}),
  \bibinfo{pages}{1--31}.
\newblock


\bibitem[\protect\citeauthoryear{Han, Liu, Xiong, and Yang}{Han
  et~al\mbox{.}}{2022}]%
        {han2022semi}
\bibfield{author}{\bibinfo{person}{Jindong Han}, \bibinfo{person}{Hao Liu},
  \bibinfo{person}{Haoyi Xiong}, {and} \bibinfo{person}{Jing Yang}.}
  \bibinfo{year}{2022}\natexlab{}.
\newblock \showarticletitle{Semi-Supervised Air Quality Forecasting via
  Self-Supervised Hierarchical Graph Neural Network}.
\newblock \bibinfo{journal}{\emph{IEEE Transactions on Knowledge and Data
  Engineering}} (\bibinfo{year}{2022}).
\newblock


\bibitem[\protect\citeauthoryear{Han, Liu, Zhu, and Xiong}{Han
  et~al\mbox{.}}{2023}]%
        {han2023kill}
\bibfield{author}{\bibinfo{person}{Jindong Han}, \bibinfo{person}{Hao Liu},
  \bibinfo{person}{Hengshu Zhu}, {and} \bibinfo{person}{Hui Xiong}.}
  \bibinfo{year}{2023}\natexlab{}.
\newblock \showarticletitle{Kill Two Birds with One Stone: A Multi-View
  Multi-Adversarial Learning Approach for Joint Air Quality and Weather
  Prediction}.
\newblock \bibinfo{journal}{\emph{IEEE Transactions on Knowledge and Data
  Engineering}} (\bibinfo{year}{2023}).
\newblock


\bibitem[\protect\citeauthoryear{Han, Liu, Zhu, Xiong, and Dou}{Han
  et~al\mbox{.}}{2021a}]%
        {han2021joint}
\bibfield{author}{\bibinfo{person}{Jindong Han}, \bibinfo{person}{Hao Liu},
  \bibinfo{person}{Hengshu Zhu}, \bibinfo{person}{Hui Xiong}, {and}
  \bibinfo{person}{Dejing Dou}.} \bibinfo{year}{2021}\natexlab{a}.
\newblock \showarticletitle{Joint air quality and weather prediction based on
  multi-adversarial spatiotemporal networks}. In
  \bibinfo{booktitle}{\emph{Proceedings of the AAAI Conference on Artificial
  Intelligence}}, Vol.~\bibinfo{volume}{35}. \bibinfo{pages}{4081--4089}.
\newblock


\bibitem[\protect\citeauthoryear{Han, Lu, and Chen}{Han et~al\mbox{.}}{2021b}]%
        {han2021fine}
\bibfield{author}{\bibinfo{person}{Qilong Han}, \bibinfo{person}{Dan Lu}, {and}
  \bibinfo{person}{Rui Chen}.} \bibinfo{year}{2021}\natexlab{b}.
\newblock \showarticletitle{Fine-Grained Air Quality Inference via
  Multi-Channel Attention Model.}. In \bibinfo{booktitle}{\emph{IJCAI}}.
  \bibinfo{pages}{2512--2518}.
\newblock


\bibitem[\protect\citeauthoryear{Han, Lam, Li, and Zhang}{Han
  et~al\mbox{.}}{2020}]%
        {han2020domain}
\bibfield{author}{\bibinfo{person}{Yang Han}, \bibinfo{person}{Jacqueline~CK
  Lam}, \bibinfo{person}{Victor~OK Li}, {and} \bibinfo{person}{Qi Zhang}.}
  \bibinfo{year}{2020}\natexlab{}.
\newblock \showarticletitle{A domain-specific Bayesian deep-learning approach
  for air pollution forecast}.
\newblock \bibinfo{journal}{\emph{IEEE Transactions on Big Data}}
  \bibinfo{volume}{8}, \bibinfo{number}{4} (\bibinfo{year}{2020}),
  \bibinfo{pages}{1034--1046}.
\newblock


\bibitem[\protect\citeauthoryear{Hasenfratz, Saukh, Walser, Hueglin, Fierz,
  Arn, Beutel, and Thiele}{Hasenfratz et~al\mbox{.}}{2015}]%
        {hasenfratz2015deriving}
\bibfield{author}{\bibinfo{person}{David Hasenfratz}, \bibinfo{person}{Olga
  Saukh}, \bibinfo{person}{Christoph Walser}, \bibinfo{person}{Christoph
  Hueglin}, \bibinfo{person}{Martin Fierz}, \bibinfo{person}{Tabita Arn},
  \bibinfo{person}{Jan Beutel}, {and} \bibinfo{person}{Lothar Thiele}.}
  \bibinfo{year}{2015}\natexlab{}.
\newblock \showarticletitle{Deriving high-resolution urban air pollution maps
  using mobile sensor nodes}.
\newblock \bibinfo{journal}{\emph{Pervasive and Mobile Computing}}
  \bibinfo{volume}{16} (\bibinfo{year}{2015}), \bibinfo{pages}{268--285}.
\newblock


\bibitem[\protect\citeauthoryear{Hasenfratz, Saukh, Walser, Hueglin, Fierz, and
  Thiele}{Hasenfratz et~al\mbox{.}}{2014}]%
        {hasenfratz2014pushing}
\bibfield{author}{\bibinfo{person}{David Hasenfratz}, \bibinfo{person}{Olga
  Saukh}, \bibinfo{person}{Christoph Walser}, \bibinfo{person}{Christoph
  Hueglin}, \bibinfo{person}{Martin Fierz}, {and} \bibinfo{person}{Lothar
  Thiele}.} \bibinfo{year}{2014}\natexlab{}.
\newblock \showarticletitle{Pushing the spatio-temporal resolution limit of
  urban air pollution maps}. In \bibinfo{booktitle}{\emph{2014 IEEE
  International Conference on Pervasive Computing and Communications
  (PerCom)}}. IEEE, \bibinfo{pages}{69--77}.
\newblock


\bibitem[\protect\citeauthoryear{Hochreiter and Schmidhuber}{Hochreiter and
  Schmidhuber}{1997}]%
        {hochreiter1997long}
\bibfield{author}{\bibinfo{person}{Sepp Hochreiter} {and}
  \bibinfo{person}{J{\"u}rgen Schmidhuber}.} \bibinfo{year}{1997}\natexlab{}.
\newblock \showarticletitle{Long short-term memory}.
\newblock \bibinfo{journal}{\emph{Neural computation}} \bibinfo{volume}{9},
  \bibinfo{number}{8} (\bibinfo{year}{1997}), \bibinfo{pages}{1735--1780}.
\newblock


\bibitem[\protect\citeauthoryear{Hoek, Beelen, De~Hoogh, Vienneau, Gulliver,
  Fischer, and Briggs}{Hoek et~al\mbox{.}}{2008}]%
        {hoek2008review}
\bibfield{author}{\bibinfo{person}{Gerard Hoek}, \bibinfo{person}{Rob Beelen},
  \bibinfo{person}{Kees De~Hoogh}, \bibinfo{person}{Danielle Vienneau},
  \bibinfo{person}{John Gulliver}, \bibinfo{person}{Paul Fischer}, {and}
  \bibinfo{person}{David Briggs}.} \bibinfo{year}{2008}\natexlab{}.
\newblock \showarticletitle{A review of land-use regression models to assess
  spatial variation of outdoor air pollution}.
\newblock \bibinfo{journal}{\emph{Atmospheric environment}}
  \bibinfo{volume}{42}, \bibinfo{number}{33} (\bibinfo{year}{2008}),
  \bibinfo{pages}{7561--7578}.
\newblock


\bibitem[\protect\citeauthoryear{Hsieh, Lin, and Zheng}{Hsieh
  et~al\mbox{.}}{2015}]%
        {hsieh2015inferring}
\bibfield{author}{\bibinfo{person}{Hsun-Ping Hsieh}, \bibinfo{person}{Shou-De
  Lin}, {and} \bibinfo{person}{Yu Zheng}.} \bibinfo{year}{2015}\natexlab{}.
\newblock \showarticletitle{Inferring air quality for station location
  recommendation based on urban big data}. In
  \bibinfo{booktitle}{\emph{Proceedings of the 21th ACM SIGKDD international
  conference on knowledge discovery and data mining}}.
  \bibinfo{pages}{437--446}.
\newblock


\bibitem[\protect\citeauthoryear{Huang, Li, Liu, Xie, Du, and Teng}{Huang
  et~al\mbox{.}}{2021}]%
        {huang2021overview}
\bibfield{author}{\bibinfo{person}{Wei Huang}, \bibinfo{person}{Tianrui Li},
  \bibinfo{person}{Jia Liu}, \bibinfo{person}{Peng Xie},
  \bibinfo{person}{Shengdong Du}, {and} \bibinfo{person}{Fei Teng}.}
  \bibinfo{year}{2021}\natexlab{}.
\newblock \showarticletitle{An overview of air quality analysis by big data
  techniques: Monitoring, forecasting, and traceability}.
\newblock \bibinfo{journal}{\emph{Information Fusion}}  \bibinfo{volume}{75}
  (\bibinfo{year}{2021}), \bibinfo{pages}{28--40}.
\newblock


\bibitem[\protect\citeauthoryear{Ji, Xue, and Carin}{Ji et~al\mbox{.}}{2008}]%
        {ji2008bayesian}
\bibfield{author}{\bibinfo{person}{Shihao Ji}, \bibinfo{person}{Ya Xue}, {and}
  \bibinfo{person}{Lawrence Carin}.} \bibinfo{year}{2008}\natexlab{}.
\newblock \showarticletitle{Bayesian compressive sensing}.
\newblock \bibinfo{journal}{\emph{IEEE Transactions on signal processing}}
  \bibinfo{volume}{56}, \bibinfo{number}{6} (\bibinfo{year}{2008}),
  \bibinfo{pages}{2346--2356}.
\newblock


\bibitem[\protect\citeauthoryear{Jiang, Coenen, and Zito}{Jiang
  et~al\mbox{.}}{2013}]%
        {jiang2013survey}
\bibfield{author}{\bibinfo{person}{Chuntao Jiang}, \bibinfo{person}{Frans
  Coenen}, {and} \bibinfo{person}{Michele Zito}.}
  \bibinfo{year}{2013}\natexlab{}.
\newblock \showarticletitle{A survey of frequent subgraph mining algorithms}.
\newblock \bibinfo{journal}{\emph{The Knowledge Engineering Review}}
  \bibinfo{volume}{28}, \bibinfo{number}{1} (\bibinfo{year}{2013}),
  \bibinfo{pages}{75--105}.
\newblock


\bibitem[\protect\citeauthoryear{Jiang, Sun, Wang, and Young}{Jiang
  et~al\mbox{.}}{2019}]%
        {jiang2019enhancing}
\bibfield{author}{\bibinfo{person}{Jyun-Yu Jiang}, \bibinfo{person}{Xue Sun},
  \bibinfo{person}{Wei Wang}, {and} \bibinfo{person}{Sean Young}.}
  \bibinfo{year}{2019}\natexlab{}.
\newblock \showarticletitle{Enhancing air quality prediction with social media
  and natural language processing}. In \bibinfo{booktitle}{\emph{Proceedings of
  the 57th Annual Meeting of the Association for Computational Linguistics}}.
  \bibinfo{pages}{2627--2632}.
\newblock


\bibitem[\protect\citeauthoryear{Jiang, Wang, Tsou, and Fu}{Jiang
  et~al\mbox{.}}{2015}]%
        {jiang2015using}
\bibfield{author}{\bibinfo{person}{Wei Jiang}, \bibinfo{person}{Yandong Wang},
  \bibinfo{person}{Ming-Hsiang Tsou}, {and} \bibinfo{person}{Xiaokang Fu}.}
  \bibinfo{year}{2015}\natexlab{}.
\newblock \showarticletitle{Using social media to detect outdoor air pollution
  and monitor air quality index (AQI): a geo-targeted spatiotemporal analysis
  framework with Sina Weibo (Chinese Twitter)}.
\newblock \bibinfo{journal}{\emph{PloS one}} \bibinfo{volume}{10},
  \bibinfo{number}{10} (\bibinfo{year}{2015}), \bibinfo{pages}{e0141185}.
\newblock


\bibitem[\protect\citeauthoryear{Jin, Liang, Fang, Huang, Zhang, and Zheng}{Jin
  et~al\mbox{.}}{2023}]%
        {jin2023spatio}
\bibfield{author}{\bibinfo{person}{Guangyin Jin}, \bibinfo{person}{Yuxuan
  Liang}, \bibinfo{person}{Yuchen Fang}, \bibinfo{person}{Jincai Huang},
  \bibinfo{person}{Junbo Zhang}, {and} \bibinfo{person}{Yu Zheng}.}
  \bibinfo{year}{2023}\natexlab{}.
\newblock \showarticletitle{Spatio-Temporal Graph Neural Networks for
  Predictive Learning in Urban Computing: A Survey}.
\newblock \bibinfo{journal}{\emph{arXiv preprint arXiv:2303.14483}}
  (\bibinfo{year}{2023}).
\newblock


\bibitem[\protect\citeauthoryear{Junninen, Niska, Tuppurainen, Ruuskanen, and
  Kolehmainen}{Junninen et~al\mbox{.}}{2004}]%
        {junninen2004methods}
\bibfield{author}{\bibinfo{person}{Heikki Junninen}, \bibinfo{person}{Harri
  Niska}, \bibinfo{person}{Kari Tuppurainen}, \bibinfo{person}{Juhani
  Ruuskanen}, {and} \bibinfo{person}{Mikko Kolehmainen}.}
  \bibinfo{year}{2004}\natexlab{}.
\newblock \showarticletitle{Methods for imputation of missing values in air
  quality data sets}.
\newblock \bibinfo{journal}{\emph{Atmospheric environment}}
  \bibinfo{volume}{38}, \bibinfo{number}{18} (\bibinfo{year}{2004}),
  \bibinfo{pages}{2895--2907}.
\newblock


\bibitem[\protect\citeauthoryear{Jutzeler, Li, and Faltings}{Jutzeler
  et~al\mbox{.}}{2014}]%
        {jutzeler2014region}
\bibfield{author}{\bibinfo{person}{Arnaud Jutzeler}, \bibinfo{person}{Jason
  Li}, {and} \bibinfo{person}{Boi Faltings}.} \bibinfo{year}{2014}\natexlab{}.
\newblock \showarticletitle{A region-based model for estimating urban air
  pollution}. In \bibinfo{booktitle}{\emph{Proceedings of the AAAI Conference
  on Artificial Intelligence}}, Vol.~\bibinfo{volume}{28}.
\newblock


\bibitem[\protect\citeauthoryear{Kaur, Singh, Jabarulla, Kumar, Kang, and
  Lee}{Kaur et~al\mbox{.}}{2023}]%
        {kaur2023computational}
\bibfield{author}{\bibinfo{person}{Manjit Kaur}, \bibinfo{person}{Dilbag
  Singh}, \bibinfo{person}{Mohamed~Yaseen Jabarulla}, \bibinfo{person}{Vijay
  Kumar}, \bibinfo{person}{Jusung Kang}, {and} \bibinfo{person}{Heung-No Lee}.}
  \bibinfo{year}{2023}\natexlab{}.
\newblock \showarticletitle{Computational deep air quality prediction
  techniques: a systematic review}.
\newblock \bibinfo{journal}{\emph{Artificial Intelligence Review}}
  (\bibinfo{year}{2023}), \bibinfo{pages}{1--46}.
\newblock


\bibitem[\protect\citeauthoryear{Keats, Yee, and Lien}{Keats
  et~al\mbox{.}}{2007}]%
        {keats2007bayesian}
\bibfield{author}{\bibinfo{person}{Andrew Keats}, \bibinfo{person}{Eugene Yee},
  {and} \bibinfo{person}{Fue-Sang Lien}.} \bibinfo{year}{2007}\natexlab{}.
\newblock \showarticletitle{Bayesian inference for source determination with
  applications to a complex urban environment}.
\newblock \bibinfo{journal}{\emph{Atmospheric environment}}
  \bibinfo{volume}{41}, \bibinfo{number}{3} (\bibinfo{year}{2007}),
  \bibinfo{pages}{465--479}.
\newblock


\bibitem[\protect\citeauthoryear{Laine and Aila}{Laine and Aila}{2016}]%
        {laine2016temporal}
\bibfield{author}{\bibinfo{person}{Samuli Laine} {and} \bibinfo{person}{Timo
  Aila}.} \bibinfo{year}{2016}\natexlab{}.
\newblock \showarticletitle{Temporal ensembling for semi-supervised learning}.
\newblock \bibinfo{journal}{\emph{arXiv preprint arXiv:1610.02242}}
  (\bibinfo{year}{2016}).
\newblock


\bibitem[\protect\citeauthoryear{Lakshminarayanan, Pritzel, and
  Blundell}{Lakshminarayanan et~al\mbox{.}}{2017}]%
        {lakshminarayanan2017simple}
\bibfield{author}{\bibinfo{person}{Balaji Lakshminarayanan},
  \bibinfo{person}{Alexander Pritzel}, {and} \bibinfo{person}{Charles
  Blundell}.} \bibinfo{year}{2017}\natexlab{}.
\newblock \showarticletitle{Simple and scalable predictive uncertainty
  estimation using deep ensembles}.
\newblock \bibinfo{journal}{\emph{Advances in neural information processing
  systems}}  \bibinfo{volume}{30} (\bibinfo{year}{2017}).
\newblock


\bibitem[\protect\citeauthoryear{Lam, Sanchez-Gonzalez, Willson, Wirnsberger,
  Fortunato, Pritzel, Ravuri, Ewalds, Alet, Eaton-Rosen, et~al\mbox{.}}{Lam
  et~al\mbox{.}}{2022}]%
        {lam2022graphcast}
\bibfield{author}{\bibinfo{person}{Remi Lam}, \bibinfo{person}{Alvaro
  Sanchez-Gonzalez}, \bibinfo{person}{Matthew Willson}, \bibinfo{person}{Peter
  Wirnsberger}, \bibinfo{person}{Meire Fortunato}, \bibinfo{person}{Alexander
  Pritzel}, \bibinfo{person}{Suman Ravuri}, \bibinfo{person}{Timo Ewalds},
  \bibinfo{person}{Ferran Alet}, \bibinfo{person}{Zach Eaton-Rosen},
  {et~al\mbox{.}}} \bibinfo{year}{2022}\natexlab{}.
\newblock \showarticletitle{GraphCast: Learning skillful medium-range global
  weather forecasting}.
\newblock \bibinfo{journal}{\emph{arXiv preprint arXiv:2212.12794}}
  (\bibinfo{year}{2022}).
\newblock


\bibitem[\protect\citeauthoryear{LeCun et~al\mbox{.}}{LeCun
  et~al\mbox{.}}{1989}]%
        {lecun1989generalization}
\bibfield{author}{\bibinfo{person}{Yann LeCun} {et~al\mbox{.}}}
  \bibinfo{year}{1989}\natexlab{}.
\newblock \showarticletitle{Generalization and network design strategies}.
\newblock \bibinfo{journal}{\emph{Connectionism in perspective}}
  \bibinfo{volume}{19}, \bibinfo{number}{143-155} (\bibinfo{year}{1989}),
  \bibinfo{pages}{18}.
\newblock


\bibitem[\protect\citeauthoryear{LeCun, Bengio, and Hinton}{LeCun
  et~al\mbox{.}}{2015}]%
        {lecun2015deep}
\bibfield{author}{\bibinfo{person}{Yann LeCun}, \bibinfo{person}{Yoshua
  Bengio}, {and} \bibinfo{person}{Geoffrey Hinton}.}
  \bibinfo{year}{2015}\natexlab{}.
\newblock \showarticletitle{Deep learning}.
\newblock \bibinfo{journal}{\emph{nature}} \bibinfo{volume}{521},
  \bibinfo{number}{7553} (\bibinfo{year}{2015}), \bibinfo{pages}{436--444}.
\newblock


\bibitem[\protect\citeauthoryear{Lee, Liu, Wang, Russell, and Edgerton}{Lee
  et~al\mbox{.}}{2008}]%
        {lee2008source}
\bibfield{author}{\bibinfo{person}{Sangil Lee}, \bibinfo{person}{Wei Liu},
  \bibinfo{person}{Yuhang Wang}, \bibinfo{person}{Armistead~G Russell}, {and}
  \bibinfo{person}{Eric~S Edgerton}.} \bibinfo{year}{2008}\natexlab{}.
\newblock \showarticletitle{Source apportionment of PM2. 5: Comparing PMF and
  CMB results for four ambient monitoring sites in the southeastern United
  States}.
\newblock \bibinfo{journal}{\emph{Atmospheric Environment}}
  \bibinfo{volume}{42}, \bibinfo{number}{18} (\bibinfo{year}{2008}),
  \bibinfo{pages}{4126--4137}.
\newblock


\bibitem[\protect\citeauthoryear{Li, Yu, Geng, Li, and Li}{Li
  et~al\mbox{.}}{2021}]%
        {li2021ddgnet}
\bibfield{author}{\bibinfo{person}{Dong Li}, \bibinfo{person}{Haomin Yu},
  \bibinfo{person}{Yangli-ao Geng}, \bibinfo{person}{Xiaobao Li}, {and}
  \bibinfo{person}{Qingyong Li}.} \bibinfo{year}{2021}\natexlab{}.
\newblock \showarticletitle{DDGNet: A Dual-Stage Dynamic Spatio-Temporal Graph
  Network for PM 2.5 Forecasting}. In \bibinfo{booktitle}{\emph{2021 IEEE
  International Conference on Big Data (Big Data)}}. IEEE,
  \bibinfo{pages}{1679--1685}.
\newblock


\bibitem[\protect\citeauthoryear{Li and Heap}{Li and Heap}{2011}]%
        {li2011review}
\bibfield{author}{\bibinfo{person}{Jin Li} {and} \bibinfo{person}{Andrew~D
  Heap}.} \bibinfo{year}{2011}\natexlab{}.
\newblock \showarticletitle{A review of comparative studies of spatial
  interpolation methods in environmental sciences: Performance and impact
  factors}.
\newblock \bibinfo{journal}{\emph{Ecological Informatics}} \bibinfo{volume}{6},
  \bibinfo{number}{3-4} (\bibinfo{year}{2011}), \bibinfo{pages}{228--241}.
\newblock


\bibitem[\protect\citeauthoryear{Li and Heap}{Li and Heap}{2014}]%
        {li2014spatial}
\bibfield{author}{\bibinfo{person}{Jin Li} {and} \bibinfo{person}{Andrew~D
  Heap}.} \bibinfo{year}{2014}\natexlab{}.
\newblock \showarticletitle{Spatial interpolation methods applied in the
  environmental sciences: A review}.
\newblock \bibinfo{journal}{\emph{Environmental Modelling \& Software}}
  \bibinfo{volume}{53} (\bibinfo{year}{2014}), \bibinfo{pages}{173--189}.
\newblock


\bibitem[\protect\citeauthoryear{Li, Faltings, Saukh, Hasenfratz, and
  Beutel}{Li et~al\mbox{.}}{2012}]%
        {li2012sensing}
\bibfield{author}{\bibinfo{person}{Jason~Jingshi Li}, \bibinfo{person}{Boi
  Faltings}, \bibinfo{person}{Olga Saukh}, \bibinfo{person}{David Hasenfratz},
  {and} \bibinfo{person}{Jan Beutel}.} \bibinfo{year}{2012}\natexlab{}.
\newblock \showarticletitle{Sensing the air we breathe—the OpenSense Zurich
  dataset}. In \bibinfo{booktitle}{\emph{Proceedings of the AAAI Conference on
  Artificial Intelligence}}, Vol.~\bibinfo{volume}{26}.
  \bibinfo{pages}{323--325}.
\newblock


\bibitem[\protect\citeauthoryear{Li, Cheng, Cong, and Chen}{Li
  et~al\mbox{.}}{2017a}]%
        {li2017discovering}
\bibfield{author}{\bibinfo{person}{Xiucheng Li}, \bibinfo{person}{Yun Cheng},
  \bibinfo{person}{Gao Cong}, {and} \bibinfo{person}{Lisi Chen}.}
  \bibinfo{year}{2017}\natexlab{a}.
\newblock \showarticletitle{Discovering pollution sources and propagation
  patterns in urban area}. In \bibinfo{booktitle}{\emph{Proceedings of the 23rd
  ACM SIGKDD international conference on knowledge discovery and data mining}}.
  \bibinfo{pages}{1863--1872}.
\newblock


\bibitem[\protect\citeauthoryear{Li, Yu, Shahabi, and Liu}{Li
  et~al\mbox{.}}{2017b}]%
        {li2017diffusion}
\bibfield{author}{\bibinfo{person}{Yaguang Li}, \bibinfo{person}{Rose Yu},
  \bibinfo{person}{Cyrus Shahabi}, {and} \bibinfo{person}{Yan Liu}.}
  \bibinfo{year}{2017}\natexlab{b}.
\newblock \showarticletitle{Diffusion convolutional recurrent neural network:
  Data-driven traffic forecasting}.
\newblock \bibinfo{journal}{\emph{arXiv preprint arXiv:1707.01926}}
  (\bibinfo{year}{2017}).
\newblock


\bibitem[\protect\citeauthoryear{Liang, Ke, Zhang, Yi, and Zheng}{Liang
  et~al\mbox{.}}{2018}]%
        {liang2018geoman}
\bibfield{author}{\bibinfo{person}{Yuxuan Liang}, \bibinfo{person}{Songyu Ke},
  \bibinfo{person}{Junbo Zhang}, \bibinfo{person}{Xiuwen Yi}, {and}
  \bibinfo{person}{Yu Zheng}.} \bibinfo{year}{2018}\natexlab{}.
\newblock \showarticletitle{Geoman: Multi-level attention networks for
  geo-sensory time series prediction.}. In \bibinfo{booktitle}{\emph{IJCAI}},
  Vol.~\bibinfo{volume}{2018}. \bibinfo{pages}{3428--3434}.
\newblock


\bibitem[\protect\citeauthoryear{Liang, Xia, Ke, Wang, Wen, Zhang, Zheng, and
  Zimmermann}{Liang et~al\mbox{.}}{2022}]%
        {liang2022airformer}
\bibfield{author}{\bibinfo{person}{Yuxuan Liang}, \bibinfo{person}{Yutong Xia},
  \bibinfo{person}{Songyu Ke}, \bibinfo{person}{Yiwei Wang},
  \bibinfo{person}{Qingsong Wen}, \bibinfo{person}{Junbo Zhang},
  \bibinfo{person}{Yu Zheng}, {and} \bibinfo{person}{Roger Zimmermann}.}
  \bibinfo{year}{2022}\natexlab{}.
\newblock \showarticletitle{AirFormer: Predicting Nationwide Air Quality in
  China with Transformers}.
\newblock \bibinfo{journal}{\emph{arXiv preprint arXiv:2211.15979}}
  (\bibinfo{year}{2022}).
\newblock


\bibitem[\protect\citeauthoryear{Lin, Chiang, Pan, Stripelis, Ambite, Eckel,
  and Habre}{Lin et~al\mbox{.}}{2017}]%
        {lin2017mining}
\bibfield{author}{\bibinfo{person}{Yijun Lin}, \bibinfo{person}{Yao-Yi Chiang},
  \bibinfo{person}{Fan Pan}, \bibinfo{person}{Dimitrios Stripelis},
  \bibinfo{person}{Jos{\'e}~Luis Ambite}, \bibinfo{person}{Sandrah~P Eckel},
  {and} \bibinfo{person}{Rima Habre}.} \bibinfo{year}{2017}\natexlab{}.
\newblock \showarticletitle{Mining public datasets for modeling intra-city PM2.
  5 concentrations at a fine spatial resolution}. In
  \bibinfo{booktitle}{\emph{Proceedings of the 25th ACM SIGSPATIAL
  international conference on advances in geographic information systems}}.
  \bibinfo{pages}{1--10}.
\newblock


\bibitem[\protect\citeauthoryear{Lin, Dong, and Chen}{Lin
  et~al\mbox{.}}{2018a}]%
        {lin2018calibrating}
\bibfield{author}{\bibinfo{person}{Yuxiang Lin}, \bibinfo{person}{Wei Dong},
  {and} \bibinfo{person}{Yuan Chen}.} \bibinfo{year}{2018}\natexlab{a}.
\newblock \showarticletitle{Calibrating low-cost sensors by a two-phase
  learning approach for urban air quality measurement}.
\newblock \bibinfo{journal}{\emph{Proceedings of the ACM on Interactive,
  Mobile, Wearable and Ubiquitous Technologies}} \bibinfo{volume}{2},
  \bibinfo{number}{1} (\bibinfo{year}{2018}), \bibinfo{pages}{1--18}.
\newblock


\bibitem[\protect\citeauthoryear{Lin, Mago, Gao, Li, Chiang, Shahabi, and
  Ambite}{Lin et~al\mbox{.}}{2018b}]%
        {lin2018exploiting}
\bibfield{author}{\bibinfo{person}{Yijun Lin}, \bibinfo{person}{Nikhit Mago},
  \bibinfo{person}{Yu Gao}, \bibinfo{person}{Yaguang Li},
  \bibinfo{person}{Yao-Yi Chiang}, \bibinfo{person}{Cyrus Shahabi}, {and}
  \bibinfo{person}{Jos{\'e}~Luis Ambite}.} \bibinfo{year}{2018}\natexlab{b}.
\newblock \showarticletitle{Exploiting spatiotemporal patterns for accurate air
  quality forecasting using deep learning}. In
  \bibinfo{booktitle}{\emph{Proceedings of the 26th ACM SIGSPATIAL
  international conference on advances in geographic information systems}}.
  \bibinfo{pages}{359--368}.
\newblock


\bibitem[\protect\citeauthoryear{Liu, Han, Fu, Li, Chen, and Xiong}{Liu
  et~al\mbox{.}}{2022a}]%
        {liu2022unified}
\bibfield{author}{\bibinfo{person}{Hao Liu}, \bibinfo{person}{Jindong Han},
  \bibinfo{person}{Yanjie Fu}, \bibinfo{person}{Yanyan Li},
  \bibinfo{person}{Kai Chen}, {and} \bibinfo{person}{Hui Xiong}.}
  \bibinfo{year}{2022}\natexlab{a}.
\newblock \showarticletitle{Unified route representation learning for
  multi-modal transportation recommendation with spatiotemporal pre-training}.
\newblock \bibinfo{journal}{\emph{The VLDB Journal}} (\bibinfo{year}{2022}),
  \bibinfo{pages}{1--18}.
\newblock


\bibitem[\protect\citeauthoryear{Liu, Wu, Zhuang, Lu, Dou, and Xiong}{Liu
  et~al\mbox{.}}{2021}]%
        {hao2021demand}
\bibfield{author}{\bibinfo{person}{Hao Liu}, \bibinfo{person}{Qiyu Wu},
  \bibinfo{person}{Fuzhen Zhuang}, \bibinfo{person}{Xinjiang Lu},
  \bibinfo{person}{Dejing Dou}, {and} \bibinfo{person}{Hui Xiong}.}
  \bibinfo{year}{2021}\natexlab{}.
\newblock \showarticletitle{Community-Aware Multi-Task Transportation Demand
  Prediction}. In \bibinfo{booktitle}{\emph{Proceedings of the Thirty-Fifth
  {AAAI} Conference on Artificial Intelligence}}, Vol.~\bibinfo{volume}{35}.
  \bibinfo{pages}{320--327}.
\newblock


\bibitem[\protect\citeauthoryear{Liu, Liu, Zheng, Ma, and Zhang}{Liu
  et~al\mbox{.}}{2018}]%
        {liu2018third}
\bibfield{author}{\bibinfo{person}{Liang Liu}, \bibinfo{person}{Wu Liu},
  \bibinfo{person}{Yu Zheng}, \bibinfo{person}{Huadong Ma}, {and}
  \bibinfo{person}{Cheng Zhang}.} \bibinfo{year}{2018}\natexlab{}.
\newblock \showarticletitle{Third-eye: A mobilephone-enabled crowdsensing
  system for air quality monitoring}.
\newblock \bibinfo{journal}{\emph{Proceedings of the ACM on Interactive,
  Mobile, Wearable and Ubiquitous Technologies}} \bibinfo{volume}{2},
  \bibinfo{number}{1} (\bibinfo{year}{2018}), \bibinfo{pages}{1--26}.
\newblock


\bibitem[\protect\citeauthoryear{Liu, Lu, Zhang, Liu, and Jiang}{Liu
  et~al\mbox{.}}{2022b}]%
        {liu2022data}
\bibfield{author}{\bibinfo{person}{Xian Liu}, \bibinfo{person}{Dawei Lu},
  \bibinfo{person}{Aiqian Zhang}, \bibinfo{person}{Qian Liu}, {and}
  \bibinfo{person}{Guibin Jiang}.} \bibinfo{year}{2022}\natexlab{b}.
\newblock \showarticletitle{Data-driven machine learning in environmental
  pollution: gains and problems}.
\newblock \bibinfo{journal}{\emph{Environmental science \& technology}}
  \bibinfo{volume}{56}, \bibinfo{number}{4} (\bibinfo{year}{2022}),
  \bibinfo{pages}{2124--2133}.
\newblock


\bibitem[\protect\citeauthoryear{Liu, Zhao, Lin, Li, Wang, Zhang, Gao, and
  Chai}{Liu et~al\mbox{.}}{2022c}]%
        {liu2022fine}
\bibfield{author}{\bibinfo{person}{Xiliang Liu}, \bibinfo{person}{Junjie Zhao},
  \bibinfo{person}{Shaofu Lin}, \bibinfo{person}{Jianqiang Li},
  \bibinfo{person}{Shaohua Wang}, \bibinfo{person}{Yumin Zhang},
  \bibinfo{person}{Yuyao Gao}, {and} \bibinfo{person}{Jinchuan Chai}.}
  \bibinfo{year}{2022}\natexlab{c}.
\newblock \showarticletitle{Fine-Grained Individual Air Quality Index (IAQI)
  Prediction Based on Spatial-Temporal Causal Convolution Network: A Case Study
  of Shanghai}.
\newblock \bibinfo{journal}{\emph{Atmosphere}} \bibinfo{volume}{13},
  \bibinfo{number}{6} (\bibinfo{year}{2022}), \bibinfo{pages}{959}.
\newblock


\bibitem[\protect\citeauthoryear{Luo, Cai, Zhang, Xu, et~al\mbox{.}}{Luo
  et~al\mbox{.}}{2018}]%
        {luo2018multivariate}
\bibfield{author}{\bibinfo{person}{Yonghong Luo}, \bibinfo{person}{Xiangrui
  Cai}, \bibinfo{person}{Ying Zhang}, \bibinfo{person}{Jun Xu},
  {et~al\mbox{.}}} \bibinfo{year}{2018}\natexlab{}.
\newblock \showarticletitle{Multivariate time series imputation with generative
  adversarial networks}.
\newblock \bibinfo{journal}{\emph{Advances in neural information processing
  systems}}  \bibinfo{volume}{31} (\bibinfo{year}{2018}).
\newblock


\bibitem[\protect\citeauthoryear{Luo, Zhang, Cai, and Yuan}{Luo
  et~al\mbox{.}}{2019b}]%
        {luo2019e2gan}
\bibfield{author}{\bibinfo{person}{Yonghong Luo}, \bibinfo{person}{Ying Zhang},
  \bibinfo{person}{Xiangrui Cai}, {and} \bibinfo{person}{Xiaojie Yuan}.}
  \bibinfo{year}{2019}\natexlab{b}.
\newblock \showarticletitle{E2gan: End-to-end generative adversarial network
  for multivariate time series imputation}. In
  \bibinfo{booktitle}{\emph{Proceedings of the 28th international joint
  conference on artificial intelligence}}. AAAI Press,
  \bibinfo{pages}{3094--3100}.
\newblock


\bibitem[\protect\citeauthoryear{Luo, Huang, Hu, Li, and Zhang}{Luo
  et~al\mbox{.}}{2019a}]%
        {luo2019accuair}
\bibfield{author}{\bibinfo{person}{Zhipeng Luo}, \bibinfo{person}{Jianqiang
  Huang}, \bibinfo{person}{Ke Hu}, \bibinfo{person}{Xue Li}, {and}
  \bibinfo{person}{Peng Zhang}.} \bibinfo{year}{2019}\natexlab{a}.
\newblock \showarticletitle{AccuAir: Winning solution to air quality prediction
  for KDD Cup 2018}. In \bibinfo{booktitle}{\emph{Proceedings of the 25th ACM
  SIGKDD International Conference on Knowledge Discovery \& Data Mining}}.
  \bibinfo{pages}{1842--1850}.
\newblock


\bibitem[\protect\citeauthoryear{Ma, Liu, Xu, Wang, Noh, Zhang, and Zhang}{Ma
  et~al\mbox{.}}{2020}]%
        {ma2020fine}
\bibfield{author}{\bibinfo{person}{Rui Ma}, \bibinfo{person}{Ning Liu},
  \bibinfo{person}{Xiangxiang Xu}, \bibinfo{person}{Yue Wang},
  \bibinfo{person}{Hae~Young Noh}, \bibinfo{person}{Pei Zhang}, {and}
  \bibinfo{person}{Lin Zhang}.} \bibinfo{year}{2020}\natexlab{}.
\newblock \showarticletitle{Fine-grained air pollution inference with mobile
  sensing systems: A weather-related deep autoencoder model}.
\newblock \bibinfo{journal}{\emph{Proceedings of the ACM on Interactive,
  Mobile, Wearable and Ubiquitous Technologies}} \bibinfo{volume}{4},
  \bibinfo{number}{2} (\bibinfo{year}{2020}), \bibinfo{pages}{1--21}.
\newblock


\bibitem[\protect\citeauthoryear{Maag, Zhou, and Thiele}{Maag
  et~al\mbox{.}}{2018a}]%
        {maag2018survey}
\bibfield{author}{\bibinfo{person}{Balz Maag}, \bibinfo{person}{Zimu Zhou},
  {and} \bibinfo{person}{Lothar Thiele}.} \bibinfo{year}{2018}\natexlab{a}.
\newblock \showarticletitle{A survey on sensor calibration in air pollution
  monitoring deployments}.
\newblock \bibinfo{journal}{\emph{IEEE Internet of Things Journal}}
  \bibinfo{volume}{5}, \bibinfo{number}{6} (\bibinfo{year}{2018}),
  \bibinfo{pages}{4857--4870}.
\newblock


\bibitem[\protect\citeauthoryear{Maag, Zhou, and Thiele}{Maag
  et~al\mbox{.}}{2018b}]%
        {maag2018w}
\bibfield{author}{\bibinfo{person}{Balz Maag}, \bibinfo{person}{Zimu Zhou},
  {and} \bibinfo{person}{Lothar Thiele}.} \bibinfo{year}{2018}\natexlab{b}.
\newblock \showarticletitle{W-air: Enabling personal air pollution monitoring
  on wearables}.
\newblock \bibinfo{journal}{\emph{Proceedings of the ACM on Interactive,
  Mobile, Wearable and Ubiquitous Technologies}} \bibinfo{volume}{2},
  \bibinfo{number}{1} (\bibinfo{year}{2018}), \bibinfo{pages}{1--25}.
\newblock


\bibitem[\protect\citeauthoryear{Makin, Moses, and Chang}{Makin
  et~al\mbox{.}}{2020}]%
        {makin2020machine}
\bibfield{author}{\bibinfo{person}{Joseph~G Makin}, \bibinfo{person}{David~A
  Moses}, {and} \bibinfo{person}{Edward~F Chang}.}
  \bibinfo{year}{2020}\natexlab{}.
\newblock \showarticletitle{Machine translation of cortical activity to text
  with an encoder--decoder framework}.
\newblock \bibinfo{journal}{\emph{Nature neuroscience}} \bibinfo{volume}{23},
  \bibinfo{number}{4} (\bibinfo{year}{2020}), \bibinfo{pages}{575--582}.
\newblock


\bibitem[\protect\citeauthoryear{Marisca, Cini, and Alippi}{Marisca
  et~al\mbox{.}}{2022}]%
        {marisca2022learning}
\bibfield{author}{\bibinfo{person}{Ivan Marisca}, \bibinfo{person}{Andrea
  Cini}, {and} \bibinfo{person}{Cesare Alippi}.}
  \bibinfo{year}{2022}\natexlab{}.
\newblock \showarticletitle{Learning to reconstruct missing data from
  spatiotemporal graphs with sparse observations}.
\newblock \bibinfo{journal}{\emph{arXiv preprint arXiv:2205.13479}}
  (\bibinfo{year}{2022}).
\newblock


\bibitem[\protect\citeauthoryear{Martin}{Martin}{2008}]%
        {martin2008satellite}
\bibfield{author}{\bibinfo{person}{Randall~V Martin}.}
  \bibinfo{year}{2008}\natexlab{}.
\newblock \showarticletitle{Satellite remote sensing of surface air quality}.
\newblock \bibinfo{journal}{\emph{Atmospheric environment}}
  \bibinfo{volume}{42}, \bibinfo{number}{34} (\bibinfo{year}{2008}),
  \bibinfo{pages}{7823--7843}.
\newblock


\bibitem[\protect\citeauthoryear{Masson, Piedrahita, and Hannigan}{Masson
  et~al\mbox{.}}{2015}]%
        {masson2015quantification}
\bibfield{author}{\bibinfo{person}{Nicholas Masson}, \bibinfo{person}{Ricardo
  Piedrahita}, {and} \bibinfo{person}{Michael Hannigan}.}
  \bibinfo{year}{2015}\natexlab{}.
\newblock \showarticletitle{Quantification method for electrolytic sensors in
  long-term monitoring of ambient air quality}.
\newblock \bibinfo{journal}{\emph{Sensors}} \bibinfo{volume}{15},
  \bibinfo{number}{10} (\bibinfo{year}{2015}), \bibinfo{pages}{27283--27302}.
\newblock


\bibitem[\protect\citeauthoryear{Mei, Li, Fan, Zhu, and Dyer}{Mei
  et~al\mbox{.}}{2014}]%
        {mei2014inferring}
\bibfield{author}{\bibinfo{person}{Shike Mei}, \bibinfo{person}{Han Li},
  \bibinfo{person}{Jing Fan}, \bibinfo{person}{Xiaojin Zhu}, {and}
  \bibinfo{person}{Charles~R Dyer}.} \bibinfo{year}{2014}\natexlab{}.
\newblock \showarticletitle{Inferring air pollution by sniffing social media}.
  In \bibinfo{booktitle}{\emph{2014 IEEE/ACM International Conference on
  Advances in Social Networks Analysis and Mining (ASONAM 2014)}}. IEEE,
  \bibinfo{pages}{534--539}.
\newblock


\bibitem[\protect\citeauthoryear{Miao, Wu, Wang, Gao, Mao, and Yin}{Miao
  et~al\mbox{.}}{2021}]%
        {miao2021generative}
\bibfield{author}{\bibinfo{person}{Xiaoye Miao}, \bibinfo{person}{Yangyang Wu},
  \bibinfo{person}{Jun Wang}, \bibinfo{person}{Yunjun Gao},
  \bibinfo{person}{Xudong Mao}, {and} \bibinfo{person}{Jianwei Yin}.}
  \bibinfo{year}{2021}\natexlab{}.
\newblock \showarticletitle{Generative semi-supervised learning for
  multivariate time series imputation}. In
  \bibinfo{booktitle}{\emph{Proceedings of the AAAI conference on artificial
  intelligence}}, Vol.~\bibinfo{volume}{35}. \bibinfo{pages}{8983--8991}.
\newblock


\bibitem[\protect\citeauthoryear{Miller}{Miller}{2004}]%
        {miller2004tobler}
\bibfield{author}{\bibinfo{person}{Harvey~J Miller}.}
  \bibinfo{year}{2004}\natexlab{}.
\newblock \showarticletitle{Tobler's first law and spatial analysis}.
\newblock \bibinfo{journal}{\emph{Annals of the association of American
  geographers}} \bibinfo{volume}{94}, \bibinfo{number}{2}
  (\bibinfo{year}{2004}), \bibinfo{pages}{284--289}.
\newblock


\bibitem[\protect\citeauthoryear{Motlagh, Lagerspetz, Nurmi, Li, Varjonen,
  Mineraud, Siekkinen, Rebeiro-Hargrave, Hussein, Petaja,
  et~al\mbox{.}}{Motlagh et~al\mbox{.}}{2020}]%
        {motlagh2020toward}
\bibfield{author}{\bibinfo{person}{Naser~Hossein Motlagh},
  \bibinfo{person}{Eemil Lagerspetz}, \bibinfo{person}{Petteri Nurmi},
  \bibinfo{person}{Xin Li}, \bibinfo{person}{Samu Varjonen},
  \bibinfo{person}{Julien Mineraud}, \bibinfo{person}{Matti Siekkinen},
  \bibinfo{person}{Andrew Rebeiro-Hargrave}, \bibinfo{person}{Tareq Hussein},
  \bibinfo{person}{Tuukka Petaja}, {et~al\mbox{.}}}
  \bibinfo{year}{2020}\natexlab{}.
\newblock \showarticletitle{Toward massive scale air quality monitoring}.
\newblock \bibinfo{journal}{\emph{IEEE Communications Magazine}}
  \bibinfo{volume}{58}, \bibinfo{number}{2} (\bibinfo{year}{2020}),
  \bibinfo{pages}{54--59}.
\newblock


\bibitem[\protect\citeauthoryear{Neal}{Neal}{2012}]%
        {neal2012bayesian}
\bibfield{author}{\bibinfo{person}{Radford~M Neal}.}
  \bibinfo{year}{2012}\natexlab{}.
\newblock \bibinfo{booktitle}{\emph{Bayesian learning for neural networks}}.
  Vol.~\bibinfo{volume}{118}.
\newblock \bibinfo{publisher}{Springer Science \& Business Media}.
\newblock


\bibitem[\protect\citeauthoryear{Nguyen, Liu, and Chen}{Nguyen
  et~al\mbox{.}}{2016}]%
        {nguyen2016discovering}
\bibfield{author}{\bibinfo{person}{Hoang Nguyen}, \bibinfo{person}{Wei Liu},
  {and} \bibinfo{person}{Fang Chen}.} \bibinfo{year}{2016}\natexlab{}.
\newblock \showarticletitle{Discovering congestion propagation patterns in
  spatio-temporal traffic data}.
\newblock \bibinfo{journal}{\emph{IEEE Transactions on Big Data}}
  \bibinfo{volume}{3}, \bibinfo{number}{2} (\bibinfo{year}{2016}),
  \bibinfo{pages}{169--180}.
\newblock


\bibitem[\protect\citeauthoryear{Oord, Dieleman, Zen, Simonyan, Vinyals,
  Graves, Kalchbrenner, Senior, and Kavukcuoglu}{Oord et~al\mbox{.}}{2016}]%
        {oord2016wavenet}
\bibfield{author}{\bibinfo{person}{Aaron van~den Oord}, \bibinfo{person}{Sander
  Dieleman}, \bibinfo{person}{Heiga Zen}, \bibinfo{person}{Karen Simonyan},
  \bibinfo{person}{Oriol Vinyals}, \bibinfo{person}{Alex Graves},
  \bibinfo{person}{Nal Kalchbrenner}, \bibinfo{person}{Andrew Senior}, {and}
  \bibinfo{person}{Koray Kavukcuoglu}.} \bibinfo{year}{2016}\natexlab{}.
\newblock \showarticletitle{Wavenet: A generative model for raw audio}.
\newblock \bibinfo{journal}{\emph{arXiv preprint arXiv:1609.03499}}
  (\bibinfo{year}{2016}).
\newblock


\bibitem[\protect\citeauthoryear{Organization et~al\mbox{.}}{Organization
  et~al\mbox{.}}{2015}]%
        {world2015economic}
\bibfield{author}{\bibinfo{person}{World~Health Organization} {et~al\mbox{.}}}
  \bibinfo{year}{2015}\natexlab{}.
\newblock \bibinfo{booktitle}{\emph{Economic cost of the health impact of air
  pollution in Europe: Clean air, health and wealth}}.
\newblock \bibinfo{type}{{T}echnical {R}eport}. \bibinfo{institution}{World
  Health Organization. Regional Office for Europe}.
\newblock


\bibitem[\protect\citeauthoryear{Ouyang, Wu, Jiang, Almeida, Wainwright,
  Mishkin, Zhang, Agarwal, Slama, Ray, et~al\mbox{.}}{Ouyang
  et~al\mbox{.}}{2022}]%
        {ouyang2022training}
\bibfield{author}{\bibinfo{person}{Long Ouyang}, \bibinfo{person}{Jeffrey Wu},
  \bibinfo{person}{Xu Jiang}, \bibinfo{person}{Diogo Almeida},
  \bibinfo{person}{Carroll Wainwright}, \bibinfo{person}{Pamela Mishkin},
  \bibinfo{person}{Chong Zhang}, \bibinfo{person}{Sandhini Agarwal},
  \bibinfo{person}{Katarina Slama}, \bibinfo{person}{Alex Ray},
  {et~al\mbox{.}}} \bibinfo{year}{2022}\natexlab{}.
\newblock \showarticletitle{Training language models to follow instructions
  with human feedback}.
\newblock \bibinfo{journal}{\emph{Advances in Neural Information Processing
  Systems}}  \bibinfo{volume}{35} (\bibinfo{year}{2022}),
  \bibinfo{pages}{27730--27744}.
\newblock


\bibitem[\protect\citeauthoryear{Pan, Yu, Miao, and Leung}{Pan
  et~al\mbox{.}}{2017}]%
        {pan2017crowdsensing}
\bibfield{author}{\bibinfo{person}{Zhengxiang Pan}, \bibinfo{person}{Han Yu},
  \bibinfo{person}{Chunyan Miao}, {and} \bibinfo{person}{Cyril Leung}.}
  \bibinfo{year}{2017}\natexlab{}.
\newblock \showarticletitle{Crowdsensing air quality with camera-enabled mobile
  devices}. In \bibinfo{booktitle}{\emph{Proceedings of the AAAI Conference on
  Artificial Intelligence}}, Vol.~\bibinfo{volume}{31}.
  \bibinfo{pages}{4728--4733}.
\newblock


\bibitem[\protect\citeauthoryear{Patel, Purohit, Patel, Sahni, and Batra}{Patel
  et~al\mbox{.}}{2022}]%
        {patel2022accurate}
\bibfield{author}{\bibinfo{person}{Zeel~B Patel}, \bibinfo{person}{Palak
  Purohit}, \bibinfo{person}{Harsh~M Patel}, \bibinfo{person}{Shivam Sahni},
  {and} \bibinfo{person}{Nipun Batra}.} \bibinfo{year}{2022}\natexlab{}.
\newblock \showarticletitle{Accurate and scalable Gaussian processes for
  fine-grained air quality inference}. In \bibinfo{booktitle}{\emph{Proceedings
  of the AAAI Conference on Artificial Intelligence}},
  Vol.~\bibinfo{volume}{36}. \bibinfo{pages}{12080--12088}.
\newblock


\bibitem[\protect\citeauthoryear{Qi, Wang, Song, Hu, Li, and Zhang}{Qi
  et~al\mbox{.}}{2018}]%
        {qi2018deep}
\bibfield{author}{\bibinfo{person}{Zhongang Qi}, \bibinfo{person}{Tianchun
  Wang}, \bibinfo{person}{Guojie Song}, \bibinfo{person}{Weisong Hu},
  \bibinfo{person}{Xi Li}, {and} \bibinfo{person}{Zhongfei Zhang}.}
  \bibinfo{year}{2018}\natexlab{}.
\newblock \showarticletitle{Deep air learning: Interpolation, prediction, and
  feature analysis of fine-grained air quality}.
\newblock \bibinfo{journal}{\emph{IEEE Transactions on Knowledge and Data
  Engineering}} \bibinfo{volume}{30}, \bibinfo{number}{12}
  (\bibinfo{year}{2018}), \bibinfo{pages}{2285--2297}.
\newblock


\bibitem[\protect\citeauthoryear{Qin, Zhan, Li, Yang, and Zheng}{Qin
  et~al\mbox{.}}{2021}]%
        {qin2021network}
\bibfield{author}{\bibinfo{person}{Huiling Qin}, \bibinfo{person}{Xianyuan
  Zhan}, \bibinfo{person}{Yuanxun Li}, \bibinfo{person}{Xiaodu Yang}, {and}
  \bibinfo{person}{Yu Zheng}.} \bibinfo{year}{2021}\natexlab{}.
\newblock \showarticletitle{Network-wide traffic states imputation using
  self-interested coalitional learning}. In
  \bibinfo{booktitle}{\emph{Proceedings of the 27th ACM SIGKDD Conference on
  Knowledge Discovery \& Data Mining}}. \bibinfo{pages}{1370--1378}.
\newblock


\bibitem[\protect\citeauthoryear{Radford, Kim, Hallacy, Ramesh, Goh, Agarwal,
  Sastry, Askell, Mishkin, Clark, et~al\mbox{.}}{Radford et~al\mbox{.}}{2021}]%
        {radford2021learning}
\bibfield{author}{\bibinfo{person}{Alec Radford}, \bibinfo{person}{Jong~Wook
  Kim}, \bibinfo{person}{Chris Hallacy}, \bibinfo{person}{Aditya Ramesh},
  \bibinfo{person}{Gabriel Goh}, \bibinfo{person}{Sandhini Agarwal},
  \bibinfo{person}{Girish Sastry}, \bibinfo{person}{Amanda Askell},
  \bibinfo{person}{Pamela Mishkin}, \bibinfo{person}{Jack Clark},
  {et~al\mbox{.}}} \bibinfo{year}{2021}\natexlab{}.
\newblock \showarticletitle{Learning transferable visual models from natural
  language supervision}. In \bibinfo{booktitle}{\emph{International conference
  on machine learning}}. PMLR, \bibinfo{pages}{8748--8763}.
\newblock


\bibitem[\protect\citeauthoryear{Ramanathan, Balzano, Burt, Estrin, Harmon,
  Harvey, Jay, Kohler, Rothenberg, and Srivastava}{Ramanathan
  et~al\mbox{.}}{2006}]%
        {ramanathan2006rapid}
\bibfield{author}{\bibinfo{person}{Nithya Ramanathan}, \bibinfo{person}{Laura
  Balzano}, \bibinfo{person}{Marci Burt}, \bibinfo{person}{Deborah Estrin},
  \bibinfo{person}{Tom Harmon}, \bibinfo{person}{Charlie Harvey},
  \bibinfo{person}{Jenny Jay}, \bibinfo{person}{Eddie Kohler},
  \bibinfo{person}{Sarah Rothenberg}, {and} \bibinfo{person}{Mani Srivastava}.}
  \bibinfo{year}{2006}\natexlab{}.
\newblock \showarticletitle{Rapid deployment with confidence: Calibration and
  fault detection in environmental sensor networks}.
\newblock  (\bibinfo{year}{2006}).
\newblock


\bibitem[\protect\citeauthoryear{Rasmus, Berglund, Honkala, Valpola, and
  Raiko}{Rasmus et~al\mbox{.}}{2015}]%
        {rasmus2015semi}
\bibfield{author}{\bibinfo{person}{Antti Rasmus}, \bibinfo{person}{Mathias
  Berglund}, \bibinfo{person}{Mikko Honkala}, \bibinfo{person}{Harri Valpola},
  {and} \bibinfo{person}{Tapani Raiko}.} \bibinfo{year}{2015}\natexlab{}.
\newblock \showarticletitle{Semi-supervised learning with ladder networks}.
\newblock \bibinfo{journal}{\emph{Advances in neural information processing
  systems}}  \bibinfo{volume}{28} (\bibinfo{year}{2015}).
\newblock


\bibitem[\protect\citeauthoryear{Saharia, Chan, Saxena, Li, Whang, Denton,
  Ghasemipour, Gontijo~Lopes, Karagol~Ayan, Salimans, et~al\mbox{.}}{Saharia
  et~al\mbox{.}}{2022}]%
        {saharia2022photorealistic}
\bibfield{author}{\bibinfo{person}{Chitwan Saharia}, \bibinfo{person}{William
  Chan}, \bibinfo{person}{Saurabh Saxena}, \bibinfo{person}{Lala Li},
  \bibinfo{person}{Jay Whang}, \bibinfo{person}{Emily~L Denton},
  \bibinfo{person}{Kamyar Ghasemipour}, \bibinfo{person}{Raphael
  Gontijo~Lopes}, \bibinfo{person}{Burcu Karagol~Ayan}, \bibinfo{person}{Tim
  Salimans}, {et~al\mbox{.}}} \bibinfo{year}{2022}\natexlab{}.
\newblock \showarticletitle{Photorealistic text-to-image diffusion models with
  deep language understanding}.
\newblock \bibinfo{journal}{\emph{Advances in Neural Information Processing
  Systems}}  \bibinfo{volume}{35} (\bibinfo{year}{2022}),
  \bibinfo{pages}{36479--36494}.
\newblock


\bibitem[\protect\citeauthoryear{Samal, Babu, and Das}{Samal
  et~al\mbox{.}}{2021}]%
        {samal2021temporal}
\bibfield{author}{\bibinfo{person}{K~Krishna~Rani Samal},
  \bibinfo{person}{Korra~Sathya Babu}, {and} \bibinfo{person}{Santos~Kumar
  Das}.} \bibinfo{year}{2021}\natexlab{}.
\newblock \showarticletitle{Temporal convolutional denoising autoencoder
  network for air pollution prediction with missing values}.
\newblock \bibinfo{journal}{\emph{Urban Climate}}  \bibinfo{volume}{38}
  (\bibinfo{year}{2021}), \bibinfo{pages}{100872}.
\newblock


\bibitem[\protect\citeauthoryear{Segovia~Dominguez, Lee, Chen, Garay, Gorski,
  and Gel}{Segovia~Dominguez et~al\mbox{.}}{2021}]%
        {segovia2021does}
\bibfield{author}{\bibinfo{person}{Ignacio Segovia~Dominguez},
  \bibinfo{person}{Huikyo Lee}, \bibinfo{person}{Yuzhou Chen},
  \bibinfo{person}{Michael Garay}, \bibinfo{person}{Krzysztof~M Gorski}, {and}
  \bibinfo{person}{Yulia~R Gel}.} \bibinfo{year}{2021}\natexlab{}.
\newblock \showarticletitle{Does air quality really impact COVID-19 clinical
  severity: coupling NASA satellite datasets with geometric deep learning}. In
  \bibinfo{booktitle}{\emph{Proceedings of the 27th ACM SIGKDD Conference on
  Knowledge Discovery \& Data Mining}}. \bibinfo{pages}{3540--3548}.
\newblock


\bibitem[\protect\citeauthoryear{Shang, Zheng, Tong, Chang, and Yu}{Shang
  et~al\mbox{.}}{2014}]%
        {shang2014inferring}
\bibfield{author}{\bibinfo{person}{Jingbo Shang}, \bibinfo{person}{Yu Zheng},
  \bibinfo{person}{Wenzhu Tong}, \bibinfo{person}{Eric Chang}, {and}
  \bibinfo{person}{Yong Yu}.} \bibinfo{year}{2014}\natexlab{}.
\newblock \showarticletitle{Inferring gas consumption and pollution emission of
  vehicles throughout a city}. In \bibinfo{booktitle}{\emph{Proceedings of the
  20th ACM SIGKDD international conference on Knowledge discovery and data
  mining}}. \bibinfo{pages}{1027--1036}.
\newblock


\bibitem[\protect\citeauthoryear{Shi, Chen, Wang, Yeung, Wong, and Woo}{Shi
  et~al\mbox{.}}{2015}]%
        {shi2015convolutional}
\bibfield{author}{\bibinfo{person}{Xingjian Shi}, \bibinfo{person}{Zhourong
  Chen}, \bibinfo{person}{Hao Wang}, \bibinfo{person}{Dit-Yan Yeung},
  \bibinfo{person}{Wai-Kin Wong}, {and} \bibinfo{person}{Wang-chun Woo}.}
  \bibinfo{year}{2015}\natexlab{}.
\newblock \showarticletitle{Convolutional LSTM network: A machine learning
  approach for precipitation nowcasting}.
\newblock \bibinfo{journal}{\emph{Advances in neural information processing
  systems}}  \bibinfo{volume}{28} (\bibinfo{year}{2015}).
\newblock


\bibitem[\protect\citeauthoryear{Song, Yang, Xu, and King}{Song
  et~al\mbox{.}}{2022}]%
        {song2022graph}
\bibfield{author}{\bibinfo{person}{Zixing Song}, \bibinfo{person}{Xiangli
  Yang}, \bibinfo{person}{Zenglin Xu}, {and} \bibinfo{person}{Irwin King}.}
  \bibinfo{year}{2022}\natexlab{}.
\newblock \showarticletitle{Graph-based semi-supervised learning: A
  comprehensive review}.
\newblock \bibinfo{journal}{\emph{IEEE Transactions on Neural Networks and
  Learning Systems}} (\bibinfo{year}{2022}).
\newblock


\bibitem[\protect\citeauthoryear{Sutskever, Vinyals, and Le}{Sutskever
  et~al\mbox{.}}{2014}]%
        {sutskever2014sequence}
\bibfield{author}{\bibinfo{person}{Ilya Sutskever}, \bibinfo{person}{Oriol
  Vinyals}, {and} \bibinfo{person}{Quoc~V Le}.}
  \bibinfo{year}{2014}\natexlab{}.
\newblock \showarticletitle{Sequence to sequence learning with neural
  networks}.
\newblock \bibinfo{journal}{\emph{Advances in neural information processing
  systems}}  \bibinfo{volume}{27} (\bibinfo{year}{2014}).
\newblock


\bibitem[\protect\citeauthoryear{Tarvainen and Valpola}{Tarvainen and
  Valpola}{2017}]%
        {tarvainen2017mean}
\bibfield{author}{\bibinfo{person}{Antti Tarvainen} {and}
  \bibinfo{person}{Harri Valpola}.} \bibinfo{year}{2017}\natexlab{}.
\newblock \showarticletitle{Mean teachers are better role models:
  Weight-averaged consistency targets improve semi-supervised deep learning
  results}.
\newblock \bibinfo{journal}{\emph{Advances in neural information processing
  systems}}  \bibinfo{volume}{30} (\bibinfo{year}{2017}).
\newblock


\bibitem[\protect\citeauthoryear{Tipping and Faul}{Tipping and Faul}{2003}]%
        {tipping2003fast}
\bibfield{author}{\bibinfo{person}{Michael~E Tipping} {and}
  \bibinfo{person}{Anita~C Faul}.} \bibinfo{year}{2003}\natexlab{}.
\newblock \showarticletitle{Fast marginal likelihood maximisation for sparse
  Bayesian models}. In \bibinfo{booktitle}{\emph{International workshop on
  artificial intelligence and statistics}}. PMLR, \bibinfo{pages}{276--283}.
\newblock


\bibitem[\protect\citeauthoryear{Tong, Zhou, Zeng, Chen, and Shahabi}{Tong
  et~al\mbox{.}}{2020}]%
        {tong2020spatial}
\bibfield{author}{\bibinfo{person}{Yongxin Tong}, \bibinfo{person}{Zimu Zhou},
  \bibinfo{person}{Yuxiang Zeng}, \bibinfo{person}{Lei Chen}, {and}
  \bibinfo{person}{Cyrus Shahabi}.} \bibinfo{year}{2020}\natexlab{}.
\newblock \showarticletitle{Spatial crowdsourcing: a survey}.
\newblock \bibinfo{journal}{\emph{The VLDB Journal}}  \bibinfo{volume}{29}
  (\bibinfo{year}{2020}), \bibinfo{pages}{217--250}.
\newblock


\bibitem[\protect\citeauthoryear{Van~Donkelaar, Martin, and Park}{Van~Donkelaar
  et~al\mbox{.}}{2006}]%
        {van2006estimating}
\bibfield{author}{\bibinfo{person}{Aaron Van~Donkelaar},
  \bibinfo{person}{Randall~V Martin}, {and} \bibinfo{person}{Rokjin~J Park}.}
  \bibinfo{year}{2006}\natexlab{}.
\newblock \showarticletitle{Estimating ground-level PM2. 5 using aerosol
  optical depth determined from satellite remote sensing}.
\newblock \bibinfo{journal}{\emph{Journal of Geophysical Research:
  Atmospheres}} \bibinfo{volume}{111}, \bibinfo{number}{D21}
  (\bibinfo{year}{2006}).
\newblock


\bibitem[\protect\citeauthoryear{Van~Engelen and Hoos}{Van~Engelen and
  Hoos}{2020}]%
        {van2020survey}
\bibfield{author}{\bibinfo{person}{Jesper~E Van~Engelen} {and}
  \bibinfo{person}{Holger~H Hoos}.} \bibinfo{year}{2020}\natexlab{}.
\newblock \showarticletitle{A survey on semi-supervised learning}.
\newblock \bibinfo{journal}{\emph{Machine Learning}} \bibinfo{volume}{109},
  \bibinfo{number}{2} (\bibinfo{year}{2020}), \bibinfo{pages}{373--440}.
\newblock


\bibitem[\protect\citeauthoryear{Vardoulakis, Fisher, Pericleous, and
  Gonzalez-Flesca}{Vardoulakis et~al\mbox{.}}{2003}]%
        {vardoulakis2003modelling}
\bibfield{author}{\bibinfo{person}{Sotiris Vardoulakis},
  \bibinfo{person}{Bernard~EA Fisher}, \bibinfo{person}{Koulis Pericleous},
  {and} \bibinfo{person}{Norbert Gonzalez-Flesca}.}
  \bibinfo{year}{2003}\natexlab{}.
\newblock \showarticletitle{Modelling air quality in street canyons: a review}.
\newblock \bibinfo{journal}{\emph{Atmospheric environment}}
  \bibinfo{volume}{37}, \bibinfo{number}{2} (\bibinfo{year}{2003}),
  \bibinfo{pages}{155--182}.
\newblock


\bibitem[\protect\citeauthoryear{Wang, Lu, Yan, Luo, Li, Zheng, and Zhang}{Wang
  et~al\mbox{.}}{2019a}]%
        {wang2019deep}
\bibfield{author}{\bibinfo{person}{Bin Wang}, \bibinfo{person}{Jie Lu},
  \bibinfo{person}{Zheng Yan}, \bibinfo{person}{Huaishao Luo},
  \bibinfo{person}{Tianrui Li}, \bibinfo{person}{Yu Zheng}, {and}
  \bibinfo{person}{Guangquan Zhang}.} \bibinfo{year}{2019}\natexlab{a}.
\newblock \showarticletitle{Deep uncertainty quantification: A machine learning
  approach for weather forecasting}. In \bibinfo{booktitle}{\emph{Proceedings
  of the 25th ACM SIGKDD International Conference on Knowledge Discovery \&
  Data Mining}}. \bibinfo{pages}{2087--2095}.
\newblock


\bibitem[\protect\citeauthoryear{Wang, Zhu, Zang, Liu, and Yu}{Wang
  et~al\mbox{.}}{2021}]%
        {wang2021modeling}
\bibfield{author}{\bibinfo{person}{Chunyang Wang}, \bibinfo{person}{Yanmin
  Zhu}, \bibinfo{person}{Tianzi Zang}, \bibinfo{person}{Haobing Liu}, {and}
  \bibinfo{person}{Jiadi Yu}.} \bibinfo{year}{2021}\natexlab{}.
\newblock \showarticletitle{Modeling inter-station relationships with attentive
  temporal graph convolutional network for air quality prediction}. In
  \bibinfo{booktitle}{\emph{Proceedings of the 14th ACM international
  conference on web search and data mining}}. \bibinfo{pages}{616--634}.
\newblock


\bibitem[\protect\citeauthoryear{Wang, Cao, and Yu}{Wang
  et~al\mbox{.}}{2020a}]%
        {wang2020deep}
\bibfield{author}{\bibinfo{person}{Senzhang Wang}, \bibinfo{person}{Jiannong
  Cao}, {and} \bibinfo{person}{Philip Yu}.} \bibinfo{year}{2020}\natexlab{a}.
\newblock \showarticletitle{Deep learning for spatio-temporal data mining: A
  survey}.
\newblock \bibinfo{journal}{\emph{IEEE transactions on knowledge and data
  engineering}} (\bibinfo{year}{2020}).
\newblock


\bibitem[\protect\citeauthoryear{Wang, Li, Zhang, Meng, Meng, and Gao}{Wang
  et~al\mbox{.}}{2020b}]%
        {wang2020pm2}
\bibfield{author}{\bibinfo{person}{Shuo Wang}, \bibinfo{person}{Yanran Li},
  \bibinfo{person}{Jiang Zhang}, \bibinfo{person}{Qingye Meng},
  \bibinfo{person}{Lingwei Meng}, {and} \bibinfo{person}{Fei Gao}.}
  \bibinfo{year}{2020}\natexlab{b}.
\newblock \showarticletitle{Pm2. 5-gnn: A domain knowledge enhanced graph
  neural network for pm2. 5 forecasting}. In
  \bibinfo{booktitle}{\emph{Proceedings of the 28th international conference on
  advances in geographic information systems}}. \bibinfo{pages}{163--166}.
\newblock


\bibitem[\protect\citeauthoryear{Wang, Meng, and Yuan}{Wang
  et~al\mbox{.}}{2018}]%
        {wang2018sparse}
\bibfield{author}{\bibinfo{person}{Yao Wang}, \bibinfo{person}{Deyu Meng},
  {and} \bibinfo{person}{Ming Yuan}.} \bibinfo{year}{2018}\natexlab{}.
\newblock \showarticletitle{Sparse recovery: from vectors to tensors}.
\newblock \bibinfo{journal}{\emph{National Science Review}}
  \bibinfo{volume}{5}, \bibinfo{number}{5} (\bibinfo{year}{2018}),
  \bibinfo{pages}{756--767}.
\newblock


\bibitem[\protect\citeauthoryear{Wang, Song, Du, and Lu}{Wang
  et~al\mbox{.}}{2019b}]%
        {wang2019real}
\bibfield{author}{\bibinfo{person}{Yun Wang}, \bibinfo{person}{Guojie Song},
  \bibinfo{person}{Lun Du}, {and} \bibinfo{person}{Zhicong Lu}.}
  \bibinfo{year}{2019}\natexlab{b}.
\newblock \showarticletitle{Real-Time Estimation of the Urban Air Quality with
  Mobile Sensor System}.
\newblock \bibinfo{journal}{\emph{ACM Transactions on Knowledge Discovery from
  Data (TKDD)}} \bibinfo{volume}{13}, \bibinfo{number}{5}
  (\bibinfo{year}{2019}), \bibinfo{pages}{11--19}.
\newblock


\bibitem[\protect\citeauthoryear{Willard, Jia, Xu, Steinbach, and
  Kumar}{Willard et~al\mbox{.}}{2020}]%
        {willard2020integrating}
\bibfield{author}{\bibinfo{person}{Jared Willard}, \bibinfo{person}{Xiaowei
  Jia}, \bibinfo{person}{Shaoming Xu}, \bibinfo{person}{Michael Steinbach},
  {and} \bibinfo{person}{Vipin Kumar}.} \bibinfo{year}{2020}\natexlab{}.
\newblock \showarticletitle{Integrating physics-based modeling with machine
  learning: A survey}.
\newblock \bibinfo{journal}{\emph{arXiv preprint arXiv:2003.04919}}
  \bibinfo{volume}{1}, \bibinfo{number}{1} (\bibinfo{year}{2020}),
  \bibinfo{pages}{1--34}.
\newblock


\bibitem[\protect\citeauthoryear{Wong, Yuan, and Perlin}{Wong
  et~al\mbox{.}}{2004}]%
        {wong2004comparison}
\bibfield{author}{\bibinfo{person}{David~W Wong}, \bibinfo{person}{Lester
  Yuan}, {and} \bibinfo{person}{Susan~A Perlin}.}
  \bibinfo{year}{2004}\natexlab{}.
\newblock \showarticletitle{Comparison of spatial interpolation methods for the
  estimation of air quality data}.
\newblock \bibinfo{journal}{\emph{Journal of Exposure Science \& Environmental
  Epidemiology}} \bibinfo{volume}{14}, \bibinfo{number}{5}
  (\bibinfo{year}{2004}), \bibinfo{pages}{404--415}.
\newblock


\bibitem[\protect\citeauthoryear{Wu, Gao, Chinazzi, Xiong, Vespignani, Ma, and
  Yu}{Wu et~al\mbox{.}}{2021}]%
        {wu2021quantifying}
\bibfield{author}{\bibinfo{person}{Dongxia Wu}, \bibinfo{person}{Liyao Gao},
  \bibinfo{person}{Matteo Chinazzi}, \bibinfo{person}{Xinyue Xiong},
  \bibinfo{person}{Alessandro Vespignani}, \bibinfo{person}{Yi-An Ma}, {and}
  \bibinfo{person}{Rose Yu}.} \bibinfo{year}{2021}\natexlab{}.
\newblock \showarticletitle{Quantifying uncertainty in deep spatiotemporal
  forecasting}. In \bibinfo{booktitle}{\emph{Proceedings of the 27th ACM SIGKDD
  Conference on Knowledge Discovery \& Data Mining}}.
  \bibinfo{pages}{1841--1851}.
\newblock


\bibitem[\protect\citeauthoryear{Wu, Xiao, Liao, Luo, Wu, Zhang, Li, and
  Guo}{Wu et~al\mbox{.}}{2020b}]%
        {wu2020sharing}
\bibfield{author}{\bibinfo{person}{Di Wu}, \bibinfo{person}{Tao Xiao},
  \bibinfo{person}{Xuewen Liao}, \bibinfo{person}{Jie Luo},
  \bibinfo{person}{Chao Wu}, \bibinfo{person}{Shigeng Zhang},
  \bibinfo{person}{Yong Li}, {and} \bibinfo{person}{Yike Guo}.}
  \bibinfo{year}{2020}\natexlab{b}.
\newblock \showarticletitle{When sharing economy meets IoT: towards
  fine-grained urban air quality monitoring through mobile crowdsensing on
  bike-share system}.
\newblock \bibinfo{journal}{\emph{Proceedings of the ACM on Interactive,
  Mobile, Wearable and Ubiquitous Technologies}} \bibinfo{volume}{4},
  \bibinfo{number}{2} (\bibinfo{year}{2020}), \bibinfo{pages}{1--26}.
\newblock


\bibitem[\protect\citeauthoryear{Wu, Pan, Chen, Long, Zhang, and Philip}{Wu
  et~al\mbox{.}}{2020a}]%
        {wu2020comprehensive}
\bibfield{author}{\bibinfo{person}{Zonghan Wu}, \bibinfo{person}{Shirui Pan},
  \bibinfo{person}{Fengwen Chen}, \bibinfo{person}{Guodong Long},
  \bibinfo{person}{Chengqi Zhang}, {and} \bibinfo{person}{S~Yu Philip}.}
  \bibinfo{year}{2020}\natexlab{a}.
\newblock \showarticletitle{A comprehensive survey on graph neural networks}.
\newblock \bibinfo{journal}{\emph{IEEE transactions on neural networks and
  learning systems}} \bibinfo{volume}{32}, \bibinfo{number}{1}
  (\bibinfo{year}{2020}), \bibinfo{pages}{4--24}.
\newblock


\bibitem[\protect\citeauthoryear{Wu, Wang, and Zhang}{Wu et~al\mbox{.}}{2019}]%
        {wu2019msstn}
\bibfield{author}{\bibinfo{person}{Zhiyuan Wu}, \bibinfo{person}{Yue Wang},
  {and} \bibinfo{person}{Lin Zhang}.} \bibinfo{year}{2019}\natexlab{}.
\newblock \showarticletitle{Msstn: Multi-scale spatial temporal network for air
  pollution prediction}. In \bibinfo{booktitle}{\emph{2019 IEEE International
  Conference on Big Data (Big Data)}}. IEEE, \bibinfo{pages}{1547--1556}.
\newblock


\bibitem[\protect\citeauthoryear{Xiong, Vahedian, Zhou, Li, and Luo}{Xiong
  et~al\mbox{.}}{2018}]%
        {xiong2018predicting}
\bibfield{author}{\bibinfo{person}{Haoyi Xiong}, \bibinfo{person}{Amin
  Vahedian}, \bibinfo{person}{Xun Zhou}, \bibinfo{person}{Yanhua Li}, {and}
  \bibinfo{person}{Jun Luo}.} \bibinfo{year}{2018}\natexlab{}.
\newblock \showarticletitle{Predicting traffic congestion propagation patterns:
  A propagation graph approach}. In \bibinfo{booktitle}{\emph{Proceedings of
  the 11th ACM SIGSPATIAL International Workshop on Computational
  Transportation Science}}. \bibinfo{pages}{60--69}.
\newblock


\bibitem[\protect\citeauthoryear{Xu, Cui, Young, Wang, Hsieh, Wan, Zhang,
  et~al\mbox{.}}{Xu et~al\mbox{.}}{2020}]%
        {xu2020air}
\bibfield{author}{\bibinfo{person}{Kaijie Xu}, \bibinfo{person}{Kangping Cui},
  \bibinfo{person}{Li-Hao Young}, \bibinfo{person}{Ya-Fen Wang},
  \bibinfo{person}{Yen-Kung Hsieh}, \bibinfo{person}{Shun Wan},
  \bibinfo{person}{Jiajia Zhang}, {et~al\mbox{.}}}
  \bibinfo{year}{2020}\natexlab{}.
\newblock \showarticletitle{Air quality index, indicatory air pollutants and
  impact of COVID-19 event on the air quality near central China}.
\newblock \bibinfo{journal}{\emph{Aerosol and Air Quality Research}}
  \bibinfo{volume}{20}, \bibinfo{number}{6} (\bibinfo{year}{2020}),
  \bibinfo{pages}{1204--1221}.
\newblock


\bibitem[\protect\citeauthoryear{Xu and Zhu}{Xu and Zhu}{2016}]%
        {xu2016remote}
\bibfield{author}{\bibinfo{person}{Yanan Xu} {and} \bibinfo{person}{Yanmin
  Zhu}.} \bibinfo{year}{2016}\natexlab{}.
\newblock \showarticletitle{When remote sensing data meet ubiquitous urban
  data: Fine-grained air quality inference}. In \bibinfo{booktitle}{\emph{2016
  IEEE International Conference on Big Data (Big Data)}}. IEEE,
  \bibinfo{pages}{1252--1261}.
\newblock


\bibitem[\protect\citeauthoryear{Xu, Zhu, Shen, and Yu}{Xu
  et~al\mbox{.}}{2019}]%
        {xu2019fine}
\bibfield{author}{\bibinfo{person}{Yanan Xu}, \bibinfo{person}{Yanmin Zhu},
  \bibinfo{person}{Yanyan Shen}, {and} \bibinfo{person}{Jiadi Yu}.}
  \bibinfo{year}{2019}\natexlab{}.
\newblock \showarticletitle{Fine-grained air quality inference with remote
  sensing data and ubiquitous urban data}.
\newblock \bibinfo{journal}{\emph{ACM Transactions on Knowledge Discovery from
  Data (TKDD)}} \bibinfo{volume}{13}, \bibinfo{number}{5}
  (\bibinfo{year}{2019}), \bibinfo{pages}{1--27}.
\newblock


\bibitem[\protect\citeauthoryear{Yan and Han}{Yan and Han}{2002}]%
        {yan2002gspan}
\bibfield{author}{\bibinfo{person}{Xifeng Yan} {and} \bibinfo{person}{Jiawei
  Han}.} \bibinfo{year}{2002}\natexlab{}.
\newblock \showarticletitle{gspan: Graph-based substructure pattern mining}. In
  \bibinfo{booktitle}{\emph{2002 IEEE International Conference on Data Mining,
  2002. Proceedings.}} IEEE, \bibinfo{pages}{721--724}.
\newblock


\bibitem[\protect\citeauthoryear{Yi, Duan, Li, Zhang, Li, and Zheng}{Yi
  et~al\mbox{.}}{2020}]%
        {yi2020predicting}
\bibfield{author}{\bibinfo{person}{Xiuwen Yi}, \bibinfo{person}{Zhewen Duan},
  \bibinfo{person}{Ruiyuan Li}, \bibinfo{person}{Junbo Zhang},
  \bibinfo{person}{Tianrui Li}, {and} \bibinfo{person}{Yu Zheng}.}
  \bibinfo{year}{2020}\natexlab{}.
\newblock \showarticletitle{Predicting fine-grained air quality based on deep
  neural networks}.
\newblock \bibinfo{journal}{\emph{IEEE Transactions on Big Data}}
  \bibinfo{volume}{8}, \bibinfo{number}{5} (\bibinfo{year}{2020}),
  \bibinfo{pages}{1326--1339}.
\newblock


\bibitem[\protect\citeauthoryear{Yi, Zhang, Wang, Li, and Zheng}{Yi
  et~al\mbox{.}}{2018}]%
        {yi2018deep}
\bibfield{author}{\bibinfo{person}{Xiuwen Yi}, \bibinfo{person}{Junbo Zhang},
  \bibinfo{person}{Zhaoyuan Wang}, \bibinfo{person}{Tianrui Li}, {and}
  \bibinfo{person}{Yu Zheng}.} \bibinfo{year}{2018}\natexlab{}.
\newblock \showarticletitle{Deep distributed fusion network for air quality
  prediction}. In \bibinfo{booktitle}{\emph{Proceedings of the 24th ACM SIGKDD
  international conference on knowledge discovery \& data mining}}.
  \bibinfo{pages}{965--973}.
\newblock


\bibitem[\protect\citeauthoryear{Yi, Zheng, Zhang, and Li}{Yi
  et~al\mbox{.}}{2016}]%
        {yi2016st}
\bibfield{author}{\bibinfo{person}{Xiuwen Yi}, \bibinfo{person}{Yu Zheng},
  \bibinfo{person}{Junbo Zhang}, {and} \bibinfo{person}{Tianrui Li}.}
  \bibinfo{year}{2016}\natexlab{}.
\newblock \showarticletitle{ST-MVL: filling missing values in geo-sensory time
  series data}. In \bibinfo{booktitle}{\emph{Proceedings of the 25th
  International Joint Conference on Artificial Intelligence}}.
\newblock


\bibitem[\protect\citeauthoryear{Yu, Hu, Zhou, Guo, Yang, and Li}{Yu
  et~al\mbox{.}}{2023}]%
        {yu2023cgf}
\bibfield{author}{\bibinfo{person}{Haomin Yu}, \bibinfo{person}{Jilin Hu},
  \bibinfo{person}{Xinyuan Zhou}, \bibinfo{person}{Chenjuan Guo},
  \bibinfo{person}{Bin Yang}, {and} \bibinfo{person}{Qingyong Li}.}
  \bibinfo{year}{2023}\natexlab{}.
\newblock \showarticletitle{CGF: A Category Guidance Based PM $ \_ $\{$2.5$\}$
  $ Sequence Forecasting Training Framework}.
\newblock \bibinfo{journal}{\emph{IEEE Transactions on Knowledge and Data
  Engineering}} (\bibinfo{year}{2023}).
\newblock


\bibitem[\protect\citeauthoryear{Yu, Li, Geng, Zhang, and Wei}{Yu
  et~al\mbox{.}}{2020}]%
        {yu2020airnet}
\bibfield{author}{\bibinfo{person}{Haomin Yu}, \bibinfo{person}{Qingyong Li},
  \bibinfo{person}{Yangli-ao Geng}, \bibinfo{person}{Yingjun Zhang}, {and}
  \bibinfo{person}{Zhi Wei}.} \bibinfo{year}{2020}\natexlab{}.
\newblock \showarticletitle{Airnet: A calibration model for low-cost air
  monitoring sensors using dual sequence encoder networks}. In
  \bibinfo{booktitle}{\emph{Proceedings of the AAAI Conference on Artificial
  Intelligence}}, Vol.~\bibinfo{volume}{34}. \bibinfo{pages}{1129--1136}.
\newblock


\bibitem[\protect\citeauthoryear{Zhan, Xu, Zhang, Zhu, Yin, and Zheng}{Zhan
  et~al\mbox{.}}{2022}]%
        {zhan2022deepthermal}
\bibfield{author}{\bibinfo{person}{Xianyuan Zhan}, \bibinfo{person}{Haoran Xu},
  \bibinfo{person}{Yue Zhang}, \bibinfo{person}{Xiangyu Zhu},
  \bibinfo{person}{Honglei Yin}, {and} \bibinfo{person}{Yu Zheng}.}
  \bibinfo{year}{2022}\natexlab{}.
\newblock \showarticletitle{Deepthermal: Combustion optimization for thermal
  power generating units using offline reinforcement learning}. In
  \bibinfo{booktitle}{\emph{Proceedings of the AAAI Conference on Artificial
  Intelligence}}, Vol.~\bibinfo{volume}{36}. \bibinfo{pages}{4680--4688}.
\newblock


\bibitem[\protect\citeauthoryear{Zhang, Zheng, Ma, and Han}{Zhang
  et~al\mbox{.}}{2015}]%
        {zhang2015assembler}
\bibfield{author}{\bibinfo{person}{Chao Zhang}, \bibinfo{person}{Yu Zheng},
  \bibinfo{person}{Xiuli Ma}, {and} \bibinfo{person}{Jiawei Han}.}
  \bibinfo{year}{2015}\natexlab{}.
\newblock \showarticletitle{Assembler: Efficient discovery of spatial
  co-evolving patterns in massive geo-sensory data}. In
  \bibinfo{booktitle}{\emph{Proceedings of the 21th ACM SIGKDD international
  conference on Knowledge discovery and data mining}}.
  \bibinfo{pages}{1415--1424}.
\newblock


\bibitem[\protect\citeauthoryear{Zhang, Yao, Sun, and Tay}{Zhang
  et~al\mbox{.}}{2019b}]%
        {zhang2019deep}
\bibfield{author}{\bibinfo{person}{Shuai Zhang}, \bibinfo{person}{Lina Yao},
  \bibinfo{person}{Aixin Sun}, {and} \bibinfo{person}{Yi Tay}.}
  \bibinfo{year}{2019}\natexlab{b}.
\newblock \showarticletitle{Deep learning based recommender system: A survey
  and new perspectives}.
\newblock \bibinfo{journal}{\emph{ACM computing surveys (CSUR)}}
  \bibinfo{volume}{52}, \bibinfo{number}{1} (\bibinfo{year}{2019}),
  \bibinfo{pages}{1--38}.
\newblock


\bibitem[\protect\citeauthoryear{Zhang, Liu, Han, Ge, and Xiong}{Zhang
  et~al\mbox{.}}{2022}]%
        {zhang2022multi}
\bibfield{author}{\bibinfo{person}{Weijia Zhang}, \bibinfo{person}{Hao Liu},
  \bibinfo{person}{Jindong Han}, \bibinfo{person}{Yong Ge}, {and}
  \bibinfo{person}{Hui Xiong}.} \bibinfo{year}{2022}\natexlab{}.
\newblock \showarticletitle{Multi-agent graph convolutional reinforcement
  learning for dynamic electric vehicle charging pricing}. In
  \bibinfo{booktitle}{\emph{Proceedings of the 28th ACM SIGKDD conference on
  knowledge discovery and data mining}}. \bibinfo{pages}{2471--2481}.
\newblock


\bibitem[\protect\citeauthoryear{Zhang, Liu, Liu, Zhou, and Xiong}{Zhang
  et~al\mbox{.}}{2020}]%
        {zhang2020semi}
\bibfield{author}{\bibinfo{person}{Weijia Zhang}, \bibinfo{person}{Hao Liu},
  \bibinfo{person}{Yanchi Liu}, \bibinfo{person}{Jingbo Zhou}, {and}
  \bibinfo{person}{Hui Xiong}.} \bibinfo{year}{2020}\natexlab{}.
\newblock \showarticletitle{Semi-supervised hierarchical recurrent graph neural
  network for city-wide parking availability prediction}. In
  \bibinfo{booktitle}{\emph{Proceedings of the AAAI Conference on Artificial
  Intelligence}}, Vol.~\bibinfo{volume}{34}. \bibinfo{pages}{1186--1193}.
\newblock


\bibitem[\protect\citeauthoryear{Zhang, Liu, Wang, Xu, Xin, Dou, and
  Xiong}{Zhang et~al\mbox{.}}{2021b}]%
        {zhang2021intelligent}
\bibfield{author}{\bibinfo{person}{Weijia Zhang}, \bibinfo{person}{Hao Liu},
  \bibinfo{person}{Fan Wang}, \bibinfo{person}{Tong Xu},
  \bibinfo{person}{Haoran Xin}, \bibinfo{person}{Dejing Dou}, {and}
  \bibinfo{person}{Hui Xiong}.} \bibinfo{year}{2021}\natexlab{b}.
\newblock \showarticletitle{Intelligent electric vehicle charging
  recommendation based on multi-agent reinforcement learning}. In
  \bibinfo{booktitle}{\emph{Proceedings of the Web Conference 2021}}.
  \bibinfo{pages}{1856--1867}.
\newblock


\bibitem[\protect\citeauthoryear{Zhang, Bocquet, Mallet, Seigneur, and
  Baklanov}{Zhang et~al\mbox{.}}{2012a}]%
        {zhang2012real1}
\bibfield{author}{\bibinfo{person}{Yang Zhang}, \bibinfo{person}{Marc Bocquet},
  \bibinfo{person}{Vivien Mallet}, \bibinfo{person}{Christian Seigneur}, {and}
  \bibinfo{person}{Alexander Baklanov}.} \bibinfo{year}{2012}\natexlab{a}.
\newblock \showarticletitle{Real-time air quality forecasting, part I: History,
  techniques, and current status}.
\newblock \bibinfo{journal}{\emph{Atmospheric Environment}}
  \bibinfo{volume}{60} (\bibinfo{year}{2012}), \bibinfo{pages}{632--655}.
\newblock


\bibitem[\protect\citeauthoryear{Zhang, Bocquet, Mallet, Seigneur, and
  Baklanov}{Zhang et~al\mbox{.}}{2012b}]%
        {zhang2012real2}
\bibfield{author}{\bibinfo{person}{Yang Zhang}, \bibinfo{person}{Marc Bocquet},
  \bibinfo{person}{Vivien Mallet}, \bibinfo{person}{Christian Seigneur}, {and}
  \bibinfo{person}{Alexander Baklanov}.} \bibinfo{year}{2012}\natexlab{b}.
\newblock \showarticletitle{Real-time air quality forecasting, part II: State
  of the science, current research needs, and future prospects}.
\newblock \bibinfo{journal}{\emph{Atmospheric Environment}}
  \bibinfo{volume}{60} (\bibinfo{year}{2012}), \bibinfo{pages}{656--676}.
\newblock


\bibitem[\protect\citeauthoryear{Zhang, Hannigan, and Lv}{Zhang
  et~al\mbox{.}}{2021a}]%
        {zhang2021air}
\bibfield{author}{\bibinfo{person}{Yawen Zhang}, \bibinfo{person}{Michael
  Hannigan}, {and} \bibinfo{person}{Qin Lv}.} \bibinfo{year}{2021}\natexlab{a}.
\newblock \showarticletitle{Air Pollution Hotspot Detection and Source Feature
  Analysis using Cross-Domain Urban Data}. In
  \bibinfo{booktitle}{\emph{Proceedings of the 29th International Conference on
  Advances in Geographic Information Systems}}. \bibinfo{pages}{592--595}.
\newblock


\bibitem[\protect\citeauthoryear{Zhang, Lv, Gao, Shen, Dick, Hannigan, and
  Liu}{Zhang et~al\mbox{.}}{2019a}]%
        {zhang2019multi}
\bibfield{author}{\bibinfo{person}{Yawen Zhang}, \bibinfo{person}{Qin Lv},
  \bibinfo{person}{Duanfeng Gao}, \bibinfo{person}{Si Shen},
  \bibinfo{person}{Robert~P Dick}, \bibinfo{person}{Michael Hannigan}, {and}
  \bibinfo{person}{Qi Liu}.} \bibinfo{year}{2019}\natexlab{a}.
\newblock \showarticletitle{Multi-Group Encoder-Decoder Networks to Fuse
  Heterogeneous Data for Next-Day Air Quality Prediction.}. In
  \bibinfo{booktitle}{\emph{IJCAI}}. \bibinfo{pages}{4341--4347}.
\newblock


\bibitem[\protect\citeauthoryear{Zhao, Xu, Fu, Chen, and Guo}{Zhao
  et~al\mbox{.}}{2017}]%
        {zhao2017incorporating}
\bibfield{author}{\bibinfo{person}{Xiangyu Zhao}, \bibinfo{person}{Tong Xu},
  \bibinfo{person}{Yanjie Fu}, \bibinfo{person}{Enhong Chen}, {and}
  \bibinfo{person}{Hao Guo}.} \bibinfo{year}{2017}\natexlab{}.
\newblock \showarticletitle{Incorporating spatio-temporal smoothness for air
  quality inference}. In \bibinfo{booktitle}{\emph{2017 IEEE International
  Conference on Data Mining (ICDM)}}. IEEE, \bibinfo{pages}{1177--1182}.
\newblock


\bibitem[\protect\citeauthoryear{Zheng, Wang, Sun, Zhang, and Kahn}{Zheng
  et~al\mbox{.}}{2019}]%
        {zheng2019air}
\bibfield{author}{\bibinfo{person}{Siqi Zheng}, \bibinfo{person}{Jianghao
  Wang}, \bibinfo{person}{Cong Sun}, \bibinfo{person}{Xiaonan Zhang}, {and}
  \bibinfo{person}{Matthew~E Kahn}.} \bibinfo{year}{2019}\natexlab{}.
\newblock \showarticletitle{Air pollution lowers Chinese urbanites’ expressed
  happiness on social media}.
\newblock \bibinfo{journal}{\emph{Nature human behaviour}} \bibinfo{volume}{3},
  \bibinfo{number}{3} (\bibinfo{year}{2019}), \bibinfo{pages}{237--243}.
\newblock


\bibitem[\protect\citeauthoryear{Zheng}{Zheng}{2015}]%
        {zheng2015methodologies}
\bibfield{author}{\bibinfo{person}{Yu Zheng}.} \bibinfo{year}{2015}\natexlab{}.
\newblock \showarticletitle{Methodologies for cross-domain data fusion: An
  overview}.
\newblock \bibinfo{journal}{\emph{IEEE transactions on big data}}
  \bibinfo{volume}{1}, \bibinfo{number}{1} (\bibinfo{year}{2015}),
  \bibinfo{pages}{16--34}.
\newblock


\bibitem[\protect\citeauthoryear{Zheng, Capra, Wolfson, and Yang}{Zheng
  et~al\mbox{.}}{2014}]%
        {zheng2014urban}
\bibfield{author}{\bibinfo{person}{Yu Zheng}, \bibinfo{person}{Licia Capra},
  \bibinfo{person}{Ouri Wolfson}, {and} \bibinfo{person}{Hai Yang}.}
  \bibinfo{year}{2014}\natexlab{}.
\newblock \showarticletitle{Urban computing: concepts, methodologies, and
  applications}.
\newblock \bibinfo{journal}{\emph{ACM Transactions on Intelligent Systems and
  Technology (TIST)}} \bibinfo{volume}{5}, \bibinfo{number}{3}
  (\bibinfo{year}{2014}), \bibinfo{pages}{1--55}.
\newblock


\bibitem[\protect\citeauthoryear{Zheng, Liu, and Hsieh}{Zheng
  et~al\mbox{.}}{2013}]%
        {zheng2013u}
\bibfield{author}{\bibinfo{person}{Yu Zheng}, \bibinfo{person}{Furui Liu},
  {and} \bibinfo{person}{Hsun-Ping Hsieh}.} \bibinfo{year}{2013}\natexlab{}.
\newblock \showarticletitle{U-air: When urban air quality inference meets big
  data}. In \bibinfo{booktitle}{\emph{Proceedings of the 19th ACM SIGKDD
  international conference on Knowledge discovery and data mining}}.
  \bibinfo{pages}{1436--1444}.
\newblock


\bibitem[\protect\citeauthoryear{Zheng, Yi, Li, Li, Shan, Chang, and Li}{Zheng
  et~al\mbox{.}}{2015}]%
        {zheng2015forecasting}
\bibfield{author}{\bibinfo{person}{Yu Zheng}, \bibinfo{person}{Xiuwen Yi},
  \bibinfo{person}{Ming Li}, \bibinfo{person}{Ruiyuan Li},
  \bibinfo{person}{Zhangqing Shan}, \bibinfo{person}{Eric Chang}, {and}
  \bibinfo{person}{Tianrui Li}.} \bibinfo{year}{2015}\natexlab{}.
\newblock \showarticletitle{Forecasting fine-grained air quality based on big
  data}. In \bibinfo{booktitle}{\emph{Proceedings of the 21th ACM SIGKDD
  international conference on knowledge discovery and data mining}}.
  \bibinfo{pages}{2267--2276}.
\newblock


\bibitem[\protect\citeauthoryear{Zhong, Zhang, Bagheri, Burken, Gu, Li, Ma,
  Marrone, Ren, Schrier, et~al\mbox{.}}{Zhong et~al\mbox{.}}{2021}]%
        {zhong2021machine}
\bibfield{author}{\bibinfo{person}{Shifa Zhong}, \bibinfo{person}{Kai Zhang},
  \bibinfo{person}{Majid Bagheri}, \bibinfo{person}{Joel~G Burken},
  \bibinfo{person}{April Gu}, \bibinfo{person}{Baikun Li},
  \bibinfo{person}{Xingmao Ma}, \bibinfo{person}{Babetta~L Marrone},
  \bibinfo{person}{Zhiyong~Jason Ren}, \bibinfo{person}{Joshua Schrier},
  {et~al\mbox{.}}} \bibinfo{year}{2021}\natexlab{}.
\newblock \showarticletitle{Machine learning: new ideas and tools in
  environmental science and engineering}.
\newblock \bibinfo{journal}{\emph{Environmental Science \& Technology}}
  \bibinfo{volume}{55}, \bibinfo{number}{19} (\bibinfo{year}{2021}),
  \bibinfo{pages}{12741--12754}.
\newblock


\bibitem[\protect\citeauthoryear{Zhou and Li}{Zhou and Li}{2005}]%
        {zhou2005tri}
\bibfield{author}{\bibinfo{person}{Zhi-Hua Zhou} {and} \bibinfo{person}{Ming
  Li}.} \bibinfo{year}{2005}\natexlab{}.
\newblock \showarticletitle{Tri-training: Exploiting unlabeled data using three
  classifiers}.
\newblock \bibinfo{journal}{\emph{IEEE Transactions on knowledge and Data
  Engineering}} \bibinfo{volume}{17}, \bibinfo{number}{11}
  (\bibinfo{year}{2005}), \bibinfo{pages}{1529--1541}.
\newblock


\bibitem[\protect\citeauthoryear{Zhu, Zhang, Zhang, Zhi, Li, Han, and
  Zheng}{Zhu et~al\mbox{.}}{2017}]%
        {zhu2017pg}
\bibfield{author}{\bibinfo{person}{Julie~Yixuan Zhu}, \bibinfo{person}{Chao
  Zhang}, \bibinfo{person}{Huichu Zhang}, \bibinfo{person}{Shi Zhi},
  \bibinfo{person}{Victor~OK Li}, \bibinfo{person}{Jiawei Han}, {and}
  \bibinfo{person}{Yu Zheng}.} \bibinfo{year}{2017}\natexlab{}.
\newblock \showarticletitle{pg-causality: Identifying spatiotemporal causal
  pathways for air pollutants with urban big data}.
\newblock \bibinfo{journal}{\emph{IEEE Transactions on Big Data}}
  \bibinfo{volume}{4}, \bibinfo{number}{4} (\bibinfo{year}{2017}),
  \bibinfo{pages}{571--585}.
\newblock


\end{thebibliography}
\bibliographystyle{ACM-Reference-Format}

\end{document}